\documentclass{article} 
\usepackage{iclr2026_conference,times}

\usepackage{amsmath,amsfonts,bm}

\def\eqref#1{equation~\ref{#1}}

\def\1{\bm{1}}

\DeclareMathAlphabet{\mathsfit}{\encodingdefault}{\sfdefault}{m}{sl}
\SetMathAlphabet{\mathsfit}{bold}{\encodingdefault}{\sfdefault}{bx}{n}

\usepackage{hyperref}
\usepackage{url}
\usepackage{amsmath,amssymb,amsfonts}
\usepackage{algorithmic}
\usepackage{algorithm}
\usepackage{array}
\usepackage[caption=false,font=normalsize,labelfont=sf,textfont=sf]{subfig}
\usepackage{textcomp}
\usepackage{stfloats}
\usepackage{verbatim}
\usepackage{graphicx}
\usepackage{tabularx}
\usepackage{multirow}
\usepackage{booktabs}
\usepackage[english]{babel}
\usepackage{amsthm}
\usepackage{bm}
\usepackage{dsfont}

\title{Autonomous Source Knowledge Selection in Multi-Domain Adaptation}

\author{Keqiuyin Li \\
Australian Artificial Intelligence Institute\\
University of Technology Sydney\\
Sydney, NSW, Australia \\
\texttt{\{keqiuyin.li\}@uts.edu.au} \\
\And
Jie Lu \\
Australian Artificial Intelligence Institute\\
University of Technology Sydney\\
Sydney, NSW, Australia \\
\texttt{\{jie.lu\}@uts.edu.au} \\
\And
Hua Zuo \\
Australian Artificial Intelligence Institute\\
University of Technology Sydney\\
Sydney, NSW, Australia \\
\texttt{\{hua.zuo\}@uts.edu.au} \\
\And
Guangquan Zhang \\
Australian Artificial Intelligence Institute\\
University of Technology Sydney\\
Sydney, NSW, Australia \\
\texttt{\{guangquan.zhang\}@uts.edu.au} 
}

\iclrfinalcopy 
\begin{document}

\maketitle

\begin{abstract}
Unsupervised multi-domain adaptation plays a key role in transfer learning by leveraging acquired rich source information from multiple source domains to solve target task from an unlabeled target domain. However, multiple source domains often contain much redundant or unrelated information which can harm transfer performance, especially when in massive-source domain settings. It is urgent to develop effective strategies for identifying and selecting the most transferable knowledge from massive source domains to address the target task. In this paper, we propose a multi-domain adaptation method named \underline{\textit{Auto}}nomous Source Knowledge \underline{\textit{S}}election (AutoS) to autonomosly select source training samples and models, enabling the prediction of target task using more relevant and transferable source information. The proposed method employs a density-driven selection strategy to choose source samples during training and to determine which source models should contribute to target prediction. Simulteneously, a pseudo-label enhancement module built on a pre-trained multimodal modal is employed to mitigate target label noise and improve self-supervision. Experiments on real-world datasets indicate the superiority of the proposed method.
\end{abstract}

\section{Introduction}
Domain adaptation has achieved significant progress in transfer learning by addressing data scarcity in the target domain through the utilization of knowledge from source domain(s). Typical domain adaptation methods involve various techniques to overcome distribution shifts and modal gaps, including approaches based on feature alignment \citep{624bai2024prompt, yu2024fuzzy} and those relying on model fine-tuning \citep{592zhang2023rethinking}. Feature alignment–based methods focus on matching source and target data in a latent feature space to bridge their gaps, commonly employing techniques such as maximum mean discrepancy (MMD) \citep{646ning2025multilevel}, Kullback–Leibler (KL) divergence \citep{647schlachter2025memory}, Wasserstein distance \citep{648he2024gradual}, and adversarial learning \citep{649fang2024prototype}. In contrast, model fine-tuning methods \citep{578zhang2023class, 570li2024agile} primarily rely on pseudo-label estimation and self-training strategies to adapt the model, making label denoising techniques a critical component for ensuring performance \citep{589litrico2023guiding}.

To enhance the generality of domain adaptation in handling complex scenarios, category shifts have been considered, leading to the development of partial \citep{505kong2022partial}, open-set  \citep{595wan2024unveiling}, and universal \citep{550qu2024lead} domain adaptation, where out-of-distribution detection techniques serve as an effective solution.
Simultaneously, extracting transferable knowledge from multiple source domains \citep{556ma2024multi} and modalities \citep{489zhang2023rethinking, yangadapting, 627zhang2025release} has emerged as a prominent approach for leveraging richer information to enhance transfer performance. To integrate knowledge, linear combination is a widely used approach for fusing features or predictions from multiple source domains or modalities, including both simple averaging \citep{109zhao2020multi}, weighted averaging \citep{330dong2021confident} and federated learning \citep{535huang2023rethinking} strategies. 

\begin{figure}[!htbp]
\centering
\includegraphics[height=3.5cm,width=10cm]{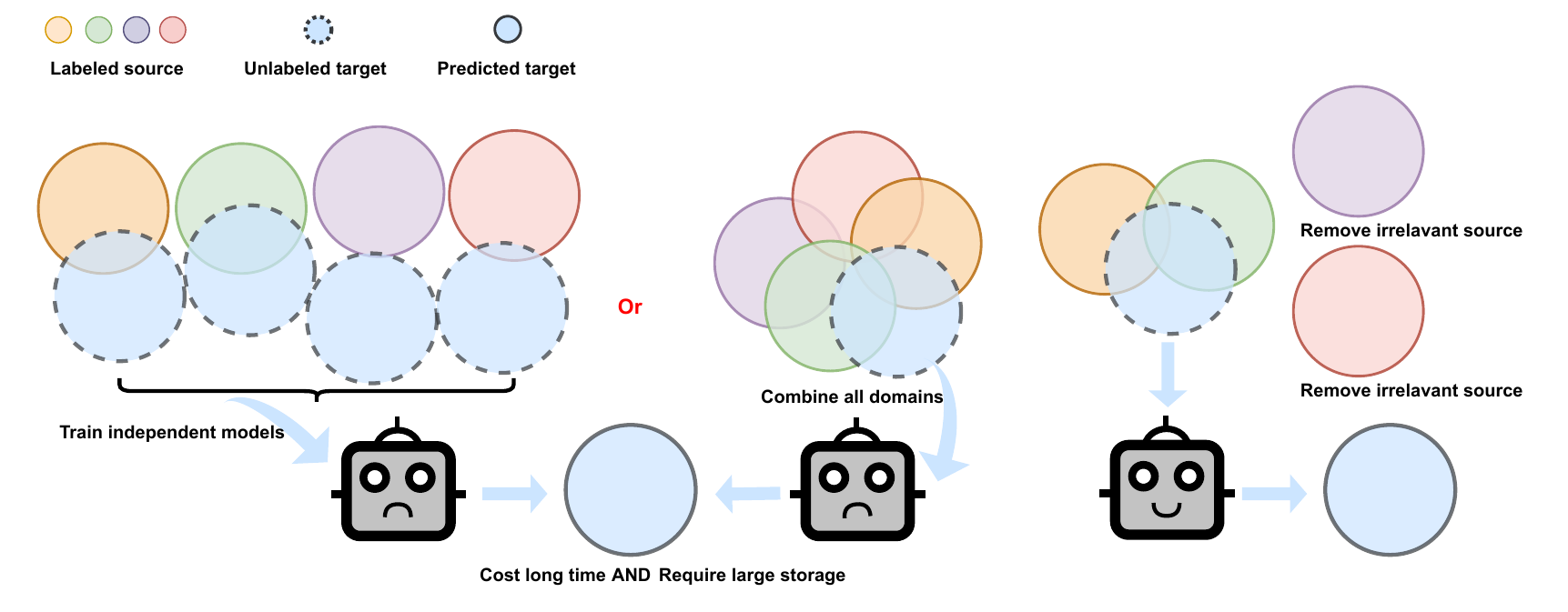}
\caption{Illustration of multi-domain adaptation. The left figure exemplifies that training independent models by matching each source-target pair or by combining all domains can cause high computational and storage costs, while overlooking the fact that irrelevant sources can degrade transfer performance. The right figure depicts an ideal solution in which irrelevant source domains are removed while relevant ones are retained.}\label{massive}
\end{figure}

However, knowledge from different sources also introduce challenges, particularly when dealing with massive numbers of samples or domains, as shown in Figure \ref{massive}. Previous multi-domain adaptation methods typically combine information from all source domains, but seldom focus on selecting one or more relevant domains. Limited research has explored reducing the number of transferred source domains, and such distillation processes often involve manual operations. One represent work is   sample and source distillation, \citep{457li2023multidomain}. It selects source samples and domains that are more similar to the target domain to enhance transfer performance. However, this method requires training source models twice, once for selection and once for adaptation, which is impractical when dealing with massive source domains due to the high computational cost. Automatically and effectively selecting source knowledge from massive domains remains an urgent challenge in multi-domain adaptation.

\graphicspath{{img/}}

Considering the mentioned gaps, in this paper, we propose an autonomous source knowledge selection method to achieve target-oriented domain adaptation, which reduces the number of source domains involved in transfer without degrading performance. The proposed method introduces a density-controlled sample collection strategy to construct an intermediate domain composed of high-confidence source and target samples. Simultaneously, guided by the proportion of confident target samples and the target data density relative to each source domain, source domains with low similarity are removed during training and subsequent adaptation. In this way, multi-domain adaptation can be achieved by automatically identifying useful source domains while progressively discarding irrelevant domains during training, thereby avoiding the introduction of noisy knowledge and facilitating learning, particularly when dealing with massive source domains.

Our contributions can be summarized as: (1) An autonomous source knowledge selection method for multi-domain adaptation. This benefits the target domain by pruning redundant or dissimilar source domains and suppressing noisy information. (2) A target-oriented multi-domain adaptation approach for enhancing downstream task performance through prompt tuning across different modalities. This enables domain adaptation without data matching, making it flexible for transfer learning with or without access to source data. (3) A density-controlled sample collection strategy for gathering high-confident source and target samples. This benefits domain adaptation by constructing an intermediate domain and introducing self-supervision for the target task.

The remainder of this paper is organized as follows: Section \ref{rw} reviews related work; Section \ref{tpm} details the proposed autonomous source knowledge selection method; Section \ref{exp} presents experiments and analysis on real-world datasets; and Section \ref{future} concludes with directions for future research.

\section{Related Work}
\label{rw}
\subsection{Multi-Domain Adaptation}
Multi-domain adaptation achieve great progress in recent. 
Contrastive adversarial learning \citep{652wilson2023calda} deals with multi-source time series domain adaptation by aligning cross-source label information. In this framework, adversarial learning guides a domain classifier to distinguish whether a sample is from the source or target domain, while contrastive learning enforces intra-class closer together and inter-class separation.
Prototype-based mean teacher \citep{651belal2024multi} employs class prototypes to encode specific information from multiple domains, where the contrastive loss is used to align intra-class knowledge while separating inter-class knowledge across domains.
Multiple adaptation network \citep{650lu2025multiple} addresses multi-source and multi-target domain adaptation, which uses multiple alignment strategies to align both features and classifiers that are relevant. Style information transfer is also considered to fully leverage knowledge from multiple target domains. 

\subsection{Federated Learning}
Federated learning is a widely used strategy that avoids training on independent source domains while leveraging the benefits of multiple sources. 
Universal federated domain adaptation \citep{558liu2024ufda} defines a hot-learning strategy with
contrastive label disambiguation, which solves category shifts through one-hot outputs of source models and detects unknown categories using a cluster-level strategy built based on the consensus knowledge across source and target domains.
Heterogeneous fuzzy domain adaptation \citep{628li2025fuzzy} integrates federated training and fuzzy logic to enable domain adaptation without requiring access to source data or models. A transformation module is introduced to address heterogeneity, while self-knowledge distillation is employed to federally construct the target model by simulating the predictions of source models.

\subsection{Multimodal Domain Adaptation}
Distilling multimodal foundation model  \citep{576tang2024source} achieves source-free domain adaptation through a two-step process, including customizing the vision–language model via prompt learning to minimize mutual information and distilling knowledge based on the target domain.
Text-image alignment network \citep{654kondapaneni2024text} extends stable diffusion to leverage perceptual knowledge for predicting visual tasks from text-based generative prompts. It employs model personalization and caption modification to adapt the pre-trained model to the target domain, achieving improvements over unaligned baselines.
Text-free graph foundation model \citep{626yu2025samgpt} introduces a novel structure alignment framework to learn multi-domain knowledge from graphs originating in multiple source domains. It adapts to unseen target domains by incorporating a set of structure tokens and dual prompts, thereby unifying domain-specific information with structural knowledge.

\section{The Proposed Autonomous Source Knowledge Selection Method}
\label{tpm}
\subsection{Overview}
\begin{figure}[h]
\centering
\includegraphics[width=\linewidth,keepaspectratio]{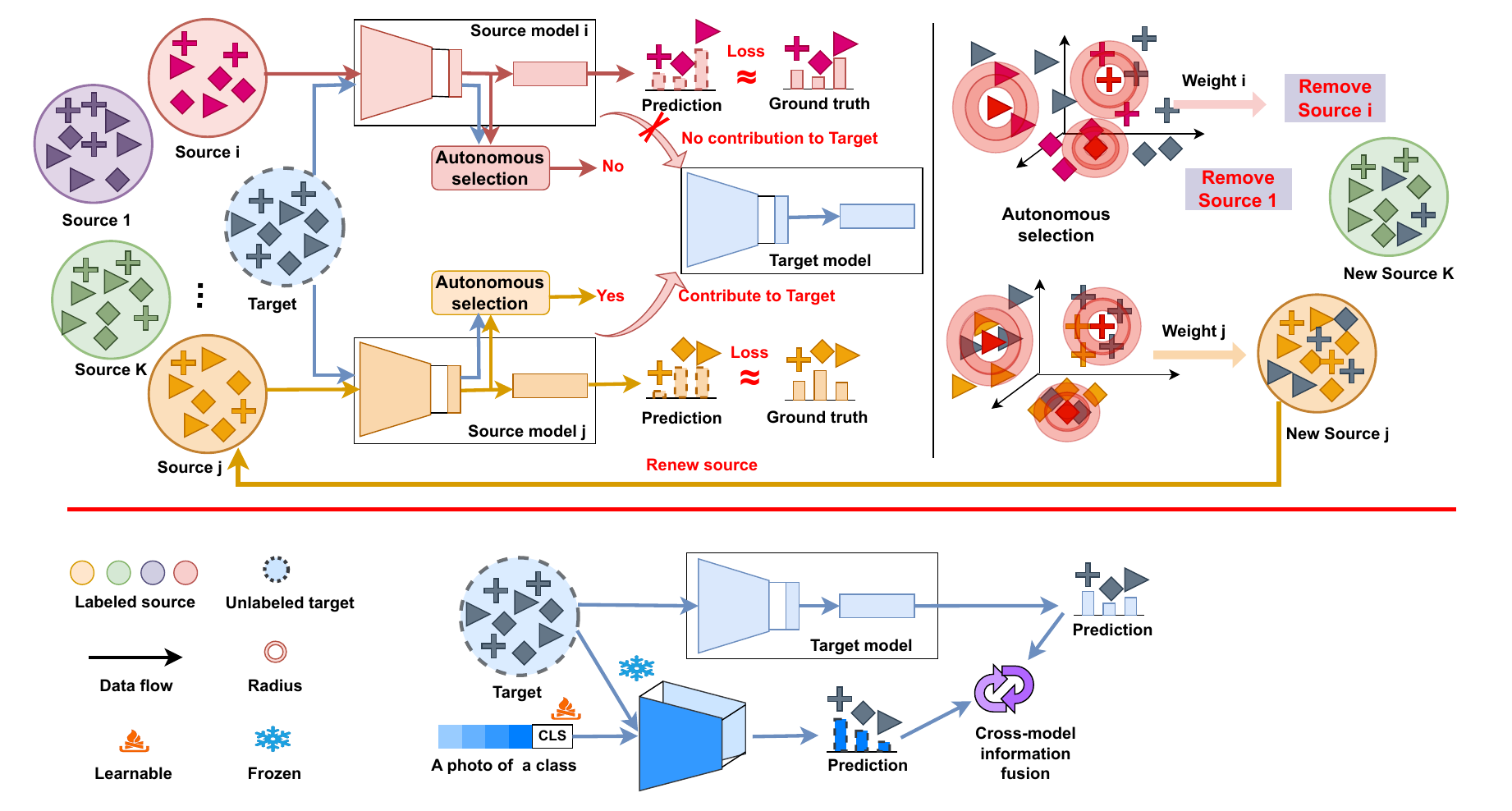}
\caption{Whole framework of the proposed autonomous source knowledge selection. The upper figure indicates source model training with target-driven autonomous source knowledge selection, the lower figure displays target model training with cross-modality transfer.}\label{framework}
\end{figure}

The whole proposed method is displayed in Figure \ref{framework}. The framework consists of two stages. 
The first stage, illustrated in the upper figure, focuses on source model training with target-driven autonomous source knowledge selection. In this phase, both source and target data are transformed into a shared latent feature space, where the similarity between each source and the target domain is evaluated based on data density. High-confidence samples are retained while irrelevant or low-similarity samples are discarded, and source domains identified as irrelevant are progressively removed during further training. 
The proposed density-controlled selection is designed not only to reduce redundancy but also to suppress noisy knowledge, thereby ensuring that the training process exploits the most transferable information.
Notably, all source models are initialized from a single shared model and updated in a federated manner, rather than being trained independently. This design avoids maintaining independent models for each source domain, thereby reducing memory requirements and computational costs. The federated model is then employed as the target model for adaptation.
The second stage, illustrated in the lower figure, involves target model training with cross-modality transfer. In this phase, knowledge selected from the previous stage is combined with a pseudo-label enhancement module and cross-modal prompts to mitigate label noise, enabling the target model to adapt without direct reliance on the full set of source data or models.

\subsection{Source Model Training}
Denote multiple source domains as $\{\mathcal{D}^s_k = \{\bm x^s_{ki}, \bm y^s_{ki}\}_{i=1}^{n_k}\}_{k=1}^K$, and the target domain as $\mathcal{D}^t = \{\bm x^t_i\}_{i=1}^n$
To handle multiple domains without training independent source models, we first initialize a global model composed of a feature extraction module $\Phi$ and a decision layer $P$. For each source domain, source samples are fed into $\Phi$ to extract features in space $\mathbb{R}^d$, which are then passed to $P$ to obtain predictions in $\mathbb{R}^{\mathcal{C}}$. By minimizing the errors between the ground truth labels and the predictions, the model is updated by:
\begin{equation}
\begin{split}
\Phi_k, P_k &= \underset{
\underset{(\bm x^s_k,\bm y^s_k) \in \mathcal{D}^s_k}{\Phi, P}}
{\arg\min} \mathcal{L}(P(\Phi(\bm x^s_k)), \mathcal{LS}({\bm y}^s_k)),  \\
\mathcal{L} &= -\frac{1}{n_k}\sum_{i=1}^{n_k} \mathcal{LS}({\bm y}^s_{ki}) \log(P(\Phi(\bm x^s_{ki}))), \\
\mathcal{LS}({\bm y}^s_k) &= (1-\mu) \bm y^s_k + \frac{\mu}{|\mathcal{C}|}, \bm {\bm y}^s_k \in \mathbb{R}^{\mathcal{C}}. 
\end{split}
\label{Ms}
\end{equation}
$\mathcal{LS}$ is the label smoothing operation to accelerate training speed.

\subsection{Autonomous Source Knowledge Selection}
To select more relevant information, we define a density-controlled selection strategy to autonomously collects highly similar source and target samples. 
Following previous findings \citep{293wang2021cross} that cluster centers are close to the mean values of normalized classifier weight vectors, we first collect cluster centers as:
\begin{equation}
\bm f^s_{kc} = \text{Norm}(P_k)c, c = 1, \cdots, \mathcal{C}; k = 1, \cdots, K.
\label{fkc}
\end{equation}
The target clustering label is:
\begin{equation}
\bm y^t_k = \underset{c}{\arg\min}(\text{Dis}(\Phi{\bm x^t}, \bm f^s_{kc})), c = 1, \cdots, \mathcal{C}; k = 1, \cdots, K.
\label{yclu}
\end{equation}
The radius of each source cluster and the radius and density of target data corresponding to source cluster centers are then calculated as:
\begin{equation}
\begin{split}
\bm r^s_{kc} = \text{Rd}(\text{Dis}(\Phi(\bm x^s_k), \bm f^s_{kc})_{\mathbb{I}_{\bm y^s_k = c}}),
\bm r^t_{kc} = \text{Rd}(\text{Dis}(\Phi(\bm x^t), \bm f^s_{kc})_{\mathbb{I}_{\bm y^t_k = c}}), 
\end{split}
\label{rkc}
\end{equation}
$\text{Dis}$ denotes the operation of computing the distance between source samples in the same cluster and their cluster center. $\text{Rd}$ means the operation for calculating cluster radius using different distance metrics, such as average distance, root mean square distance and maximum radius. Unless otherwise specified, this work employs the average radius computed using cosine distance.
Furthermore, thresholds for selecting highly confident source and target samples are defined as:
\begin{equation}
d^s_{kc} = \alpha \bm r^s_{kc} + s_{adj}, 
d^t_{kc} = \alpha \bm r^s_{kc} - t_{adj}, 
c = 1, \cdots, \mathcal{C}; k = 1, \cdots, K.
\label{dkc}
\end{equation}
$\alpha$, $s_{adj}$ and $t_{adj}$ are the parameters that control the radius used to assign values separating similar and dissimilar samples under different distance metrics. 
For any source and target sample, if  $\text{Dis}(\Phi(\bm x^s_k), \bm f^s_{kc})_{\mathbb{I}_{\bm y^s_k = c}} < d^s_{kc}$, we define the source sample as a confident sample, if  $\text{Dis}(\Phi(\bm x^t), \bm f^s_{kc})_{\underset{\bm x^t}{\arg\min}(\text{Dis}(\bm x^t, \bm f^s_{kc}))} < d^t_{kc}$, we define the target sample as a confident sample. Denote the selected source samples as $\mathcal{U}^s = \{\bm x^s_{ki}, \bm y^s_{ki}\}_{i=1}^{n^s_k}$ and target samples as $\mathcal{U}^t = \{\bm x^t_i, \bm y^t_{ki}\}_{i=1}^{n^t_k}$,
the selected high-confidence samples are grouped together to update the original source domain, retaining confident source samples, incorporating confident target samples, and discarding low-confidence source samples.

To further define if a source domain can be removed during further training, target data density can be expressed as:
\begin{equation}
\rho^t_{kc} = \frac{n^t_{kc}}{\frac{\pi^{\mathbb{R}^{d/2}}}{\Gamma(\mathbb{R}^{d/2}+1)} \bm r^t_{kc}},
c = 1, \cdots, \mathcal{C}; k = 1, \cdots, K.
\label{denst}
\end{equation}
Then, two weights are defined as:
\begin{equation}
\begin{split}
\omega_{k1} &= \frac{n^t_k}{n}, \omega_{k2} = \frac{1}{\mathcal{C}}\sum_{c=1}^{\mathcal{C}} (1-\frac{1}{1+e^{\rho^t_{kc}}}),\\
\omega_k &= \lambda \omega_{k1} + (1-\lambda) \omega_{k2}, 
c = 1, \cdots, \mathcal{C}; k = 1, \cdots, K.
\end{split}
\label{selwe}
\end{equation}
The rules for removing irrelevant source domains are defined as:
\begin{equation}
\begin{split}
\text{Keep}(\mathcal{D}^s_k) =& \left\{
\begin{array}{ccc}
TRUE, \ \text{if} \ \omega_k >= \frac{1}{K}-\sigma; \\ 
FALSE, \ \text{if} \ \omega_k < \frac{1}{K}-\sigma; \\
\end{array} \right. \\
&c = 1, \cdots, \mathcal{C}; k = 1, \cdots, K.
\end{split}
\label{keepsrc}
\end{equation}
Then we can get renewed source domains 
$\{\mathcal{D'}^s_k = \{\bm x^s_{ki}, \bm x^t_j, \bm y^s_{ki}, \bm y^t_{kj} \}_{i,j=1}^{n^s_k, n^t_k}\}_{k=1}^{K'}, \{\bm x^s_k, \bm y^s_k\} \in \mathcal{U}^s, \{\bm x^t, \bm y^t_k\} \in \mathcal{U}^t,
K' \in [1, K]$ for further training.

\subsection{Target Model Adaptation}
To adapt target model, based on the autonomously selected source domains, target model is first defined as the linear combination of models learned in equation \eqref{Ms}:
\begin{equation}
\Phi = \sum_{k=1}^{K'} \omega_k \cdot \Phi_k, P = \sum_{k=1}^{K'} \omega_k \cdot P_k, K' \in [1, K].
\label{Mt}
\end{equation}
Then the possibility of target label is predicted as:
\begin{equation}
\mathcal{P}^t = P(\Phi(\bm x^t)).
\label{Pt}
\end{equation}
At this stage, gaps remain between the pre-trained models $P$ and $\Phi$.
To bridge these gaps, we employ pseudo-labels generated by a frozen foundation model (e.g., CLIP) to self-supervise the adaptation of the target model, where only prompts are fine-tuned to take the benefits of cross-modality knowledge. Denote foundation model as $\Psi$, and original text prompts as 
$\{\mathcal{T}_c\}_{c=1}^{\mathcal{C}}$, the prediction on target domain can be expressed as:
\begin{equation}
\mathcal{P}_{FM} = \Psi(\bm x^t)_{vis} \cdot (\Psi(\mathcal{T}_c)_{txt})^T.
\label{Pfm}
\end{equation}
Following previous findings in \citep{574tang2024unified, 628li2025fuzzy} which maintain prediction consistency under a structural causal model, the external structural causal factor is defined by maximizing the correlation between the representations of random variables and the predictions of the target model, while the refinement of the desired latent factor is conditionally equivalent to a self-supervised information bottleneck. Denote latent random variables of 
$\{\mathcal{T}_c\}_{c=1}^{\mathcal{C}}$
as $\mathcal{P}_V$ following distribution of $\mathcal{P}_{FM}$, 
and the bridging values connecting text prompts and $\mathcal{P}_{FM}$ as $\mathcal{P}_V'$, the loss function of fine-tuning prompts with a Gaussian distribution is defined as following by minimizing the errors between target model prediction $\mathcal{P}^t$ and foundation model prediction $\mathcal{P}_{FM}$:
\begin{equation}
\begin{split}
\mathcal{L}_{ex} = \frac{1}{n}
\sum_{i=1}^{n} \beta [(\frac{(\mathcal{P}_{FM}^i- \mathcal{P}^{ti})^2}{\text{diag}(\mathcal{P}_{FM}^i)}  &+ \log|\text{diag}(\mathcal{P}_{FM}^i)|)  + \gamma \mathcal{KL}(g(\mathcal{P}_V')\|\mathcal{P}_V)], \\
g(\mathcal{P}_V') &= \frac{1}{\text{diag}(\mathcal{P}_V)} \odot \mathcal{P}_V. 
\end{split}
\label{exloss}
\end{equation}
$g(\cdot)$ is a learnable function providing a probability distribution with no information loss.

The cross-entropy loss for self-supervising target model is then defined as:
\begin{equation}
\begin{split}
\mathcal{L}_{in} = -\frac{1}{n}\sum_{i=1}^{n} \theta \mathcal{P}_{FM}\log{\mathcal{P}^t} + \delta \sum \bar{\mathcal{P}^t}\log \bar{\mathcal{P}^t} + \sum_{c=1}^C\mathcal{KL}(\frac{1}{n}\sum_{i=1}^{n}(\mathcal{P}^{ti})_c \| \frac{1}{C}), \bar{\mathcal{P}^t} = \frac{1}{n}\sum_{i=1}^{n} \mathcal{P}^{ti}.
\end{split}
\label{inloss}
\end{equation}

Target label is finally predicted as:
\begin{equation}
\bm y^t = \arg\max \ \text{SoftMax}(P(\Phi(\bm x^t))).
\label{yt}
\end{equation}

\section{Experiments and Analysis}
\label{exp}
{\bf{Datasets:}} The proposed method is evaluated on four real-world visual classification datasets: Office31, OfficeHome, DomainNet126 and DomainNet. Office31 contains three domains with 31 shared categories, denoted as A (Amazon), D (DSLR), and W (Webcam). OfficeHome consists of four domains with 65 shared categories, denoted as A (Art), C (Clipart), P (Product) and R (RealWorld). DomainNet126 includes four domains with 126 shared categories, including C (Clipart), P (Painting), R (Real), and S (Sketch). DomainNet has six domains with 345 shared categories, denoted as C (Clipart), I (Infograph), P (Painting), Q (Quickdraw), R (Real), and S (Sketch).

{\bf{Parameters:}} For a fair comparison, the target model adopts ResNet50 as its backbone, while CLIP with a vision transformer backbone is used as a frozen foundation model to generate target predictions that guide model adaptation under self-supervision. SGD optimizer with momentum $0.9$ is used to update the parameters. The initial learning rate $\eta = 0.001$, it changes linearly with training epochs as {$\eta = \frac{\eta_0}{(1+10p)^{0.75}}$} for all datasets. The batchsize is $64$. The trade-off parameters are set as $\beta = 0.003$, $\lambda = 0.5$, $\gamma = 0.5$, $\theta = 0.4$, $\delta = 1.0$ and $\sigma = \frac{1}{2p}$. For average radius, $\alpha = 1$.
$s_{adj}$ is defined as the median distance between source samples and their corresponding cluster centers within the same class, $t_{adj}$ is set to one-third of the source radius for each cluster.

All experiments are carried out using Pytorch 1.12.1+cu113 on a NVIDIA RTX-A5500 GPU.

{\bf{Baselines:}} Baselines used in this work include methods based on ResNet:
CAiDA \citep{330dong2021confident}(NeurIPS), 
FixBi \citep{658na2021fixbi}(CVPR), 
SSD \citep{457li2023multidomain}(TCYB),
Co-MDA \citep{483liu2023co}(TCSVT),
DCL \citep{622tian2023dcl}(TCSVT),
GSDE \citep{657westfechtel2024gradual}(WACV), 
SEAL \citep{585xia2024separation}(AAAI),
MPA \citep{568chen2024multi}(NeurIPS), 
KGCDE \citep{561wong2024graph}(PR), 
DSACDIC \citep{659zhao2024deep}(WACV)
LCFD \citep{574tang2024unified}(ArXiv), 
DAMP \citep{656du2024domain}(CVPR),
Ucon-SFDA \citep{655xu2024revisiting}(ICLR),
FuzHDA \citep{628li2025fuzzy}(TFS)
ProDe \citep{653tang2024proxy}(ICLR)
TIGM \citep{660zhu2025revisiting}(CVPR); 
and methods based on vision transformer:
DeiT \citep{661touvron2021training}(ICML),
CDTrans \citep{664xu2021cdtrans}(ICLR),
SSRT \citep{663sun2022safe}(CVPR),
DSiT \citep{567sanyal2023domain}(ICCV).

In the proposed method AutoS, when ResNet is used as the backbone, it is fully fine-tuned, whereas when a Vision transformer is used, the backbone is frozen. ``AutoS/sf" denotes AutoS under the source-free setting, while ``AutoS*" denotes AutoS based on a frozen Vision Transformer backbone.

{\bf{Results:}} Tables \ref{office} and \ref{domainnet} show the transfer performance on four datasets. The proposed AutoS consistently outperforms most baselines. On Office31 and DomainNet126, AutoS trained with source data surpasses its source-free model AutoS/sf, whereas on OfficeHome the source-free setting performs better.
Compared with multi-domain feature-alignment methods such as SSD and MPA, AutoS achieves higher accuracy while using fewer source knowledge. Against self-supervision approaches, such as CAiDA, Ucon-SFDA and methods that leverage large language models, like SEAL and LCFD, AutoS benefits from a target-confident sample selection strategy that effectively separates high- and low-confidence target samples for adaptation. Relative to federated-learning baselines Co-MDA and FuzHDA, AutoS still delivers superior performance on most tasks while relying only on selected source knowledge.

\begin{table}[htbp]
  \centering
  \caption{Results of AutoS on datasets Office31 and OfficeHome.}
  \setlength{\tabcolsep}{1.5mm}{
    \begin{tabular}{ccccc|cccccc}
    \toprule
    Method & D & W & A & Avg & Method & R & P & C & A & Avg \\
    \midrule
    CAiDA & \textbf{99.8}  & 98.9  & 75.8  & 91.5  & CAiDA & 84.2  & 84.7  & 60.5  & 75.2  & 76.2 \\
    SSD   & \textbf{99.8}  & \textbf{99.1}  & 76.0  & 91.6  & SSD   & 83.2  & 81.2  & 64.5  & 72.5  & 75.4 \\
    Co-MDA & 96.3  & 95.3  & 75.3  & 89.0  & Co-MDA & 83.9  & 85.3  & 64.0  & 74.4  & 76.9 \\
    GSDE  & 97.8  & 97.8 & 78.8 &  91.7   & GSDE  & 82.1  & 81.6  & 59.2  & 71.6  & 73.6 \\
    FixBi & 97.5 & 97.7 & 79.1 &  91.4    & MPA   & 85.7  & 86.2  & 54.9  & 74.8  & 75.4 \\
    SEAL  & 97.2  & 92.2  & 77.1  & 88.8  & SEAL  & 84.3  & 82.7  & 57.8  & 73.4  & 74.6 \\
    KGCDE & 99.8  & 99.2  & 79.4  & 92.8  & KGCDE & 85.4  & 83.7  & 68.3  & 75.6  & 78.3 \\
    DCL  &  97.3 &  95.3 & 77.2  &  89.9 & DAMP & 86.9  & 89.1  & 60.1  & 76.6  & 78.2 \\
    DSACDIC  & 95.9  & 96.8  & 75.9  &   89.5  & TIGM & 82.5  & 82.6  & 60.7  & 70.7  & 74.1 \\
    LCFD  &  93.4 &  92.5 & 82.8  &  89.7 & LCFD  & 89.7  & \textbf{90.2} & 72.2  & 80.7  & 83.2 \\
    Ucon-SFDA & 97.4  & 97.1  & 77.1  & 90.5 & Ucon-SFDA & 80.9  & 82.3  & 61.9  & 69.5 & 73.6 \\
    FuzHDA & 99.2  & 97.0  & 83.5  & \textbf{93.2} & FuzHDA & 87.9  & 89.9  & 71.4  & 81.2  & 82.6 \\
    ProDe & 96.5  & 93.9  & 79.4  & 89.9 & ProDe & 89.0  & 89.7  & 64.9  & 80.6 & 81.1 \\
    \midrule
    AutoS & 97.0  & 97.0  & \textbf{83.7} & 92.6  & AutoS & 88.7  & 90.0  & 72.2  & 80.7  & 82.9 \\
    AutoS/sf & 96.6  & 94.6  & 82.8  & 91.3  & AutoS/sf & \textbf{89.7} & 89.7  & \textbf{72.4} & \textbf{80.8} & \textbf{83.2} \\
    \bottomrule
    \bottomrule
    \end{tabular}}%
  \label{office}%
\end{table}%

\begin{table}[htbp]
  \centering
  \caption{Results of AutoS on datasets DomainNet126 and DomainNet.}
  \setlength{\tabcolsep}{1.2mm}{
    \begin{tabular}{cccccc|cccccccc}
    \toprule
    Method & C & P & R & S & Avg & Method & C & I & P & Q & R & S & Avg \\
    \midrule
    GSDE  & 83.4  & 77.8  & \textbf{91.0}  & \textbf{79.9}  & 83.1  & CAiDA   & 63.6  & 20.7  & 54.3  & 19.3   & 71.2  & 51.6  & 46.8 \\
    DAMP  &  74.5 &  76.2  & 88.4  & 71.0  & 77.5  & SSD   & \textbf{67.2} & 21.7  & 52.4  & 20.8  & 67.8  & 55.3  & 47.5 \\
    LCFD  & 79.3  & 77.5  & 88.1  & 75.3  & 80.1 & DSiT* & 55.3  & 23.4  & 47.1  & 17.6  & 63.5  & 45.5  & 42.1 \\
    Ucon-SFDA & 72.2  & 69.6  & 81.0  & 65.4  & 72.1 & DeiT* & 41.4  & 39.9  & 38.9  & 19.6  & 39.0  & 41.5  & 36.7 \\
    FuzHDA & 82.4  & 80.0  & 88.7  & 76.9  & 82.0 & SSRT* & 49.8  & 46.3  & 45.0  & 29.3  & 48.8  & 52.1  & 45.2 \\
    TIGM  & 74.7  & 72.1  & 86.2  & 67.7  & 75.2 & CDTrans* & 48.7  & \textbf{48.4}  & 46.4  & \textbf{30.7} & 45.6  & 51.5  & 45.2 \\
    \midrule
    AutoS & \textbf{83.7} & \textbf{81.0} & 90.0 & 78.1 & \textbf{83.2} & AutoS & 64.1  & 22.5  & 50.2  & 8.2   & 61.1  & 47.5  & 42.3 \\
    AutoS/sf & 82.1  & 79.7  & 89.2  & 75.0  & 81.5  & AutoS* & 64.6  & 32.7 & \textbf{58.1} & 6.2   & \textbf{69.4} & \textbf{56.5}  & \textbf{47.9} \\
    \bottomrule
    \bottomrule
    \end{tabular}}%
  \label{domainnet}%
\end{table}%

\begin{table}[htbp]
  \centering
  \caption{Ablation study of AutoS on OfficeHome.}
    \begin{tabular}{cccccc}
    \toprule
    Method & R     & P     & C     & A     & Avg \\
    \midrule
    Fedavg & 88.6  & 89.4  & 71.8  & 80.4  & 82.6 \\
    w/o TarCof & \textbf{88.9} & 89.6  & 72.1  & 80.5  & 82.8 \\
    w/o $\mathcal{L}$ & 88.1  & 89.4  & 70.1  & 79.1  & 81.7 \\
    w/o $\mathcal{L}_{in}$ & 76.5  & 73.5  & 51.9  & 64.6  & 66.6 \\
    w/o $\mathcal{L}_{ex}$   & 88.2  & 89.3  & 70.1  & 80.2  & 82.0 \\
    AutoS & 88.7  & \textbf{90.0} & \textbf{72.2} & \textbf{80.7} & \textbf{82.9} \\
    \bottomrule
    \bottomrule
    \end{tabular}%
  \label{ablation}%
\end{table}%

\begin{figure}[htbp]
\centering
\subfloat[Size of selected source data]{
\begin{minipage}[t]{0.4\linewidth}
\centering
\includegraphics[height=3cm,width=3cm]{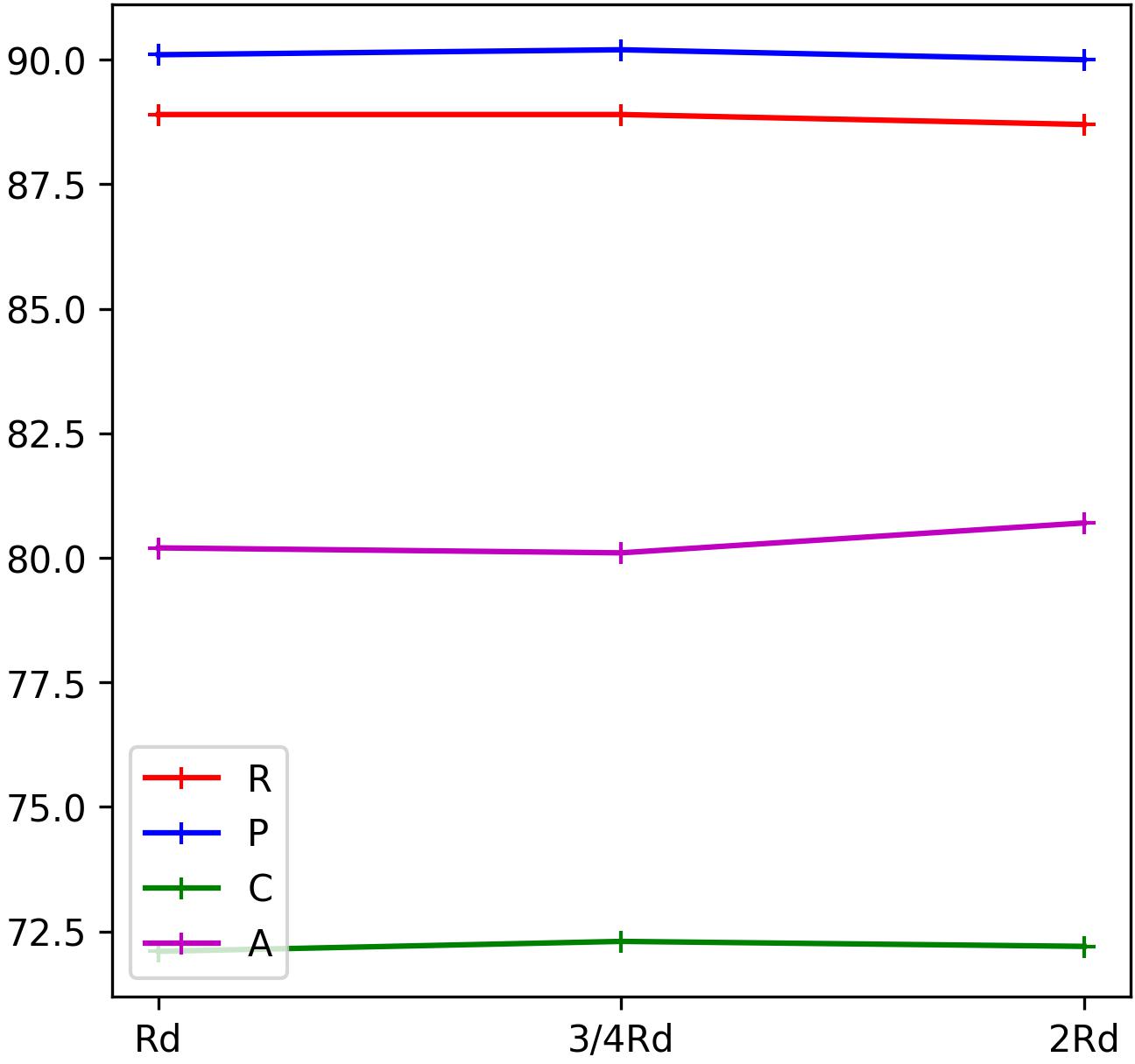}
\end{minipage}
}
\subfloat[Size of selected target data]{
\begin{minipage}[t]{0.4\linewidth}
\centering
\includegraphics[height=3cm,width=3cm]{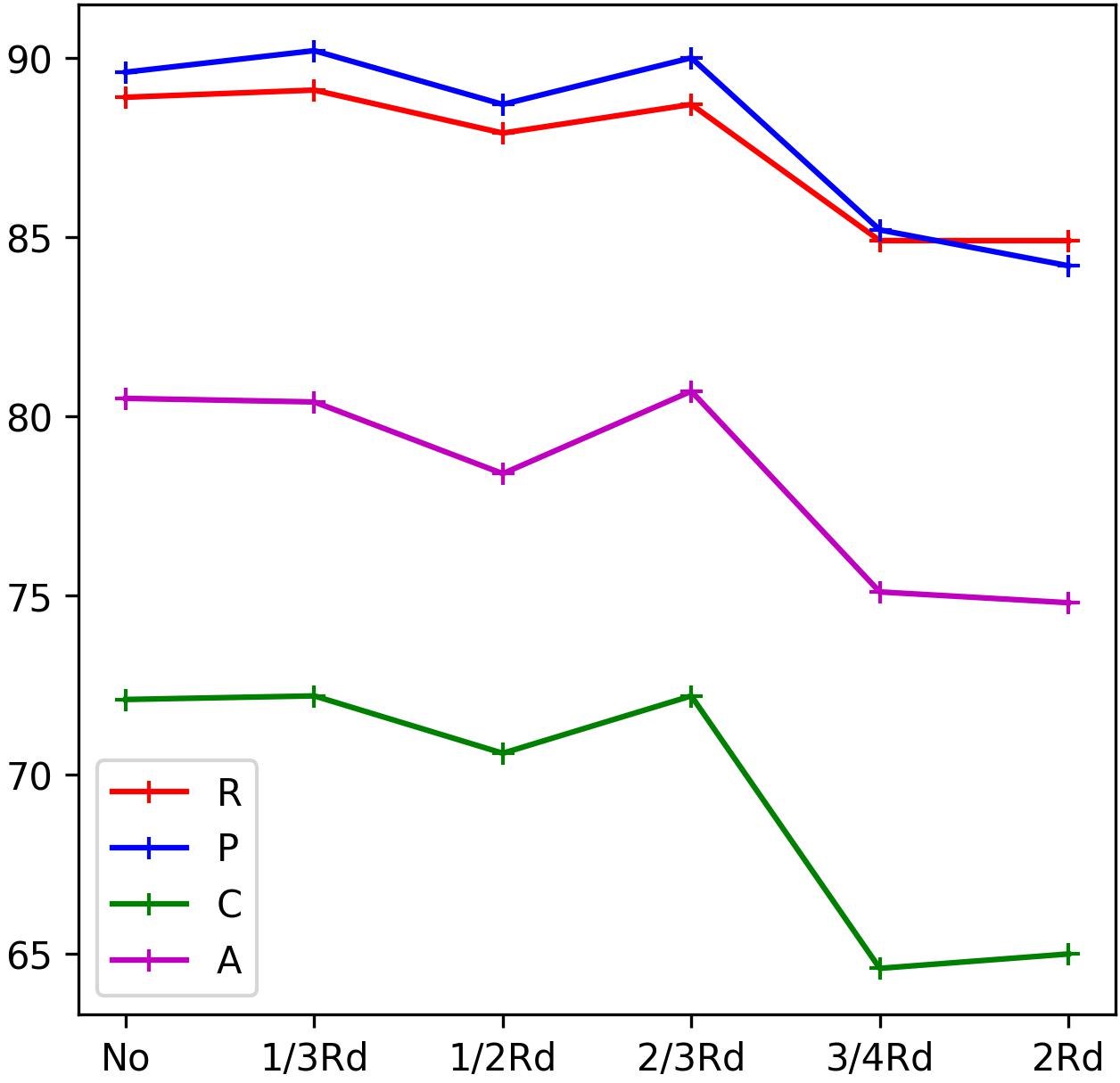}
\end{minipage}
}
\caption{Transfer performance with different size of selected source and target data.}\label{datasize}
\end{figure}

{\bf{Ablation Study:}} Table \ref{ablation} shows the performance of AutoS on OfficeHome using an ablation strategy. ``FedAvg" refers to a model trained without source-knowledge selection, combining domains through simple federated averaging. ``w/o TarCof" indicates training without adding target samples to the selected source domains.
AutoS achieves the best performance. The model with no source-target interaction ($\mathcal{L}_{in}$) gives the lowest accuracy, indicating the need to bridge domain gaps. The next most critical factor is source supervision 
($\mathcal{L}$), showing the value of supervision from source labels. Target pseudo-labels, enhanced by the frozen foundation model and fine-tuned prompts ($\mathcal{L}_{ex}$), provide the third-highest gain.

Figure \ref{datasize} shows the influence of the size of selected source and target confident samples on OfficeHome. ``Rd" denotes the average radius of the source clusters. For the source data, removing samples whose distance to their cluster centers exceeds the mean radius does not affect transfer performance, indicating that the source domains contain redundant knowledge. For the target data, the best transfer performance is achieved when using confident target samples whose distance to the mean source radius is less than one third of that radius.

\begin{table}[htbp]
  \centering
  \caption{Performance of AutoS with different radius metric.}
    \begin{tabular}{cccccc|ccccc}
    \toprule
    \multicolumn{6}{c|}{ACCURACY}                 & \multicolumn{5}{c}{TIME} \\
    \midrule
    Method & R & P & C & A & Avg & R & P & C & A & Avg \\
    \midrule
    RMS   & 88.7  & \textbf{90.1} & 69.9  & 78.0  & 81.7  & \textbf{7718.3} & 9205.8 & 7033.1 & 4867.5 & 7206.2 \\
    MAX   & \textbf{88.8} & 89.7  & 72.0  & 79.8  & 82.6  & 8314.4 & 8289.1 & 8299.8 & 5141.1 & 7511.1 \\
    MEAN  & 88.7  & 90.0  & \textbf{72.2} & \textbf{80.7} & \textbf{82.9} & 9130.3 & \textbf{6998.4} & \textbf{6492.5} & \textbf{4351.5} & \textbf{6743.2} \\
    \bottomrule
    \bottomrule
    \end{tabular}%
  \label{radius}%
\end{table}%

\begin{figure}[htbp]
\centering
\subfloat[Average radius]{
\begin{minipage}[t]{0.3\linewidth}
\centering
\includegraphics[height=2.5cm,width=2.5cm]{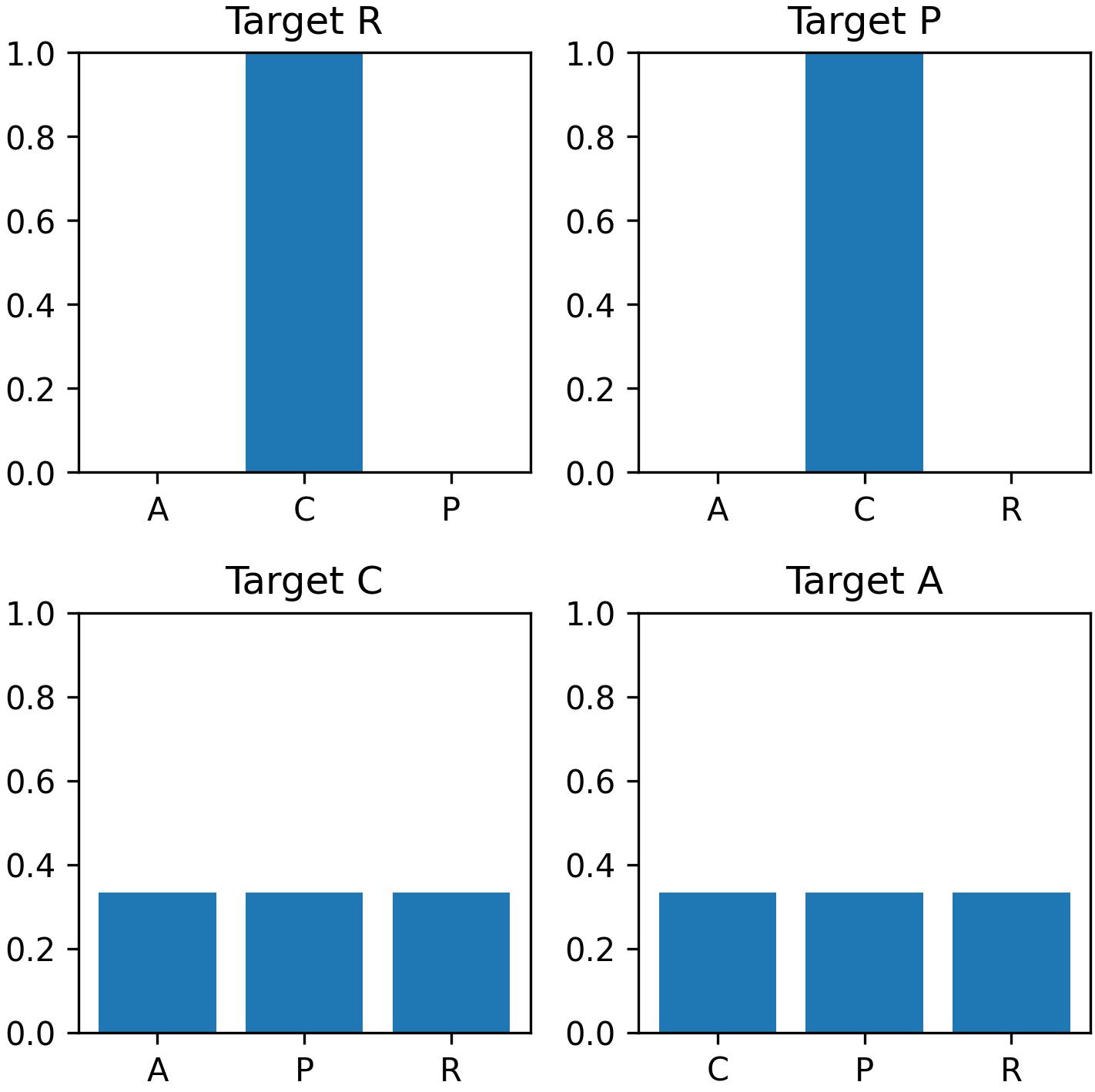}
\end{minipage}
}
\subfloat[Root mean square radius]{
\begin{minipage}[t]{0.3\linewidth}
\centering
\includegraphics[height=2.5cm,width=2.5cm]{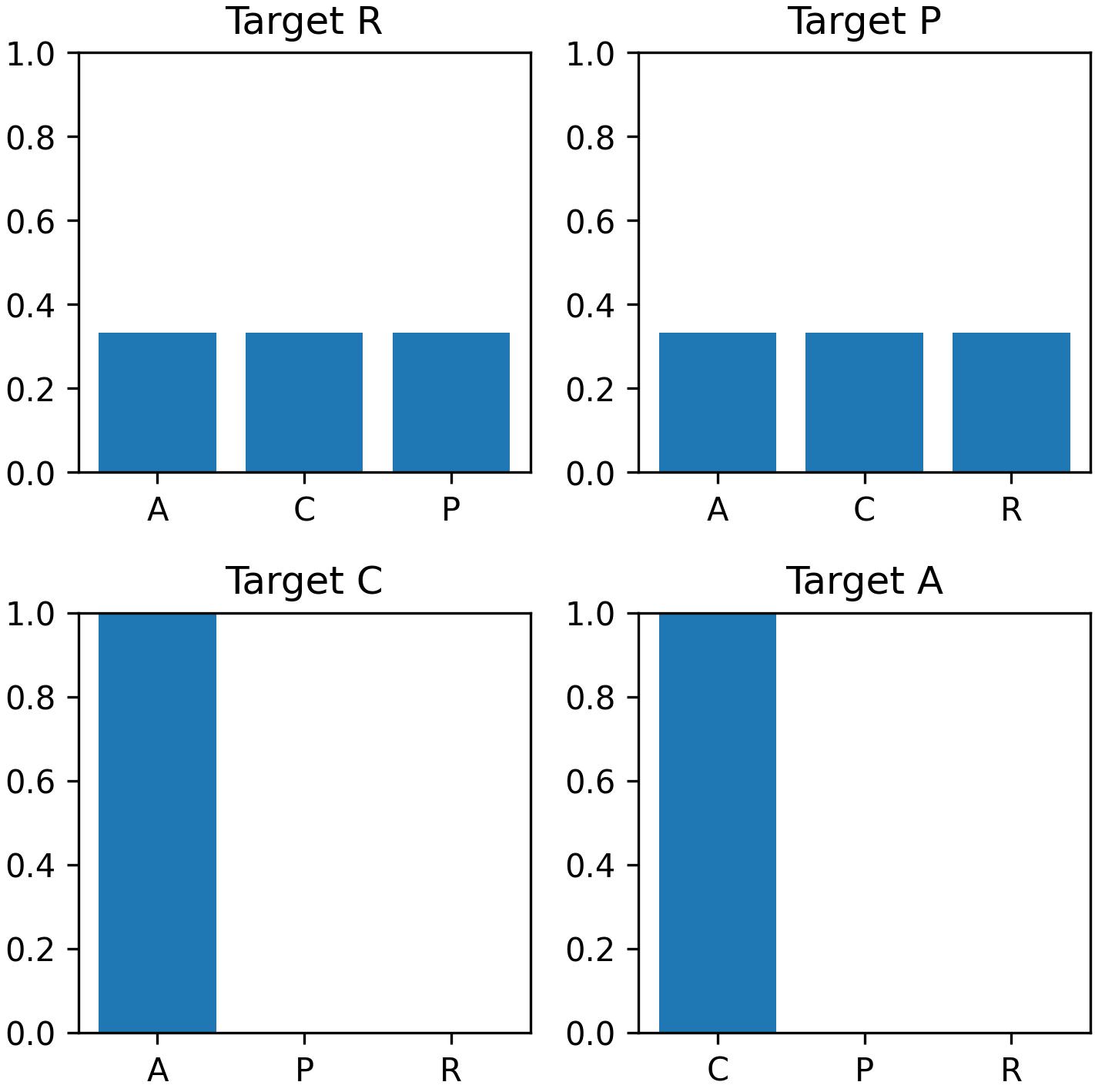}
\end{minipage}
}
\subfloat[Maxium radius]{
\begin{minipage}[t]{0.3\linewidth}
\centering
\includegraphics[height=2.5cm,width=2.5cm]{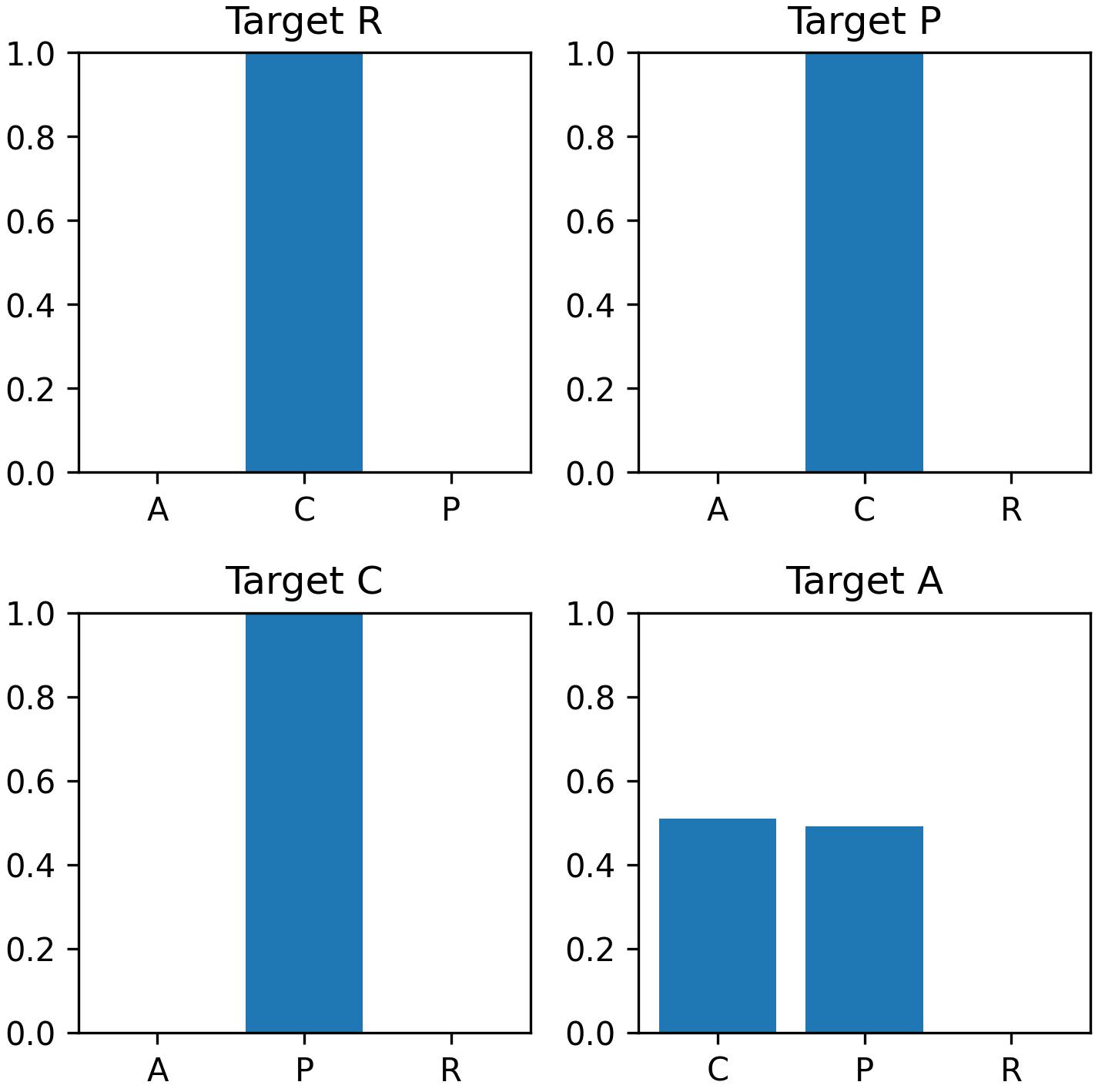}
\end{minipage}
}
\caption{Transfer performance with different distance metrics for defining radius of clusters, taking OfficeHome as example.}\label{seloff}
\end{figure}

\begin{figure}[htbp]
\centering
\subfloat[Sected by ResNet]{
\begin{minipage}[t]{0.45\linewidth}
\centering
\includegraphics[height=3cm,width=6cm]{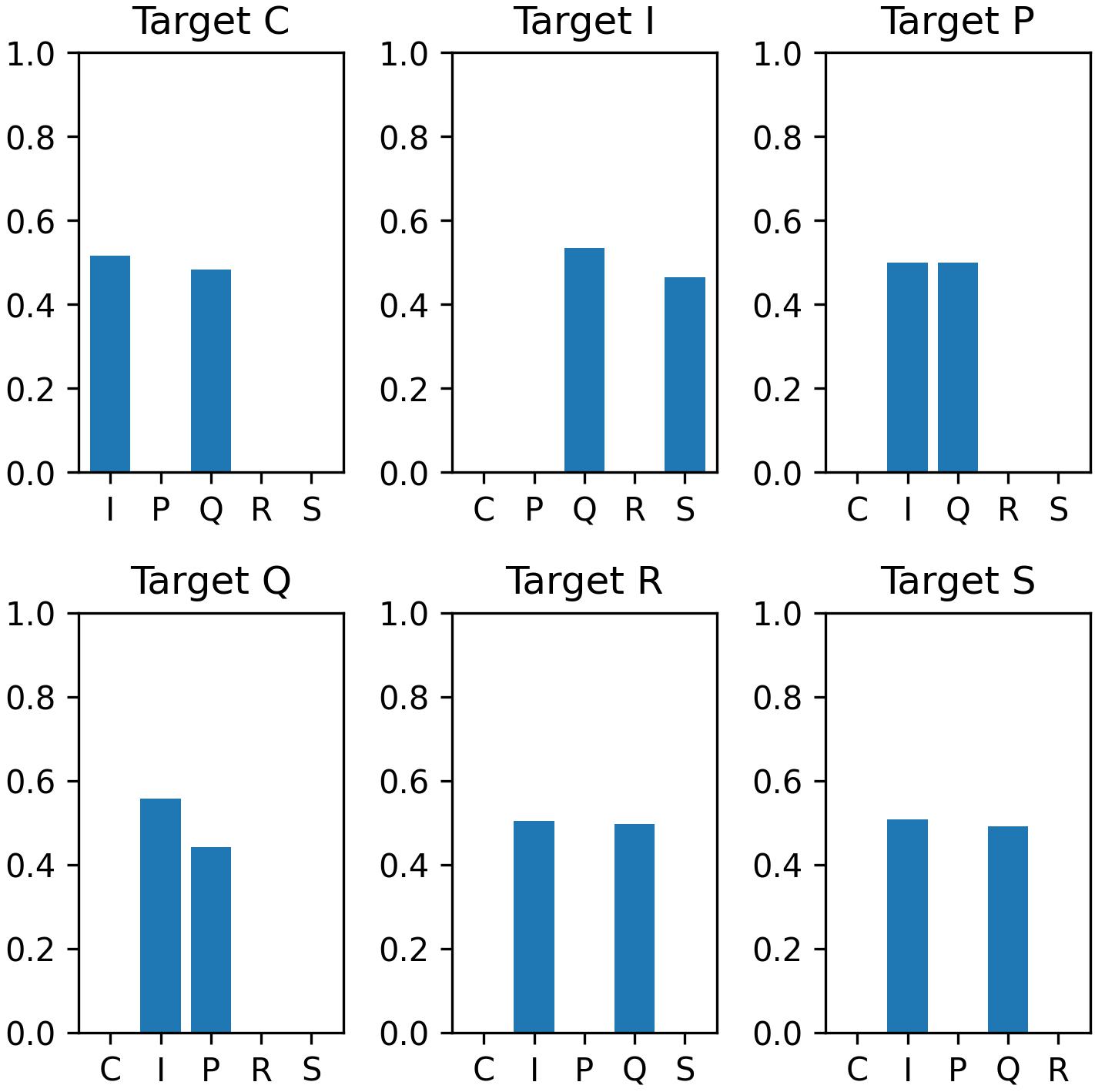}
\end{minipage}
}
\subfloat[Selected by ViT]{
\begin{minipage}[t]{0.45\linewidth}
\centering
\includegraphics[height=3cm,width=6cm]{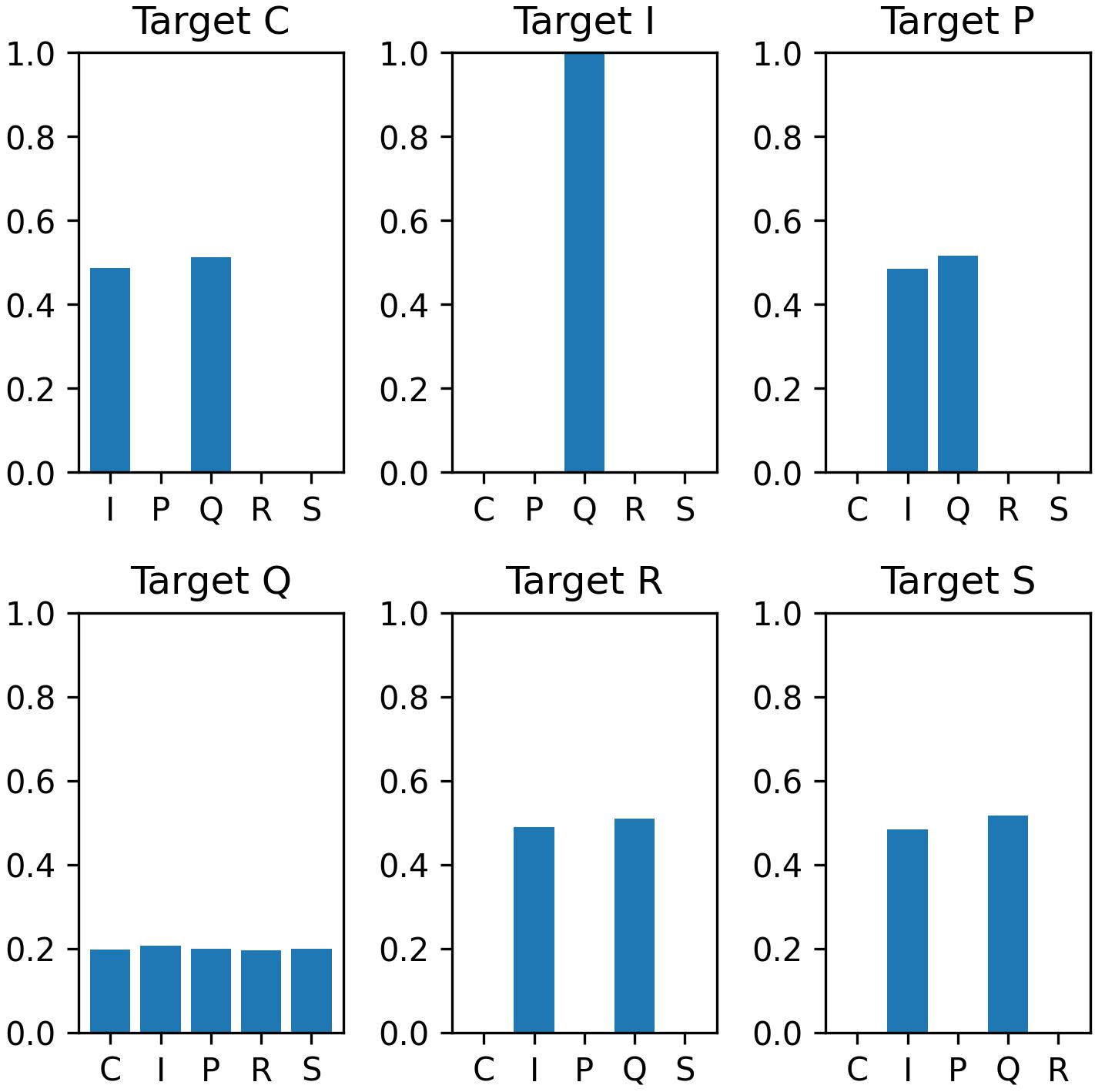}
\end{minipage}
}
\caption{Autonomously select source domains in DomainNet.}\label{seldnet}
\end{figure}

Table \ref{radius} reports the transfer performance and computational time of AutoS with different radius metrics. ``RMS" is the root mean square distance, ``MAX" is the maximum radius, and ``MEAN" is the average radius. The results show that the average radius obtains the best performance while requiring the least running time. Figures \ref{seloff} and \ref{seldnet} show the autonomously selected source domains using density-controlled strategy. For most tasks, the proposed AutoS achieves better performance than multi-domain adaptation baselines that use all sources, while requiring only one or two source domains.

\begin{table}[htbp]
  \centering
  \caption{Running time and GPU memory usage.}
    \begin{tabular}{ccccccc}
    \toprule
    Method & R     & P     & C     & A     & Avg & GPU \\
    \midrule
    SSD   & 62476.2 & 41824.8 & 42081.2 & 42190.0 & 47143.0 & 12415.0 \\
    FuzHDA & \textbf{7314.0} & 7243.1 & 7323.8 & 7467.3 & 7337.0 & 8195.0 \\
    AutoS & 9130.3 & \textbf{6998.4} & \textbf{6492.5} & \textbf{4351.5} & \textbf{6743.2} & \textbf{8157.0} \\
    \bottomrule
    \bottomrule
    \end{tabular}%
  \label{time}%
\end{table}%

Table \ref{time} reports the running time and GPU memory usage of the proposed AutoS compared with two multi-domain baselines, SSD and FuzHDA. The reported running time includes both source and target model training. It shows that AutoS requires less time and lower memory than the baselines.

\begin{figure}[htbp]
\centering
\subfloat[Selected A and R]{
\begin{minipage}[t]{0.3\linewidth}
\centering
\includegraphics[height=2.5cm,width=2.5cm]{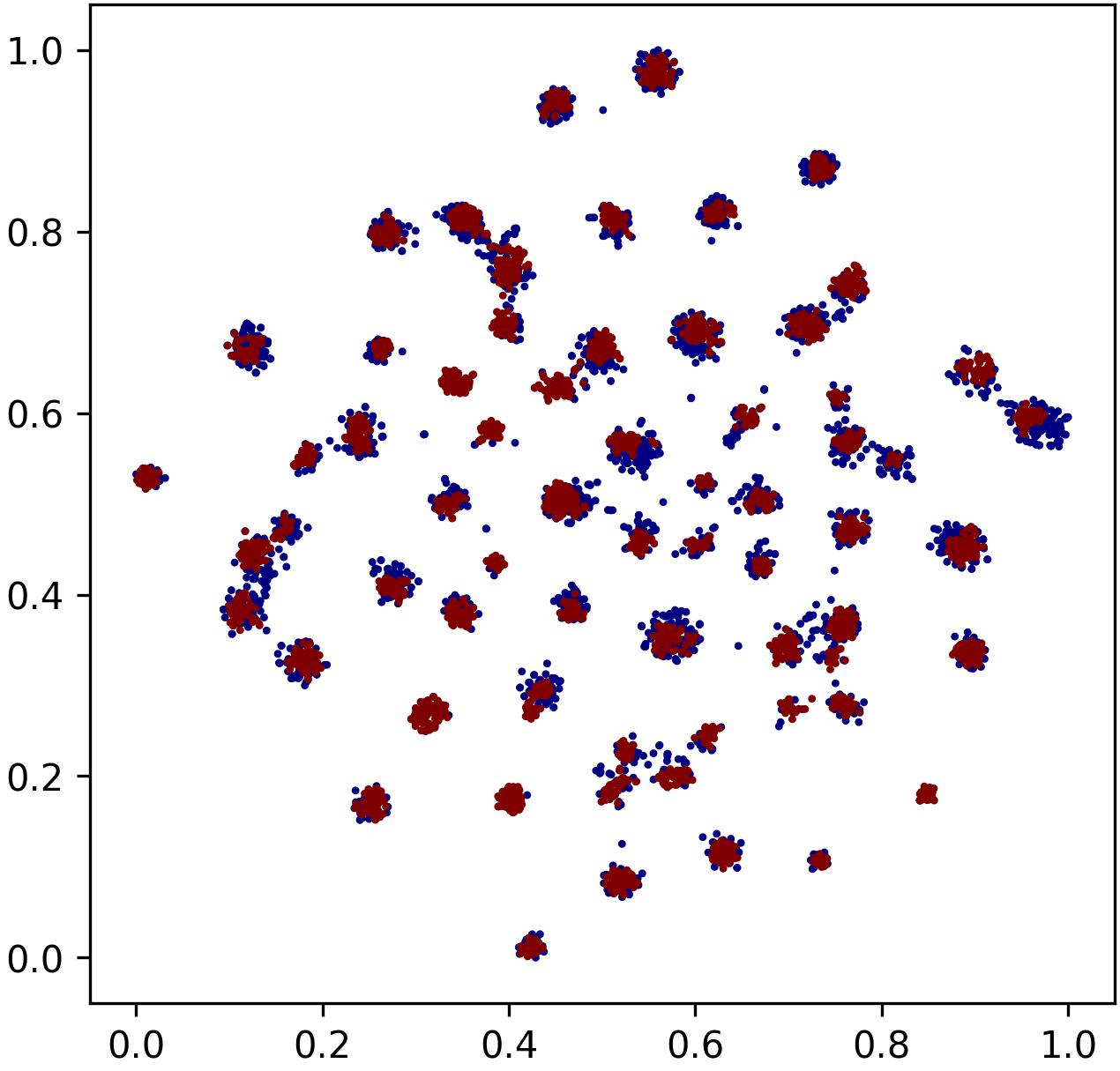}
\end{minipage}
}
\subfloat[Selected C and R]{
\begin{minipage}[t]{0.3\linewidth}
\centering
\includegraphics[height=2.5cm,width=2.5cm]{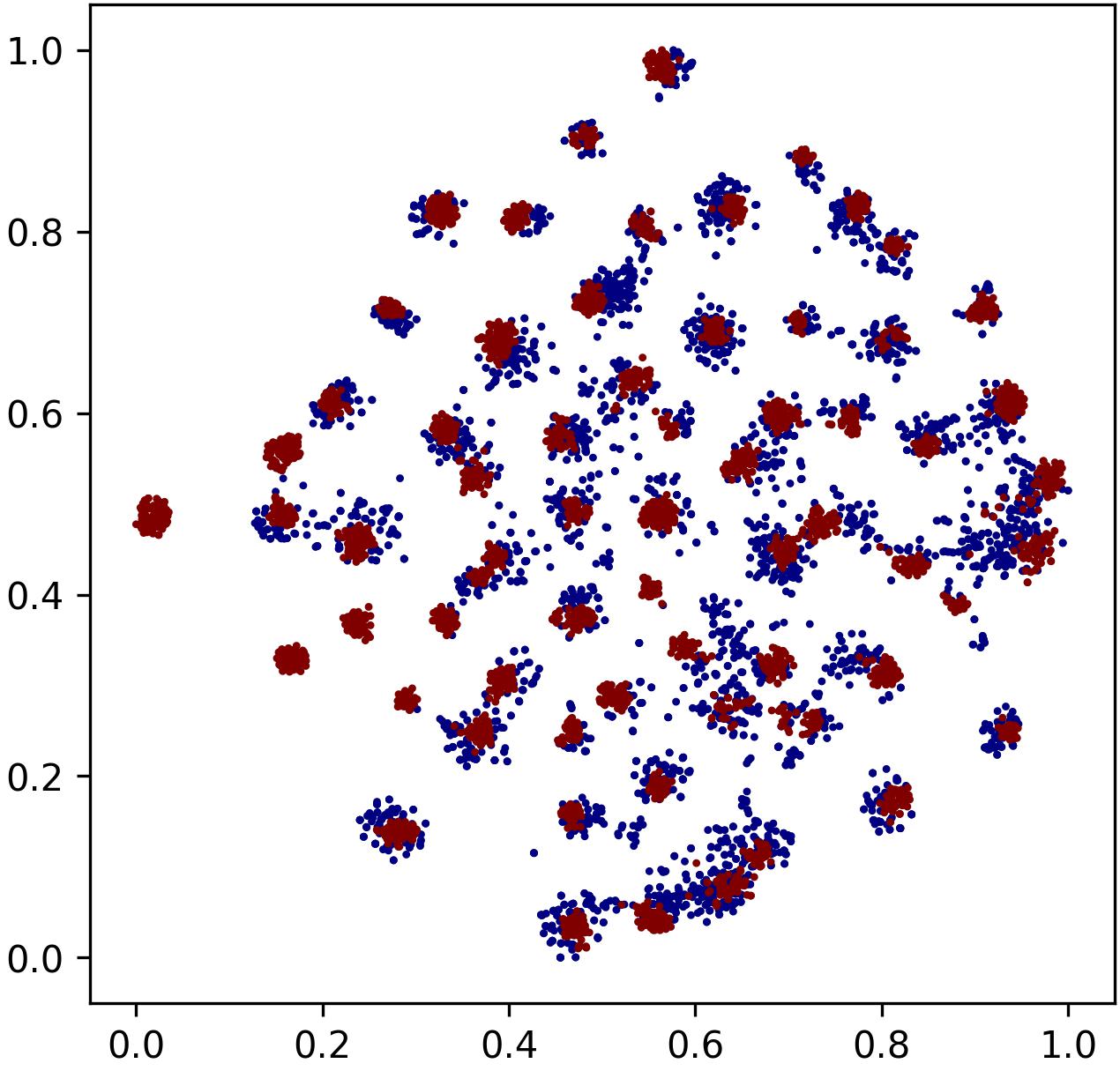}
\end{minipage}
}
\subfloat[Selected P and R]{
\begin{minipage}[t]{0.3\linewidth}
\centering
\includegraphics[height=2.5cm,width=2.5cm]{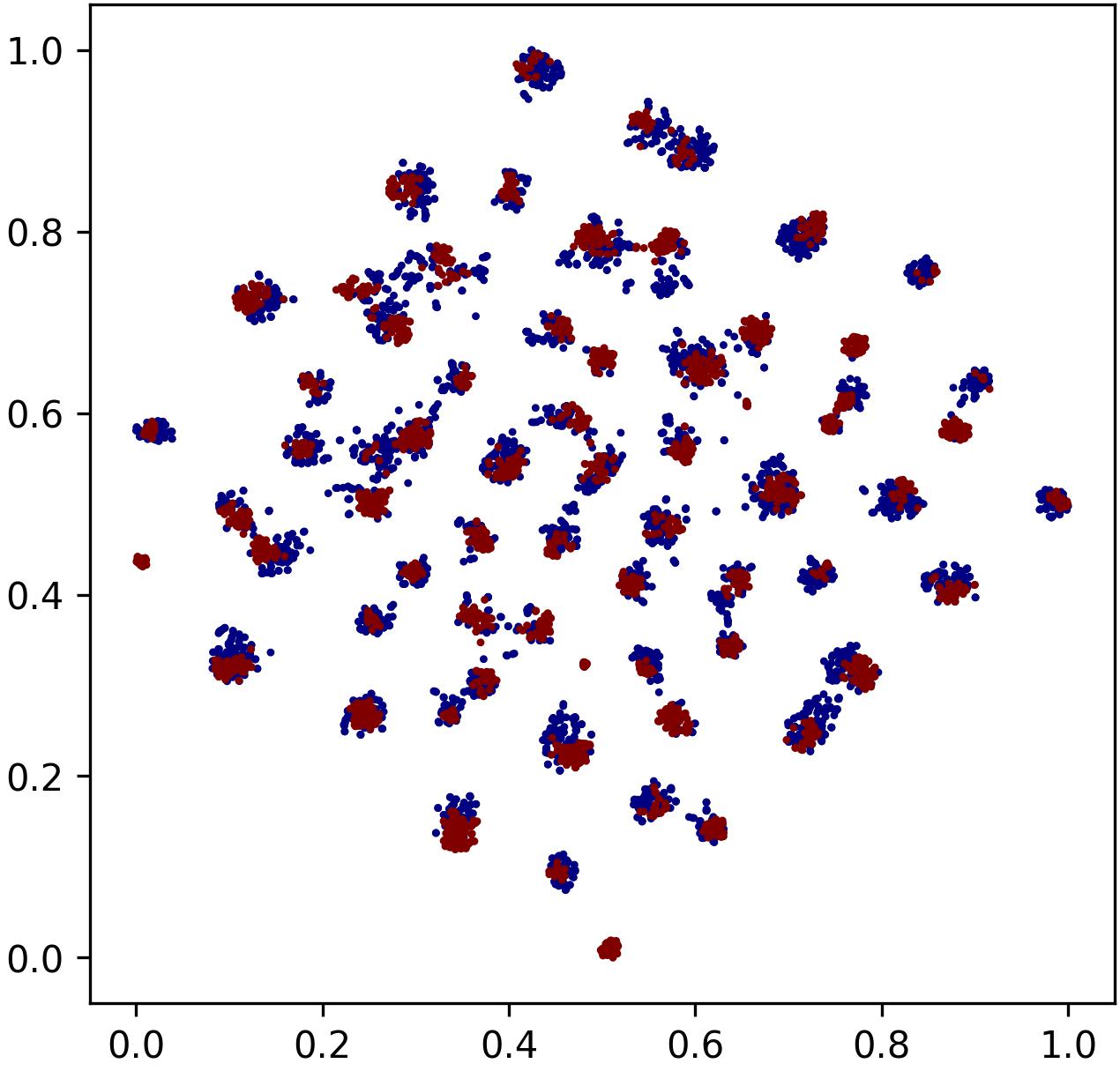}
\end{minipage}
}

\subfloat[Adapted A and R]{
\begin{minipage}[t]{0.3\linewidth}
\centering
\includegraphics[height=2.5cm,width=2.5cm]{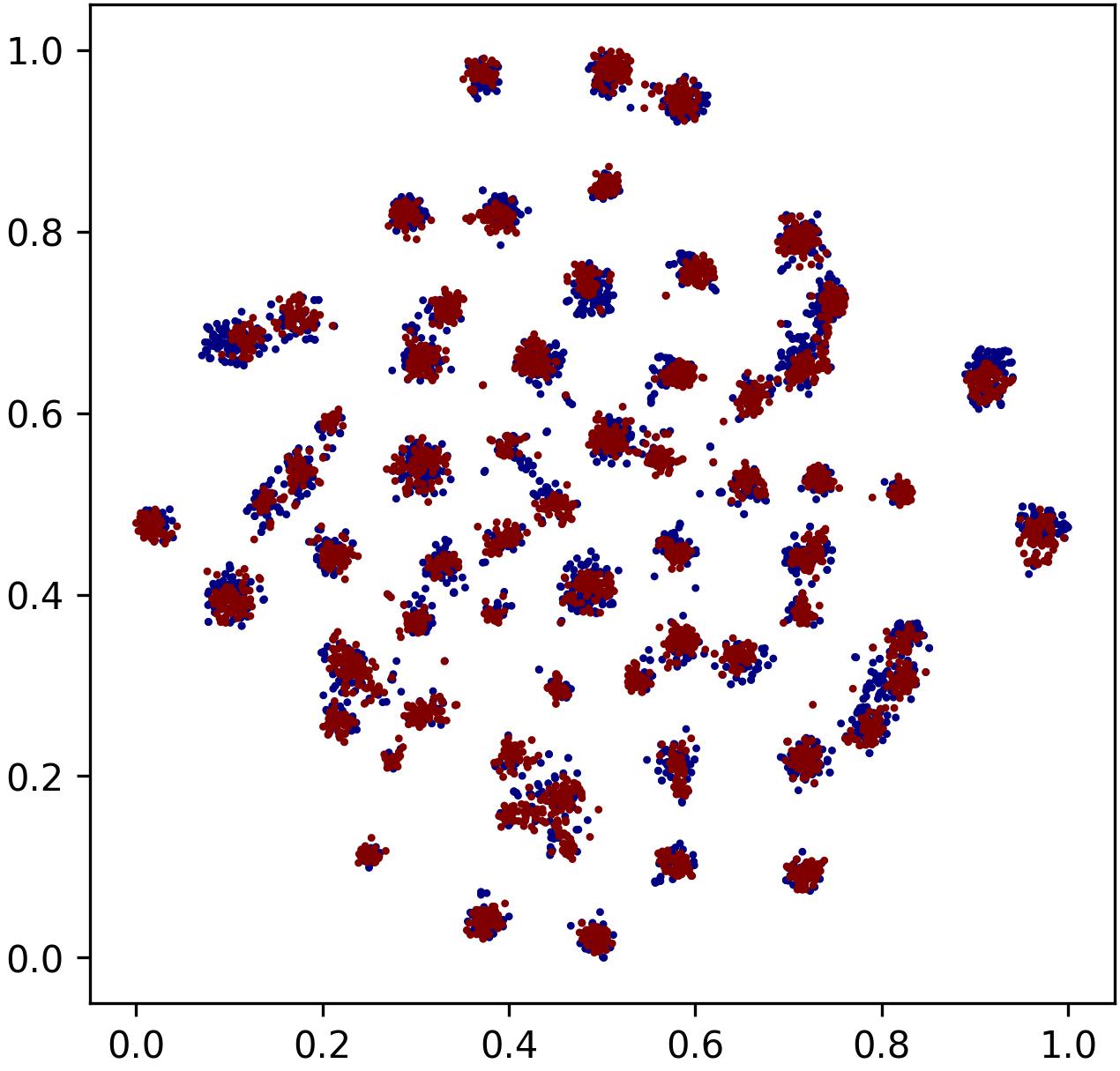}
\end{minipage}
}
\subfloat[Adapted C and R]{
\begin{minipage}[t]{0.3\linewidth}
\centering
\includegraphics[height=2.5cm,width=2.5cm]{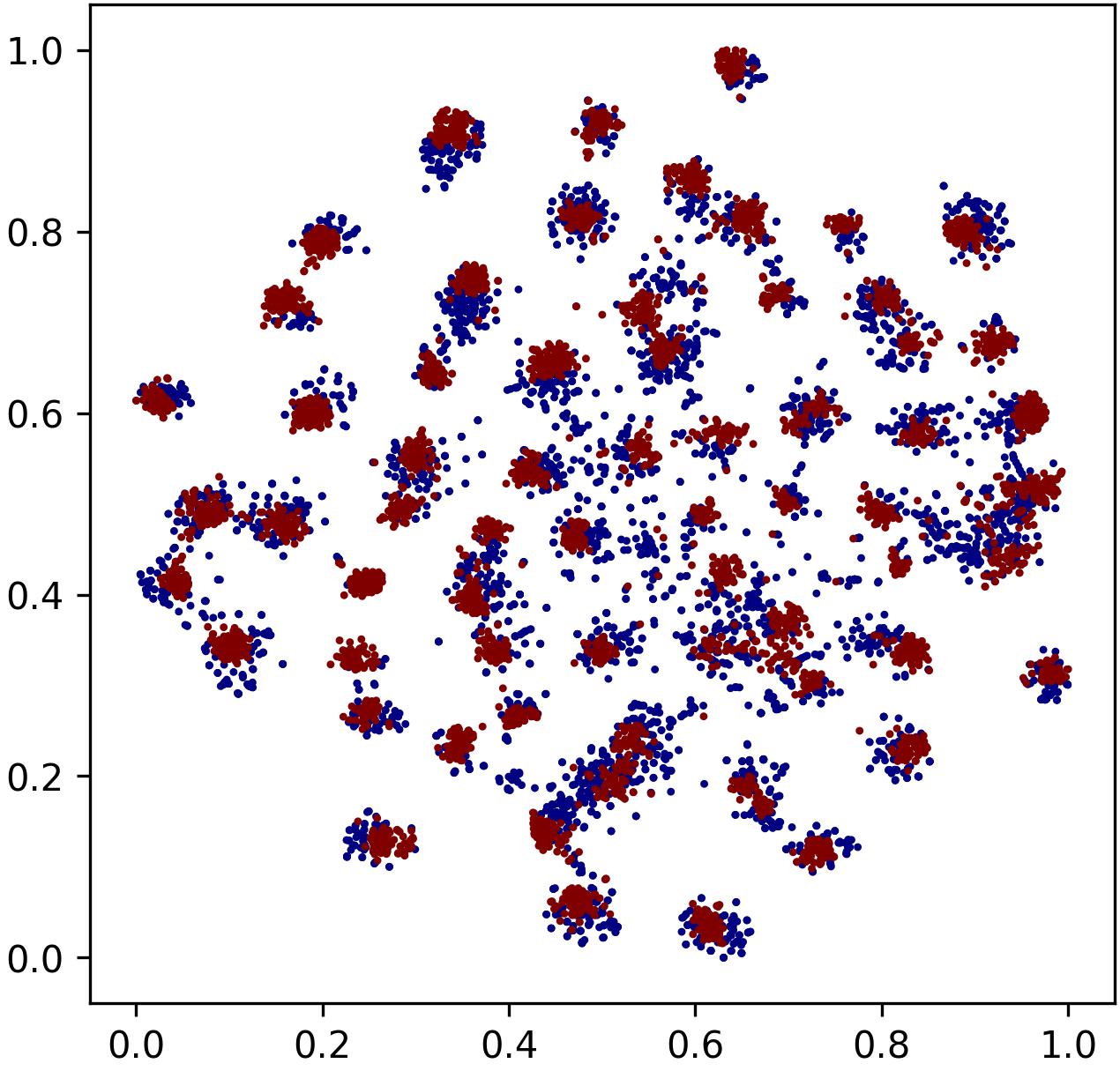}
\end{minipage}
}
\subfloat[Adapted P and R]{
\begin{minipage}[t]{0.3\linewidth}
\centering
\includegraphics[height=2.5cm,width=2.5cm]{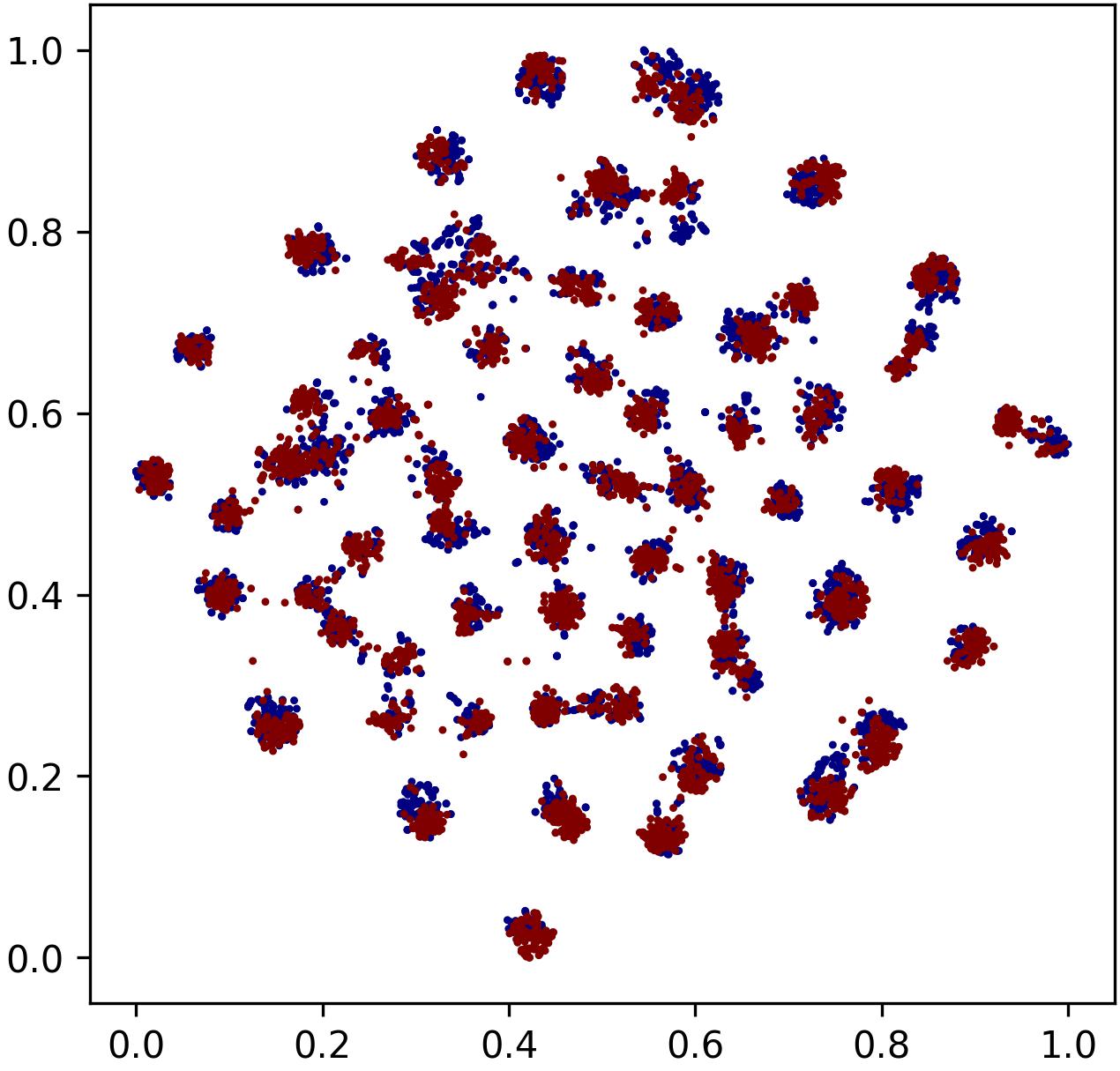}
\end{minipage}
}
\caption{T-SNE of adapted source and target data on task R from OfficeHome.}\label{tsne}
\end{figure}

{\bf{Visualization:}} Figure \ref{tsne} illustrates the distributions of the selected source and target samples, along with the adaptation of the complete source and target domains using task R from OfficeHome as an example. It shows that the three source domains align well with the target domain. The reason is that AutoS autonomously selects source domains during training, the target model initially learns from all sources but later removes those defined as dissimilar. This autonomous selection strategy allows the method to emphasize the most relevant source domains at the appropriate time while still preserving transferable knowledge from all sources.

\section{Conclusion and Future Work}\label{future}
This work introduces an autonomous source knowledge selection method that dynamically identifies relevant and irrelevant source domains with respect to the target domain during training. First, the method applies a federated learning strategy to leverage transferable knowledge from multiple sources and designs a density-controlled selection strategy to determine relevant domains while assigning combination weights for federated aggregation of source models in adapting to the target. Furthermore, rather than relying on data matching, the method adapts the target domain through self-supervision using pseudo-labels enhanced by a frozen foundation model, while fine-tuning only the prompts. This design makes the approach flexible for domain adaptation both with and without access to source data.

In futuer study, we will extend the proposed method to deal with more complex settings such as label shifts and modality shifts 


\subsubsection*{LLMs Usage}
The authors used ChatGPT for grammar and spelling checks only, with prompt "Proofread the sentences".

\newpage
\bibliography{TL-Reference-lkqy}
\bibliographystyle{iclr2026_conference}

\newpage
\appendix
\section{Appendix}
\subsection{Proof of the objective function for structural causal model}
Given target predictions provided by source federal model and frozen foundation model as $\mathcal{P}^t$ and $\mathcal{P}_{FM}$ respectively, latent values of text prompts $\{\mathcal{T}_c\}_{c=1}^{\mathcal{C}}$ as $V$, latent random values following distribution of $\mathcal{P}^t$ and $\mathcal{P}_{FM}$ as $\mathcal{P}_Y$ and $\mathcal{P}_V$, the intermediate bottleneck values connecting latent text prompts $V$ and $\mathcal{P}_{FM}$ as $\mathcal{P}'_V$.
To ensure the relationship between  target model and text prompts, we following previous works \citep{574tang2024unified, 628li2025fuzzy} to maximize the correlation between the latent random values of text prompts $V$ and target model predictions $\mathcal{P}_Y$, which is:

$\max \mathcal{I}(\mathcal{P}_Y, V) = \max \mathcal{I}(\mathcal{P}_Y, \mathcal{P}_V), 
\mathcal{P}_V = \Psi(V)_{txt}$.

Based on the statement about information loss under dimensional compression, for random values $Z$, $X$ and $Y$, there is a function $h(\cdot)$ satisfies that $X \in \mathcal{R}^m \overset{h(\cdot)} {\rightarrow} Y \in \mathcal{R}^{\mathcal{C}}$, if $\mathcal{C} < m$, then:

(1). There is $X' \neq X$ satisfying $h(X') = h(X)$;

(2) If $Z \rightarrow X  \rightarrow Y$ a Markov chain, then $\mathcal{I}(Z, Y) \leq \mathcal{I}(Z, X)$.

Let $Z = \mathcal{P}_Y$, $Y = \mathcal{P}_V$ and $X = V$, then we have 
\begin{equation}
\mathcal{I}(\mathcal{P}_Y, \mathcal{P}_V) \leq \mathcal{I}(\mathcal{P}_Y, V).   
\end{equation}

Simultaneously, for random values $Z$, $X$ and $Y$, there are intermediate value $Z'$ and intermediate value $Y'$ satisfying $Z' \overset{g(\cdot)} {\rightarrow} Y'$, and 
$Z \rightarrow Z' \rightarrow (Y, Y')$
while $g(\cdot)$ is reversible and uncompressed, then:

(1). $\mathcal{I}(Z,Y') = \mathcal{I}(Z,g(Z')) = \mathcal{I}(Z,Z')$  

(2). If $Z' \rightarrow Y' \rightarrow Y$ a Markov chain, then $\mathcal{I}(Y',Y) \leq \mathcal{I}(Z',Y)$.

Then we have:
\begin{equation}
\mathcal{I}(Z,Y) = \mathcal{I}(Z,Z') - \mathcal{I}(Z',Y) \leq
\mathcal{I}(Z,Z') -\mathcal{I}(Y', Y), Y' = g(Z').
\end{equation}

Let $Z' = \mathcal{P}'_v$, $Y' = \mathcal{P}'_Y = g(\mathcal{P}'_V)$, we have:

\begin{equation}
\begin{split}
\mathcal{I}(\mathcal{P}_Y,\mathcal{P}_V) &= \mathcal{I}(\mathcal{P}_Y,\mathcal{P}'_V) - \mathcal{I}(\mathcal{P}'_V,\mathcal{P}_V)\\
&\leq
\mathcal{I}(\mathcal{P}_Y,\mathcal{P}'_V) -\mathcal{I}(\mathcal{P}'_Y, \mathcal{P}_V) \\
&= 
\mathcal{I}(\mathcal{P}_Y,\mathcal{P}'_V) -\mathcal{I}(g(\mathcal{P}'_V), \mathcal{P}_V).
\end{split}
\end{equation}

The upper bound of $\mathcal{I}(\mathcal{P}_Y,\mathcal{P}'_V)$ is calculated as:
\begin{equation}
\begin{split}
\mathcal{I}(\mathcal{P}_Y,\mathcal{P}'_V)
&  \equiv H(\mathcal{P}'_V) - H(\mathcal{P}'_V| \mathcal{P}_Y) \equiv H(\mathcal{P}_Y) - H(\mathcal{P}_Y| \mathcal{P}'_V) \\
& = -\sum \mathbb{P}(\mathcal{P}_Y) \log\mathbb{P}(\mathcal{P}_Y)
-\sum \mathbb{P}(\mathcal{P}'_V,\mathcal{P}_Y)\log \frac{\mathbb{P}(\mathcal{P}'_V,\mathcal{P}_Y)}{\mathbb{P}(\mathcal{P}'_V)} \\
& = \mathbb{E}[\log\mathbb{P}(\mathcal{P}_Y)]
+ \mathbb{E}[\log \frac{\mathbb{P}(\mathcal{P}'_V,\mathcal{P}_Y)}{\mathbb{P}(\mathcal{P}'_V)}] \\
& = \mathbb{E}[\log\mathbb{P}(\mathcal{P}'_V)]
+ \mathbb{E}[\log \frac{\mathbb{P}(\mathcal{P}_Y,\mathcal{P}'_V)}{\mathbb{P}(\mathcal{P}_Y)}] \\
& \geq \text{const}
+ \mathbb{E}[\log \mathbb{P}(\mathcal{P}'_V|\mathcal{P}_Y))], \\
\end{split}
\end{equation}

Calculation of $\mathcal{I}(g(\mathcal{P}'_V),\mathcal{P}_V)$ is expressed as:
\begin{equation}
\begin{split}
\mathcal{I}(g(\mathcal{P}'_V),\mathcal{P}_V) &= \mathcal{KL}(g(\mathcal{P}'_V)\|\mathcal{P}_V), \\
g(\mathcal{P}'_V) &= \frac{1}{\text{diag}(\mathcal{P}_V)} \odot \mathcal{P}_V.
\end{split}  
\end{equation}

Then we can get Equation \ref{exloss} by modeling $\mathbb{P}(\mathcal{P}'_V|\mathcal{P}_Y)$ as Gaussian distribution $\mathbb{N}(\mathcal{P}_Y, \text{diag}(\mathcal{P}_Y))$ and updating the random values with $\mathcal{P}^t$ and $\mathcal{P}_{FM}$ during training.

\subsection{Algorithm of the proposed Autos}
The whole algorithm of the proposed AutoS is summarized as in Algorithm \ref{autos}:

\begin{algorithm}[htbp]
\caption{AutoS: Autonomous source knowledge selection.}\label{autos}
\begin{algorithmic}[1]
\STATE {\bf Input:} Source domains $\{\mathcal{D}^s_k = \{\bm x^s_{ki}, \bm y^s_{ki}\}_{i=1}^{n_k}\}_{k=1}^K$, target domain $\mathcal{D}^t = \{\bm x^t_i\}_{i=1}^n$;
\STATE {\bf Initialization:} Feature extraction module $\Phi$, classifier $P$, text prompts $\{\mathcal{T}_c\}_{c=1}^{\mathcal{C}}$;
\FOR {$\epsilon=1$, $\epsilon<\mathcal{I}$, $\epsilon++$, }
\STATE $\Phi_k$, $P_k$ $\leftarrow$ $\Phi$, $P$: Train source independent models $\Phi_k$, $P_k$ as in \eqref{Ms};
\STATE $\{\bm f_{kc}^s\}_{c=1}^{\mathcal{C}}$ $\leftarrow$ $P_k$: Collect the $k$th source cluster centers \eqref{fkc};
\STATE $\bm y_k^t$ $\leftarrow$ $\bm x^t$, $\bm f_{kc}^s$: Collect target clustering label as in \eqref{yclu};
\STATE $\bm r_{kc}^s$ $\leftarrow$ $\bm x^s_k$,$\bm f_{kc}^s$, $\bm r_{kc}^t$ $\leftarrow$ $\bm x^t$, $\bm f_{kc}^s$: Calculate cluster radius of source and target data as in \eqref{rkc};
\STATE $\bm d_{kc}^s$, $\bm d_{kc}^t$ $\leftarrow$ $\bm r_{kc}^s$: Define thresholds to identify confident source and target samples as in \eqref{dkc}; 
\STATE $\rho_{kc}^t$ $\leftarrow$ $\bm r_{kc}^t$: Define thresholds to identify confident source and target samples as in \eqref{denst};
\STATE Renew source domains by adding selected target samples while removing irrelevant source samples;
\STATE $\omega_k$: Calculate weights to select relevant source domains as in \eqref{selwe};
\STATE $\text{Keep}(\mathcal{D}_k^s)$: Select relevant source domains as in \eqref{keepsrc};
\STATE $\Phi$, $P$ $\leftarrow$ $\Phi_k$, $P_k$, $k \in K'$: Get target model as in \eqref{Mt};
\STATE $\mathcal{P}^t$ $\leftarrow$ $\bm x^t$: Collect target prediction as in \eqref{Pt};
\STATE $\mathcal{P}^{FM}$ $\leftarrow$ $\bm x^t$, $\mathcal{T}_c$: Collect frozen foundation model prediction as in \eqref{Pfm};
\STATE $\mathcal{T}$ $\leftarrow$ $\mathcal{P}^{FM}$, $\mathcal{P}^t$, $\mathcal{P}^V$, $\mathcal{P}'^V$: Update text prompts as in \eqref{exloss};
\STATE $\Phi$, $P$ $\leftarrow$ $\mathcal{P}^{FM}$, $\mathcal{P}^t$: Update target model as in \eqref{inloss};
\ENDFOR
\STATE {\bf Output:} Target label $\bm y_t$ via \eqref{yt}
\end{algorithmic} 
\end{algorithm}

\subsection{Selection status of source domains with training progress}
This work employs cosine distance to computing the distance between source/target samples and their cluster centers, which is expressed as:
\begin{equation*}
\text{Dis} = \frac{\bm x^*_{(\cdot)} \cdot \bm f_{kc}^s}{\|\bm x^*_{(\cdot)}\| \cdot \|\bm f_{kc}^s\|}, * \in {s,t}, (\cdot)=k \ \text{if} \ *=s
\end{equation*}

Average radius is expressed as:
\begin{equation*}
\text{Rd} = \frac{1}{n_{kc}}\sum_{i=1}^{n_{kc}} \frac{\bm x^*_{(\cdot)} \cdot \bm f_{kc}^s}{\|\bm x^*_{(\cdot)}\| \cdot \|\bm f_{kc}^s\|}, * \in {s,t}, (\cdot)=k \ \text{if} \ *=s
\end{equation*}
$n_{kc}$ is the number of $k$th source samples belonging to $c$th class.

Root mean square radius is expressed as:
\begin{equation*}
\text{Rd} = \sqrt{\frac{1}{n_{kc}}\sum_{i=1}^{n_{kc}} (\frac{\bm x^*_{(\cdot)} \cdot \bm f_{kc}^s}{\|\bm x^*_{(\cdot)}\| \cdot \|\bm f_{kc}^s\|})^2}, * \in {s,t}, (\cdot)=k \ \text{if} \ *=s
\end{equation*}
If employing root mean square radius, $\alpha = 1.5$.

Maximum radius is expressed as:
\begin{equation*}
\text{Rd} = \underset{i \in [1, n_{kc}]}{\max} \frac{\bm x^{*i}_{(\cdot)} \cdot \bm f_{kc}^s}{\|\bm x^{*i}_{(\cdot)}\| \cdot \|\bm f_{kc}^s\|}, * \in {s,t}, (\cdot)=k \ \text{if} \ *=s
\end{equation*}

\begin{figure}[htbp]
\centering
\subfloat[Target R]{
\begin{minipage}[t]{0.45\linewidth}
\centering
\includegraphics[height=2cm,width=4cm]{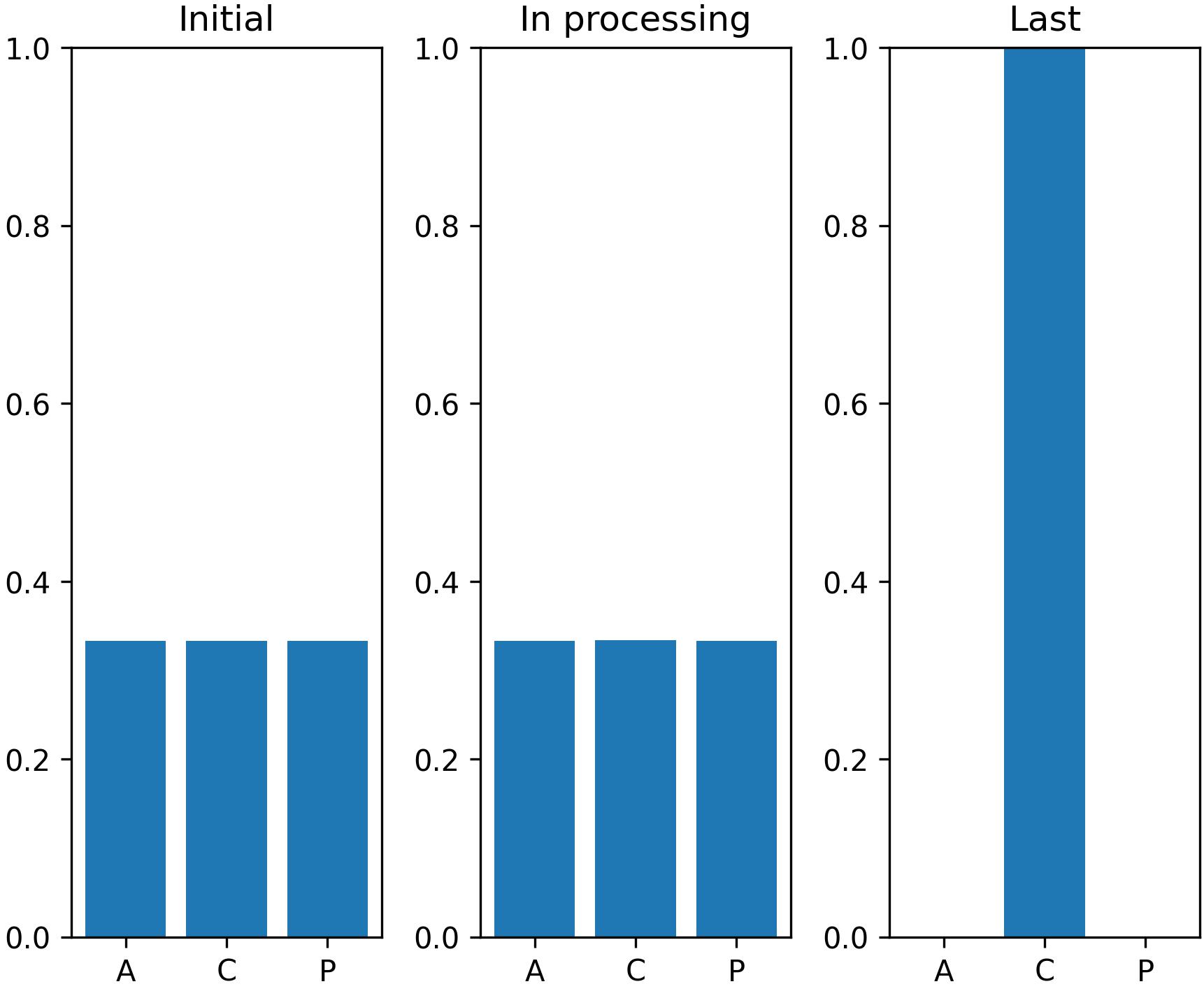}
\end{minipage}
}
\subfloat[Target P]{
\begin{minipage}[t]{0.45\linewidth}
\centering
\includegraphics[height=2cm,width=4cm]{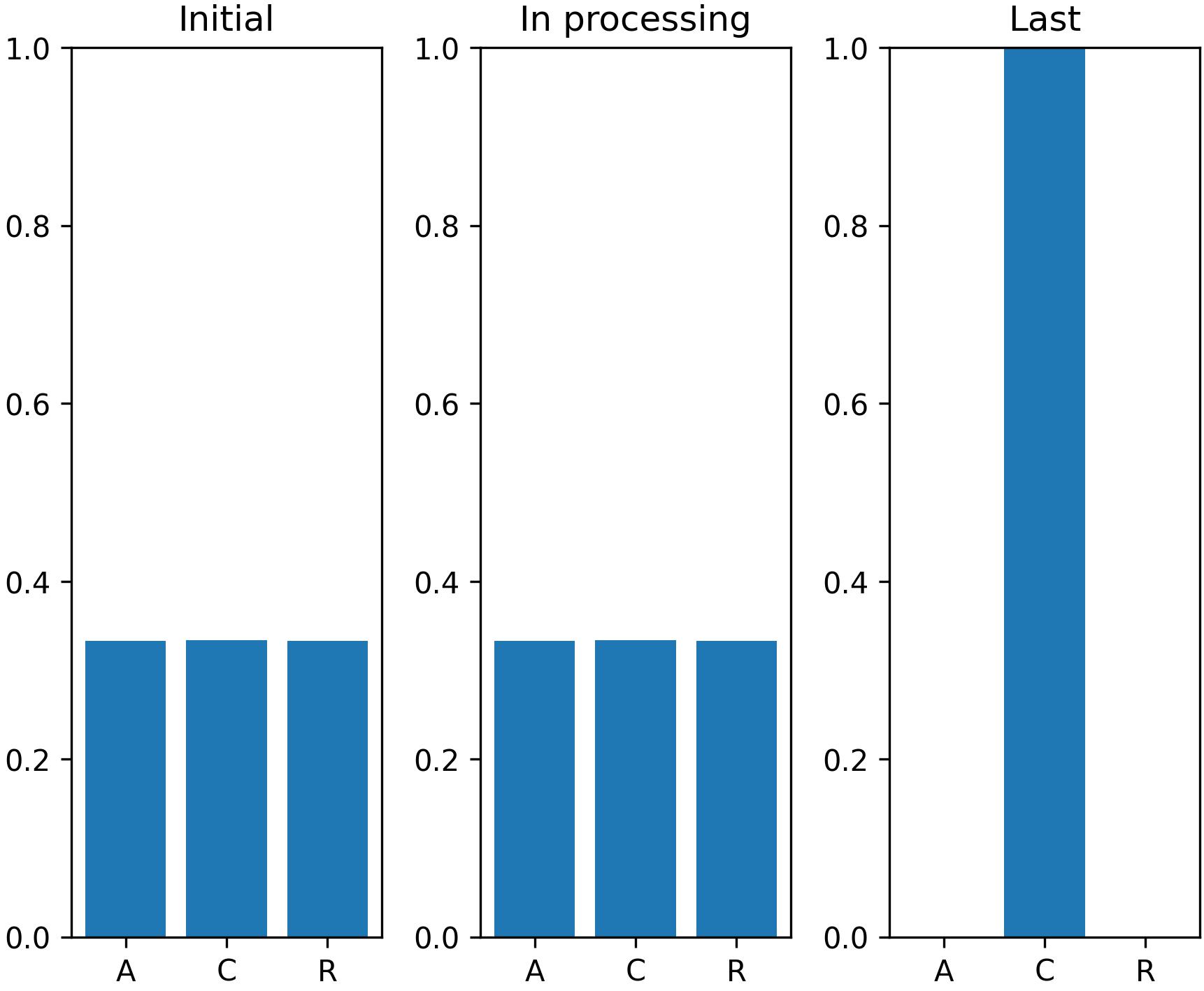}
\end{minipage}
}

\subfloat[Target C]{
\begin{minipage}[t]{0.45\linewidth}
\centering
\includegraphics[height=2cm,width=4cm]{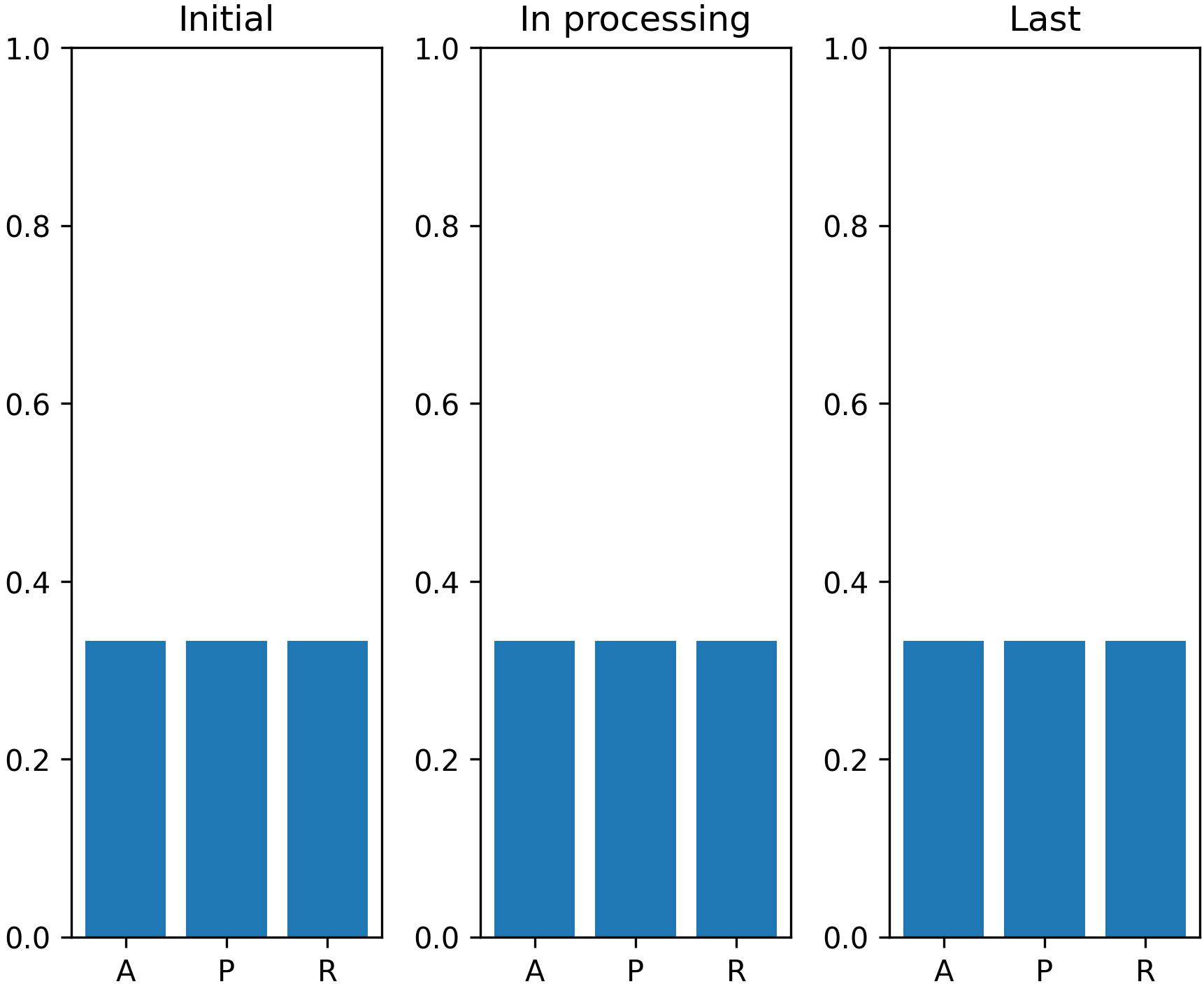}
\end{minipage}
}
\subfloat[Target A]{
\begin{minipage}[t]{0.45\linewidth}
\centering
\includegraphics[height=2cm,width=4cm]{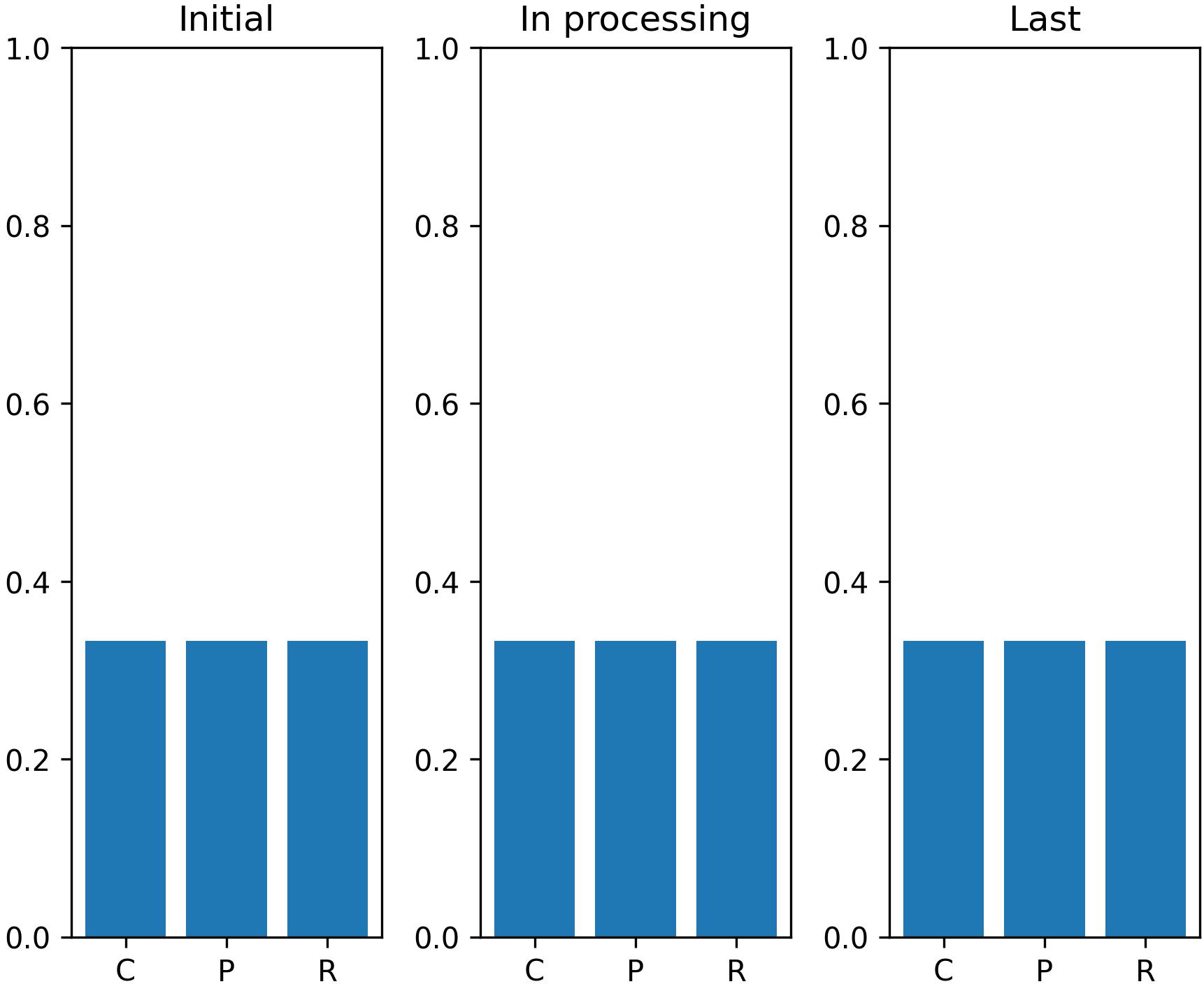}
\end{minipage}
}
\caption{Source domain selection status using average radius of OfficeHome.}\label{selmean}
\end{figure}

\begin{figure}[htbp]
\centering
\subfloat[Target R]{
\begin{minipage}[t]{0.45\linewidth}
\centering
\includegraphics[height=2cm,width=4cm]{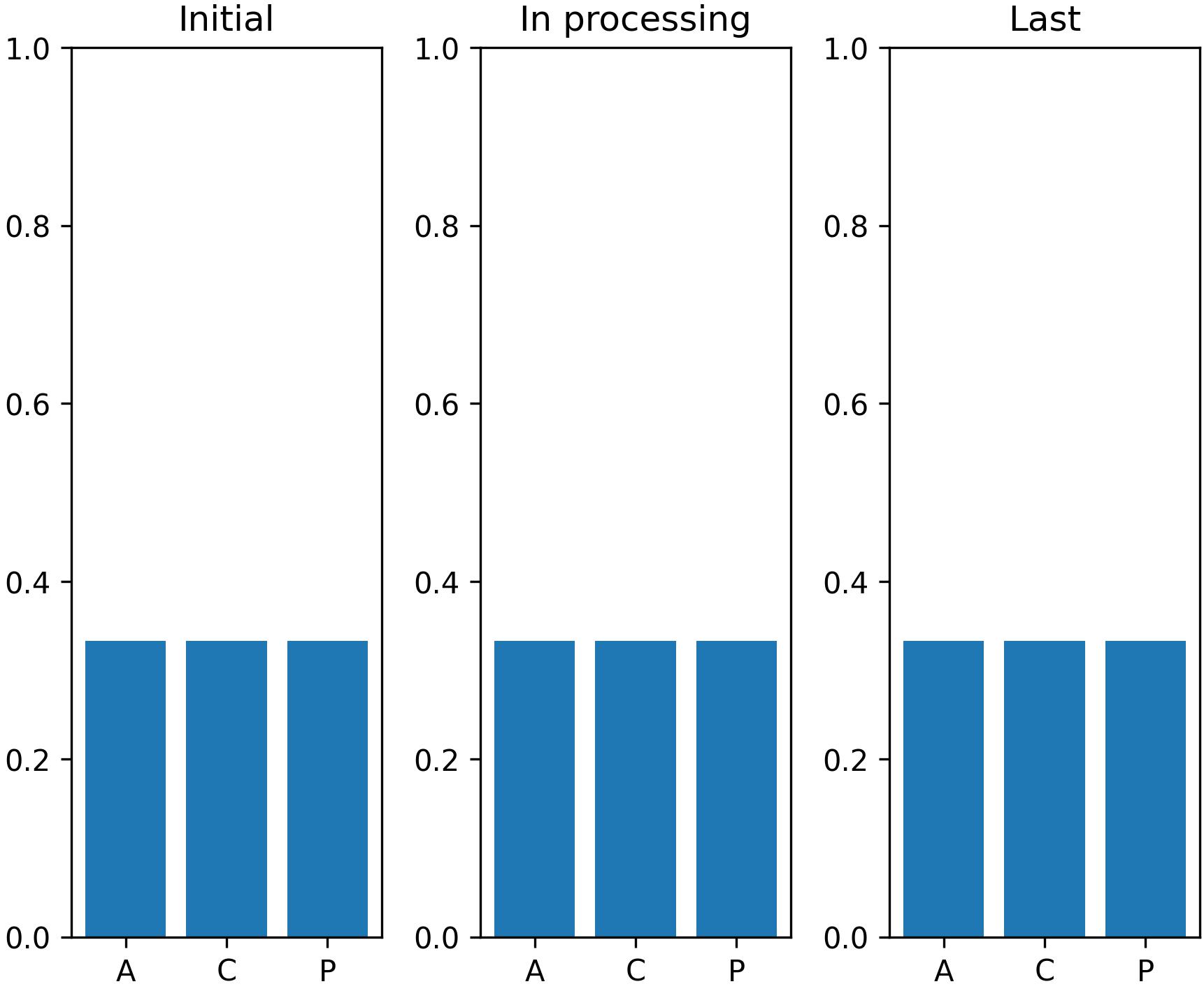}
\end{minipage}
}
\subfloat[Target P]{
\begin{minipage}[t]{0.45\linewidth}
\centering
\includegraphics[height=2cm,width=4cm]{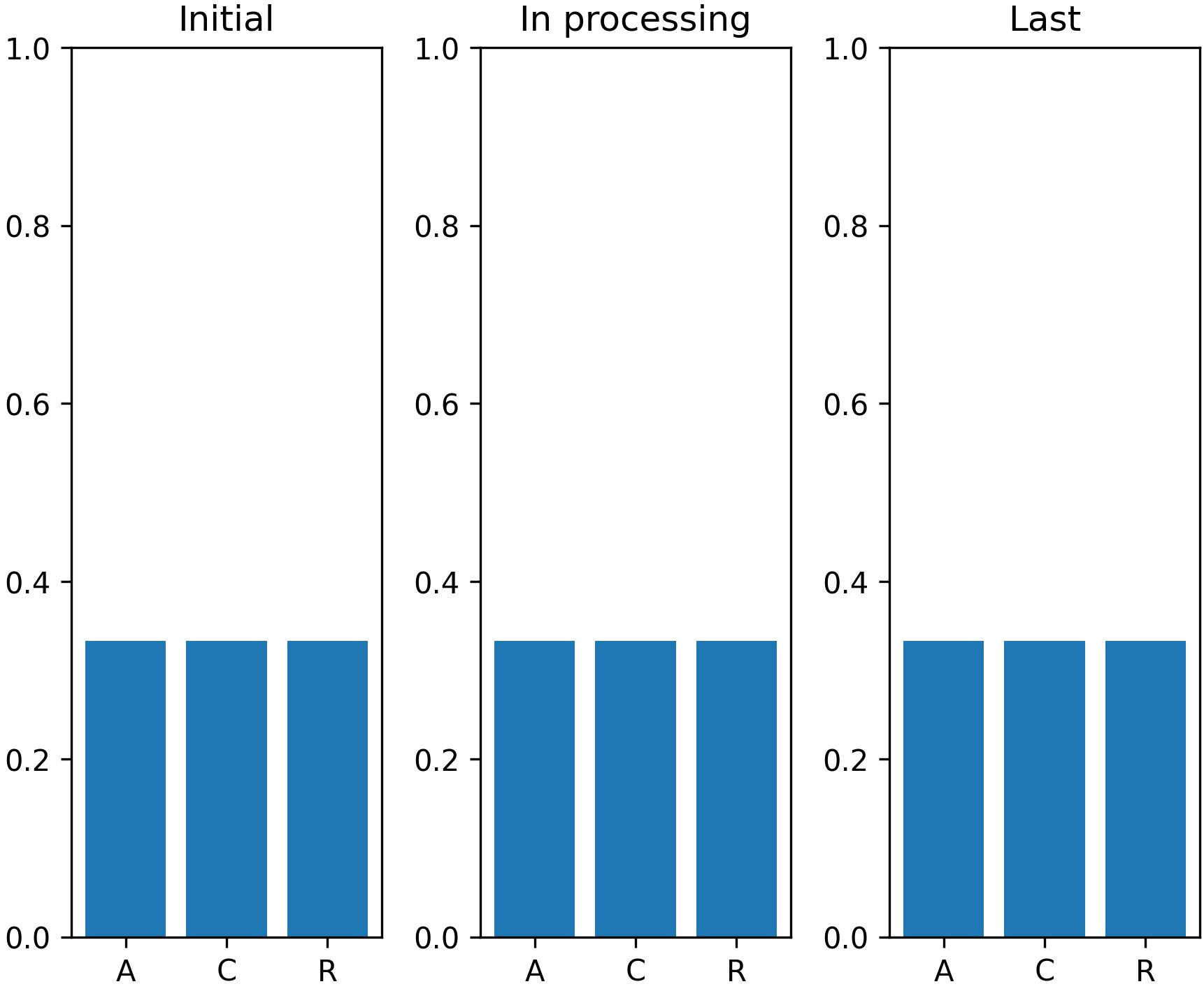}
\end{minipage}
}

\subfloat[Target C]{
\begin{minipage}[t]{0.45\linewidth}
\centering
\includegraphics[height=2cm,width=4cm]{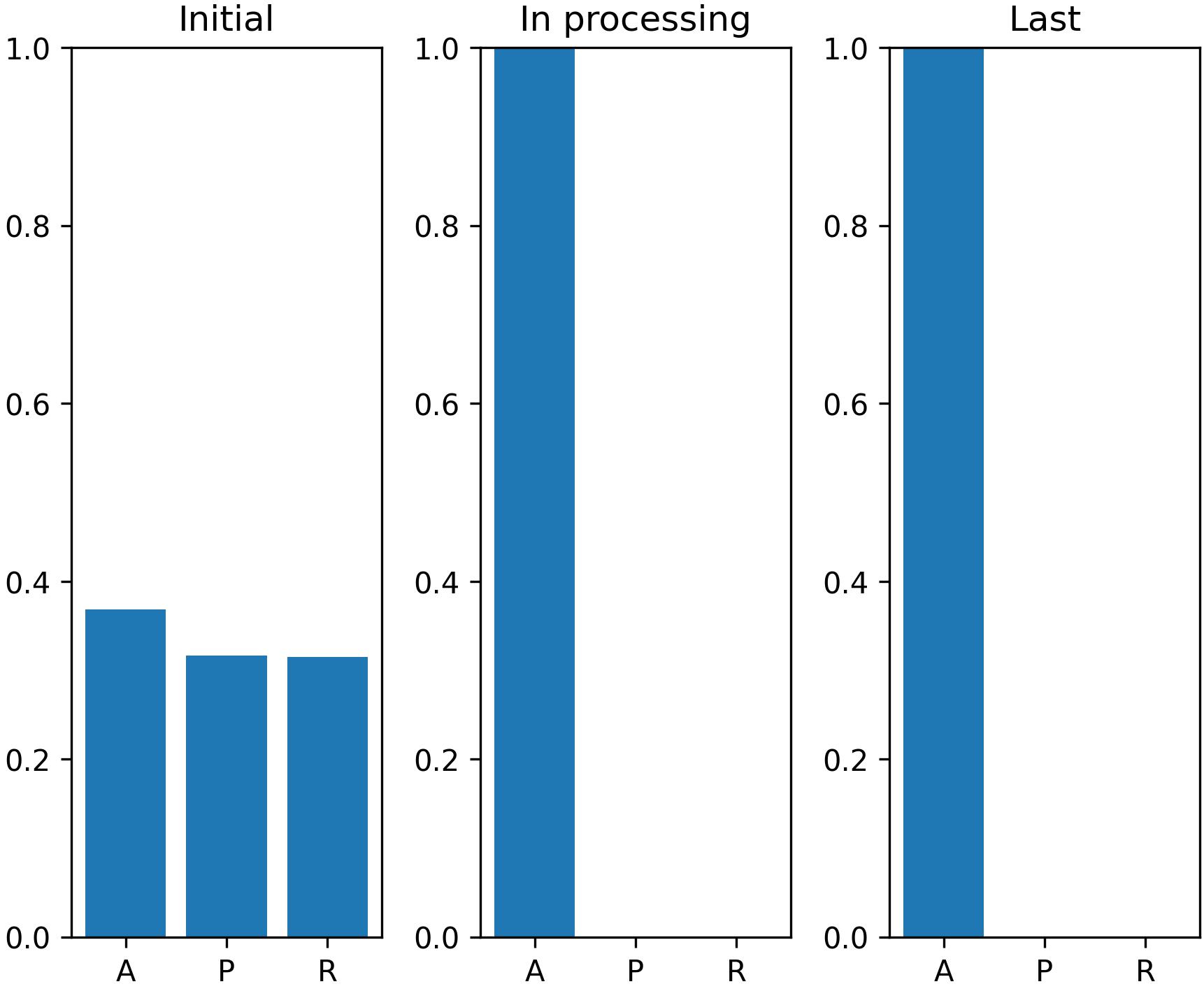}
\end{minipage}
}
\subfloat[Target A]{
\begin{minipage}[t]{0.45\linewidth}
\centering
\includegraphics[height=2cm,width=4cm]{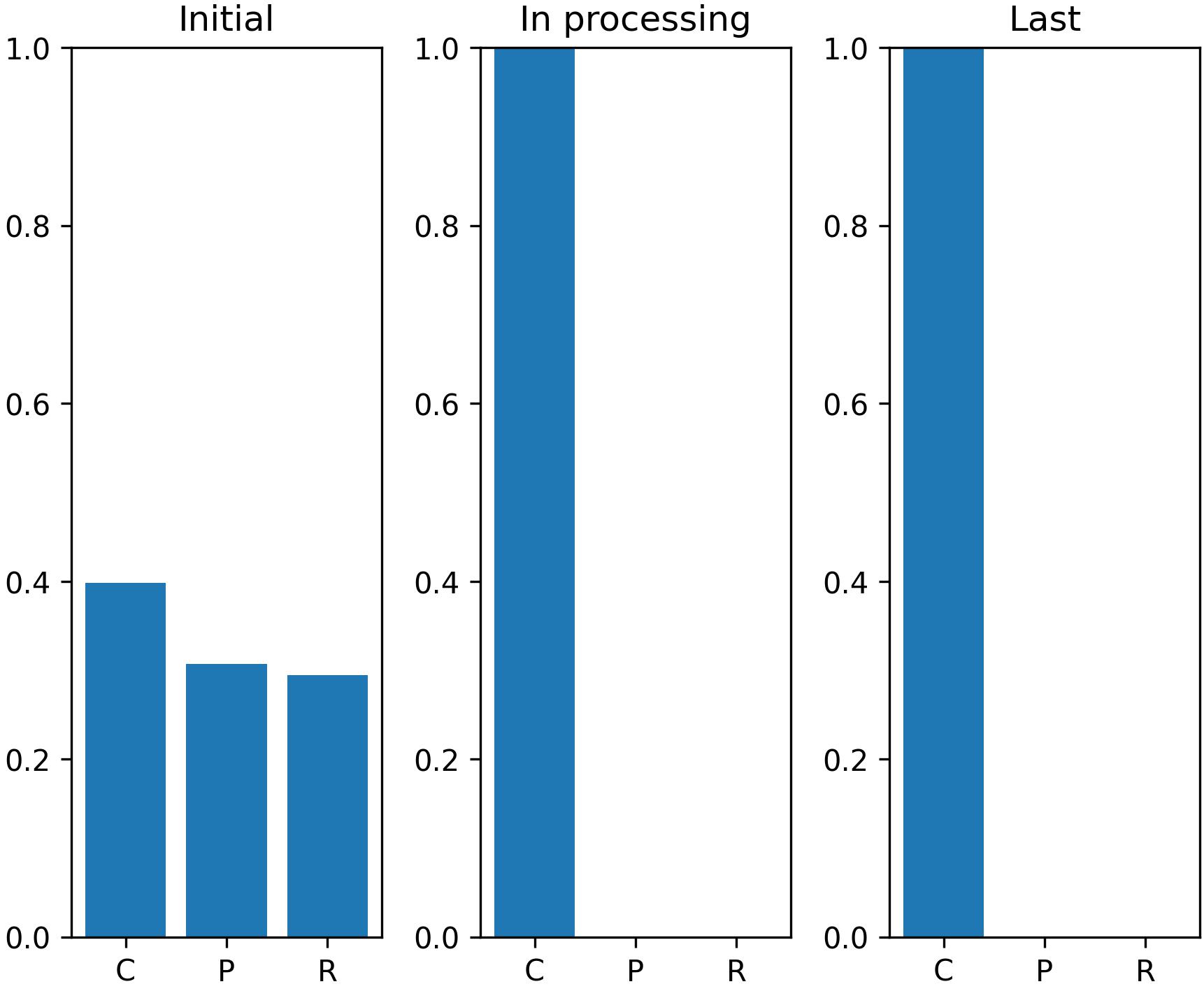}
\end{minipage}
}
\caption{Source domain selection status using root mean square radius of OfficeHome.}\label{selrms}
\end{figure}

\begin{figure}[htbp]
\centering
\subfloat[Target R]{
\begin{minipage}[t]{0.45\linewidth}
\centering
\includegraphics[height=2cm,width=4cm]{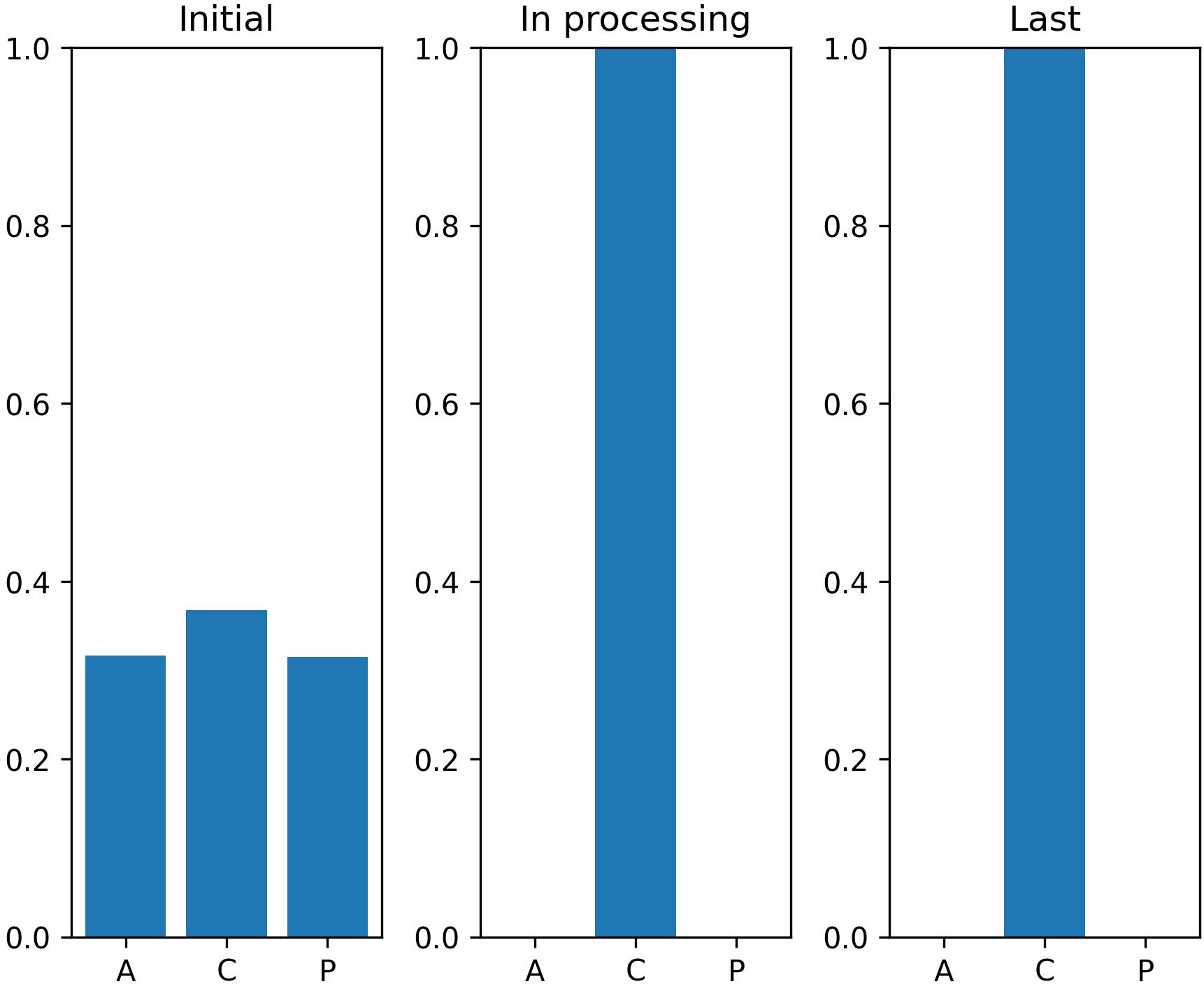}
\end{minipage}
}
\subfloat[Target P]{
\begin{minipage}[t]{0.45\linewidth}
\centering
\includegraphics[height=2cm,width=4cm]{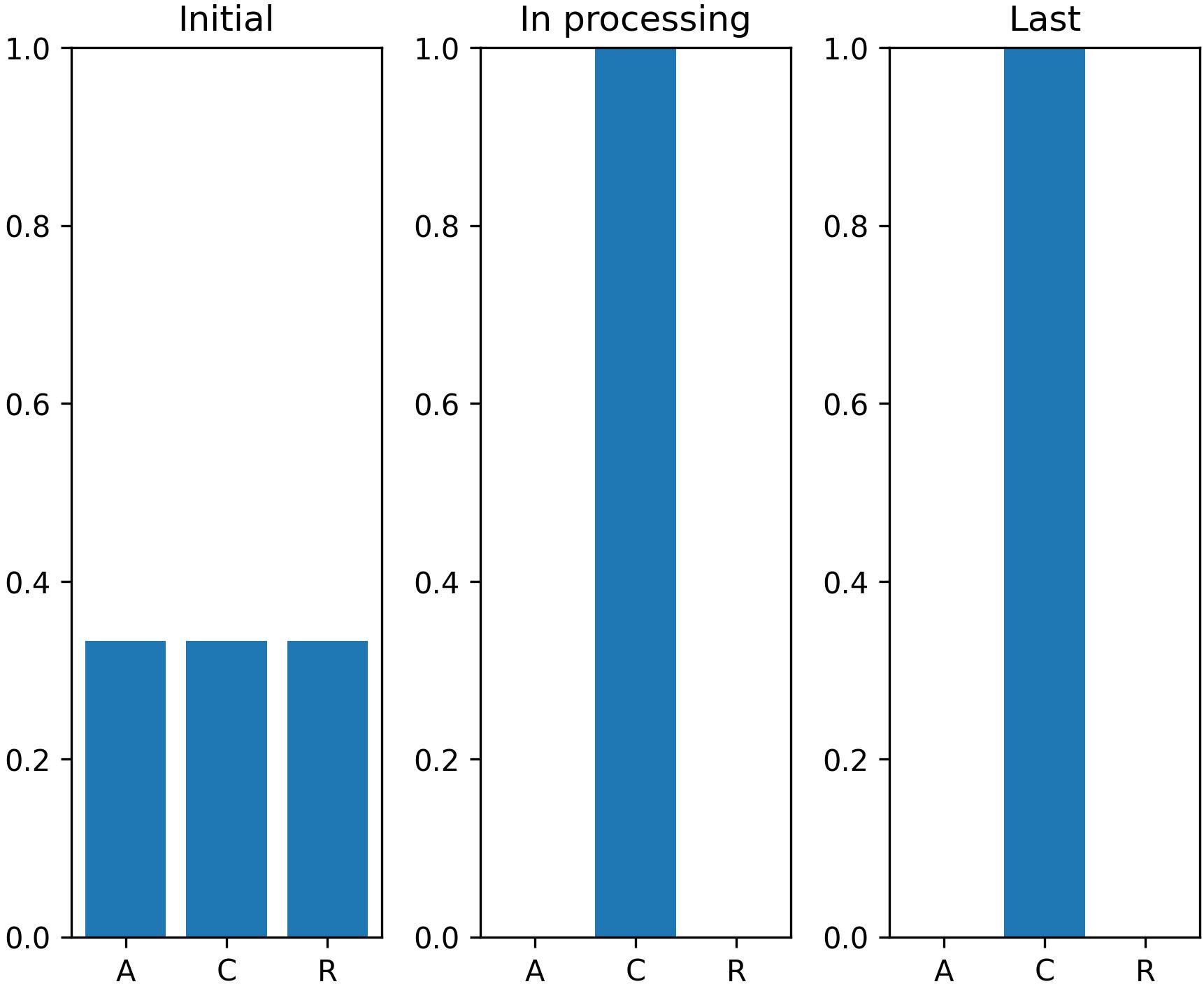}
\end{minipage}
}

\subfloat[Target C]{
\begin{minipage}[t]{0.45\linewidth}
\centering
\includegraphics[height=2cm,width=4cm]{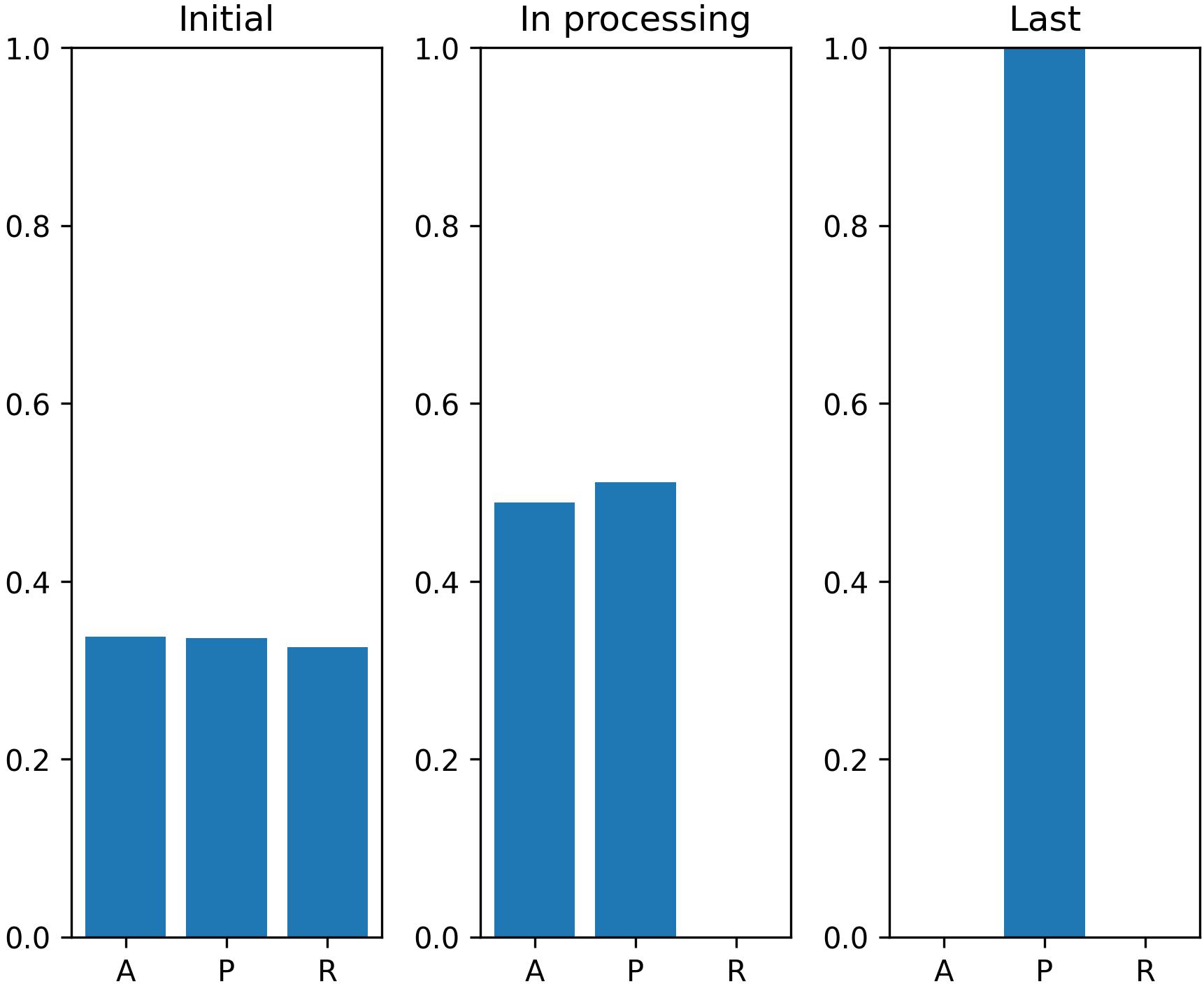}
\end{minipage}
}
\subfloat[Target A]{
\begin{minipage}[t]{0.45\linewidth}
\centering
\includegraphics[height=2cm,width=4cm]{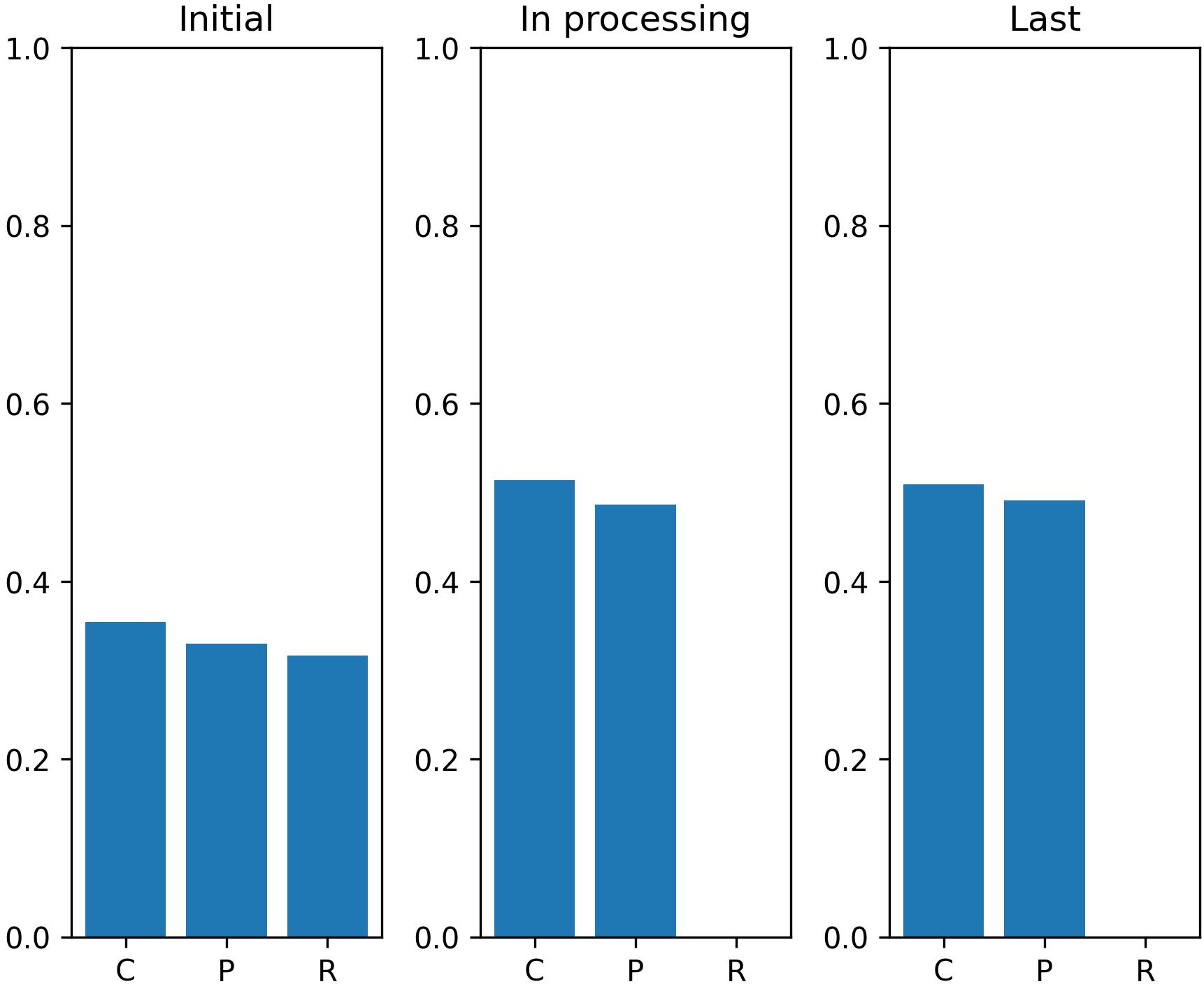}
\end{minipage}
}
\caption{Source domain selection status using maximum radius of OfficeHome.}\label{selmax}
\end{figure}

Autonomous source selection status is displayed as Figures \ref{selmean}, \ref{selrms} and \ref{selmax}.

The results show that different cluster radius metrics play a key role in autonomously selecting source domains. The mean radius is more likely to select relevant domains because the proportion of target samples falling within the cluster radius contributes to the computation of selection weights. The root mean square radius and the maximum radius generally provide larger radius values, resulting in more target samples being added to the updated source domains and thereby increasing label noise when training the federated model.

\subsection{Description of compared baselines}
Settings and core techniques of the baselines are described in Table \ref{baseline}:

\begin{table}[htbp]
  \centering
  \caption{The baselines. ``SF" means source-free, ``MS" means multi-source, ``BB" means backbone structure, while ``FFM" indicates frozen foundation model.}
  \setlength{\tabcolsep}{1.2mm}{
    \begin{tabular}{ccccccc}
    \toprule
    Baseline & Venue & SF  & MS  & BB & FFM\\
    \midrule
    CAiDA \cite{330dong2021confident} & NeurIPS'21 & \checkmark    & \checkmark     & RN  & $\times$    \\
    FixBi \cite{658na2021fixbi} & CVPR'21 & \checkmark    & $\times$     & RN     & $\times$    \\
    DeiT \cite{661touvron2021training} & ICML'21 & \checkmark    & $\times$     & ViT     & $\times$    \\
    CDTrans \cite{664xu2021cdtrans} & ICLR'21 & \checkmark    & $\times$     & ViT     & $\times$    \\
    SSRT \cite{663sun2022safe} & CVPR'22 & \checkmark    & $\times$     & ViT     & $\times$    \\
    DSiT \cite{567sanyal2023domain} & ICCV'23 & \checkmark    & $\times$     & ViT     & $\times$    \\
    SSD \cite{457li2023multidomain}   & TCYB'23 & $\times$     & \checkmark    & RN     & $\times$     \\
    Co-MDA \cite{483liu2023co} & TCSVT'23 & \checkmark    & \checkmark     & RN     & $\times$    \\
    DCL \cite{622tian2023dcl} & TCSVT'23 & \checkmark    & $\times$     & RN     & $\times$    \\
    GSDE \cite{657westfechtel2024gradual} & WACV'24 & \checkmark    & $\times$     & RN     & $\times$    \\
    MPA \cite{568chen2024multi}   & NeurIPS'23 & $\times$     & \checkmark     & RN     & \checkmark     \\
    SEAL \cite{585xia2024separation}  & AAAI'24 & \checkmark     & \checkmark     & RN    & $\times$    \\
    KGCDE \cite{561wong2024graph} & PR'24  & $\times$     & \checkmark     & RN     & $\times$     \\
    DSACDIC \cite{659zhao2024deep} & WACV'24  & $\times$     & $\times$     & RN     & $\times$     \\
    LCFD \cite{574tang2024unified}  & ArXiv'24 & \checkmark     & $\times$  & RN  & \checkmark      \\
    DAMP \cite{656du2024domain}  & CVPR'24 & \checkmark     & $\times$  & RN  & \checkmark      \\
    Ucon-SFDA \cite{655xu2024revisiting}  & ICLR'24 & \checkmark     & $\times$  & RN  & $\times$      \\
    FuzHDA \cite{628li2025fuzzy}  & TFS'25 & \checkmark     & \checkmark  & RN  & \checkmark      \\
    ProDe \cite{653tang2024proxy}  & ICLR'25 & \checkmark     & $\times$  & RN  & $\times$      \\
    TIGM \cite{660zhu2025revisiting}  & CVPR'25 & \checkmark     & $\times$  & RN  & $\times$      \\
    \bottomrule
    \bottomrule
\end{tabular}}
  \label{baseline}%
\end{table}%

\subsection{Visualization of tasks from OfficeHome with and without source data}
T-SNE visualization of all tasks from OfficeHome is displayed as following Figures.

\begin{figure}[htbp]
\centering
\subfloat[Selected C and A]{
\begin{minipage}[t]{0.3\linewidth}
\centering
\includegraphics[height=2.5cm,width=2.5cm]{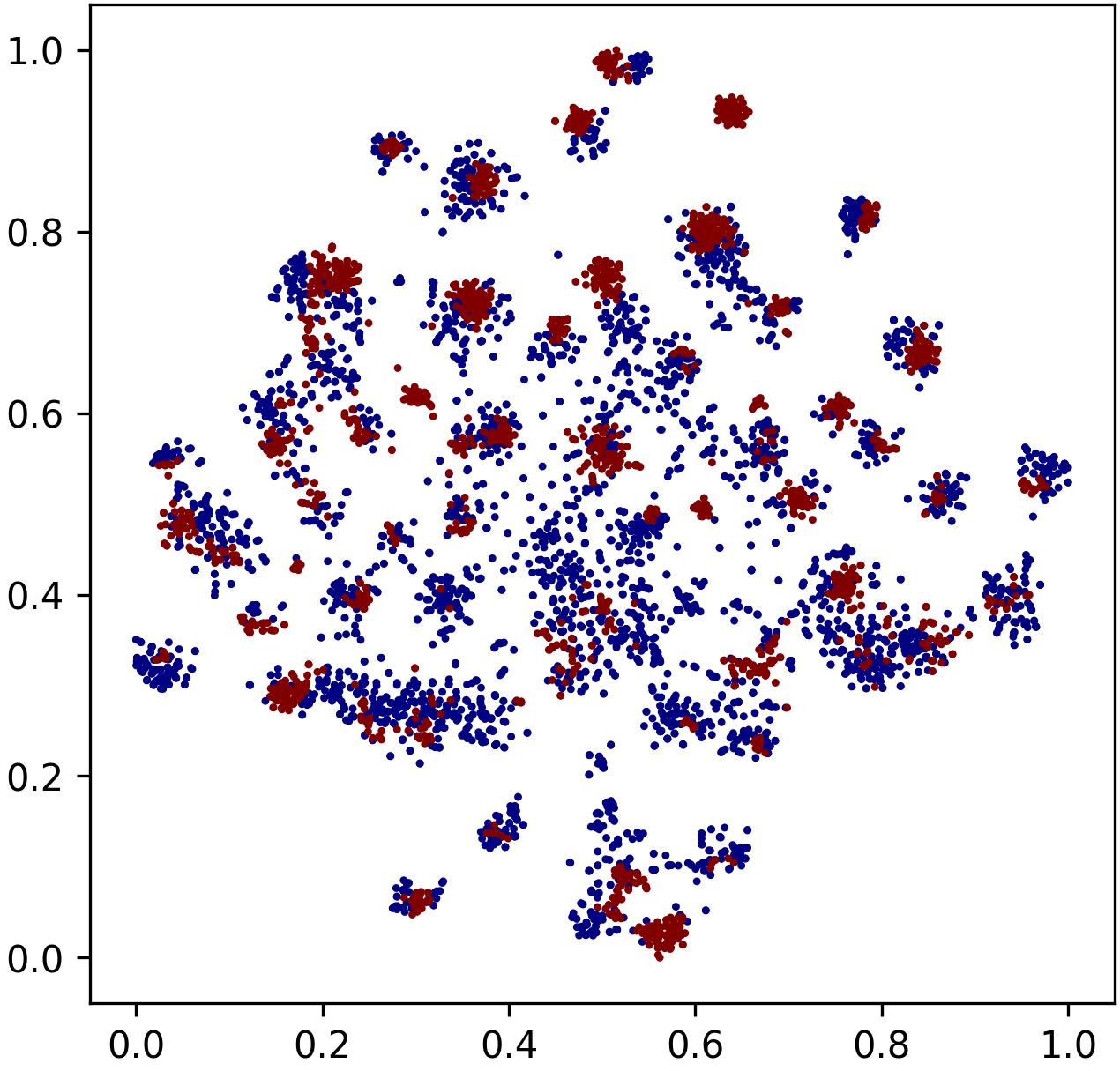}
\end{minipage}
}
\subfloat[Selected P and A]{
\begin{minipage}[t]{0.3\linewidth}
\centering
\includegraphics[height=2.5cm,width=2.5cm]{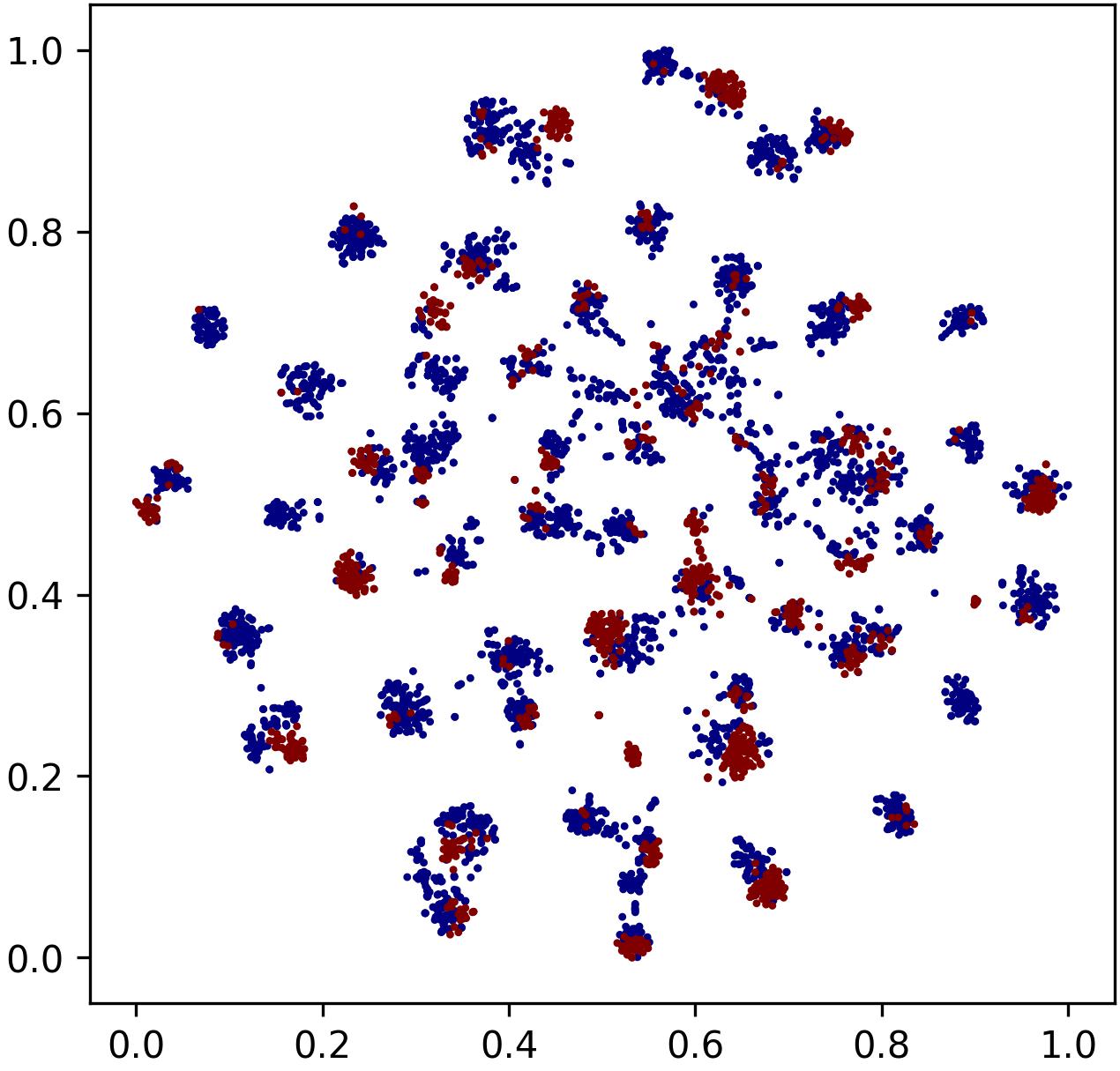}
\end{minipage}
}
\subfloat[Selected R and A]{
\begin{minipage}[t]{0.3\linewidth}
\centering
\includegraphics[height=2.5cm,width=2.5cm]{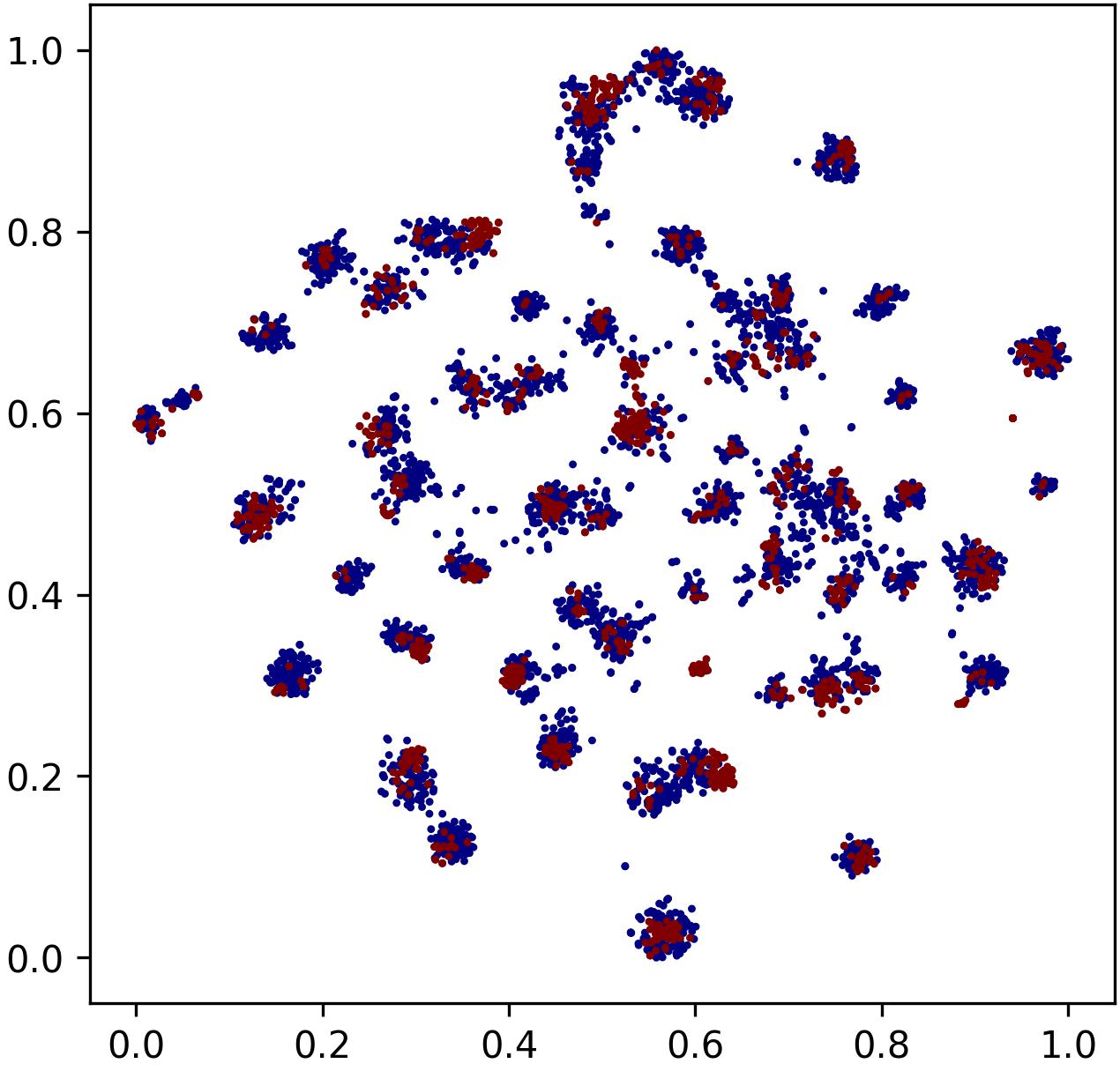}
\end{minipage}
}

\subfloat[Adapted C and A]{
\begin{minipage}[t]{0.3\linewidth}
\centering
\includegraphics[height=2.5cm,width=2.5cm]{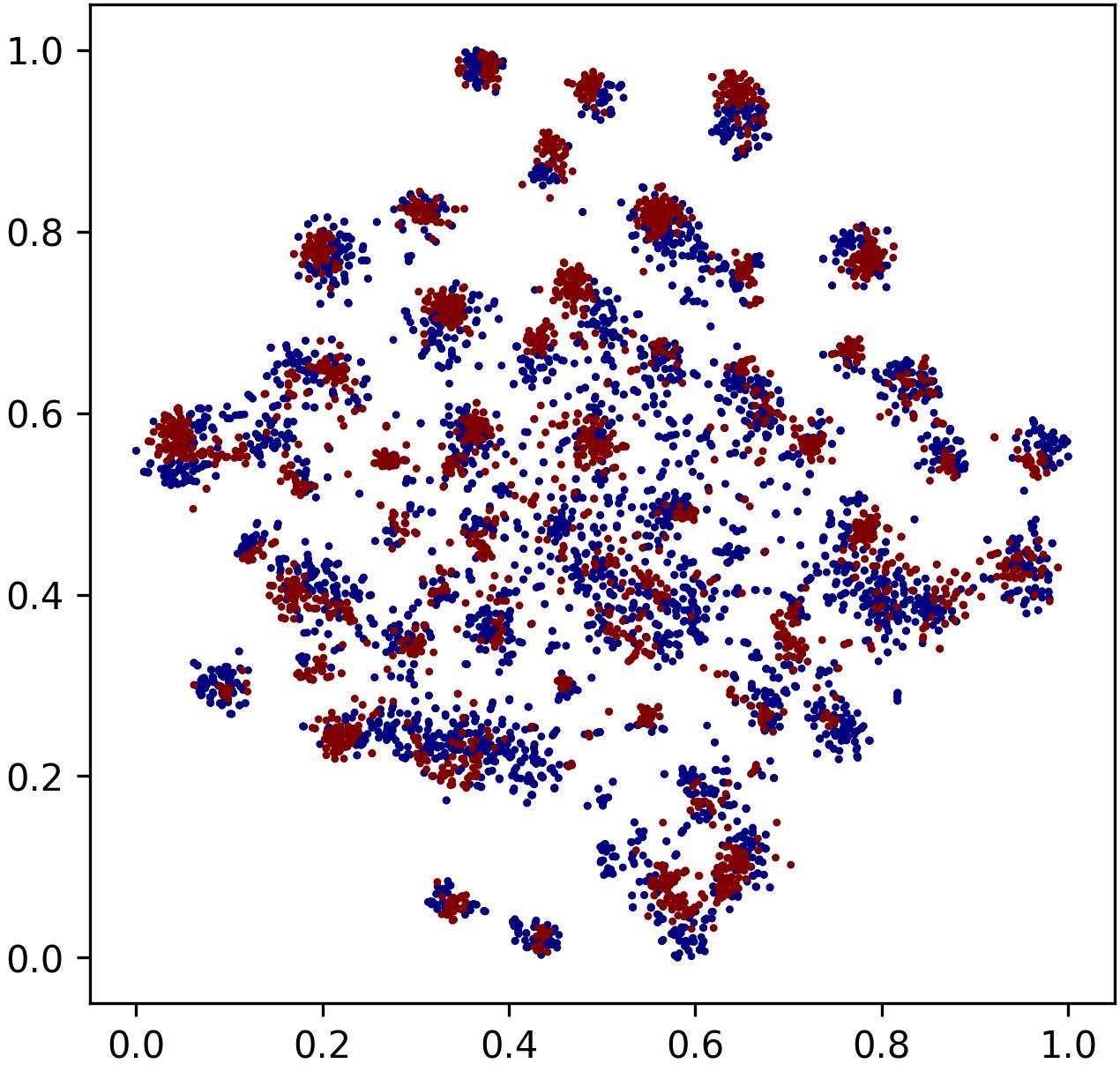}
\end{minipage}
}
\subfloat[Adapted P and A]{
\begin{minipage}[t]{0.3\linewidth}
\centering
\includegraphics[height=2.5cm,width=2.5cm]{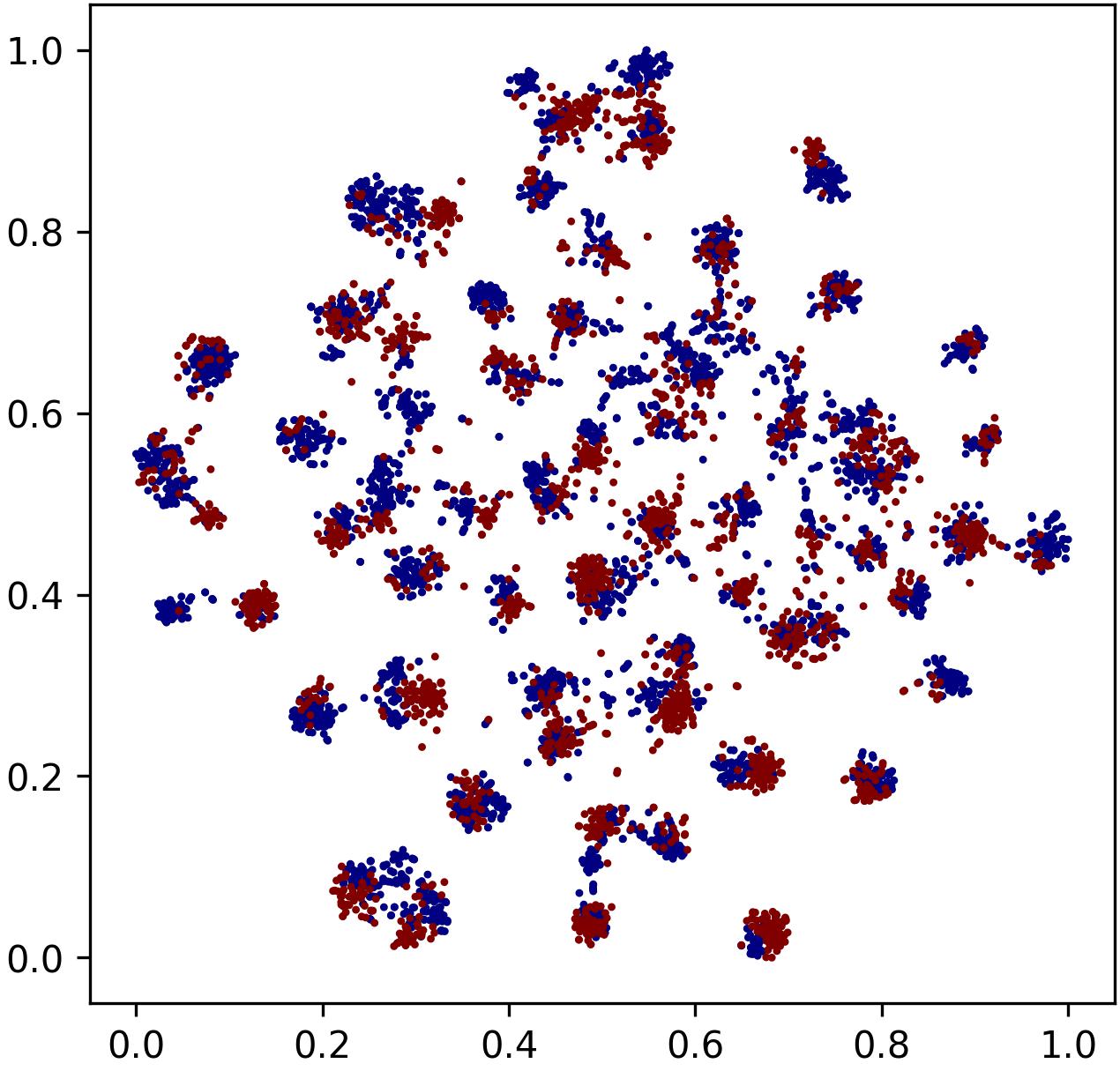}
\end{minipage}
}
\subfloat[Adapted R and A]{
\begin{minipage}[t]{0.3\linewidth}
\centering
\includegraphics[height=2.5cm,width=2.5cm]{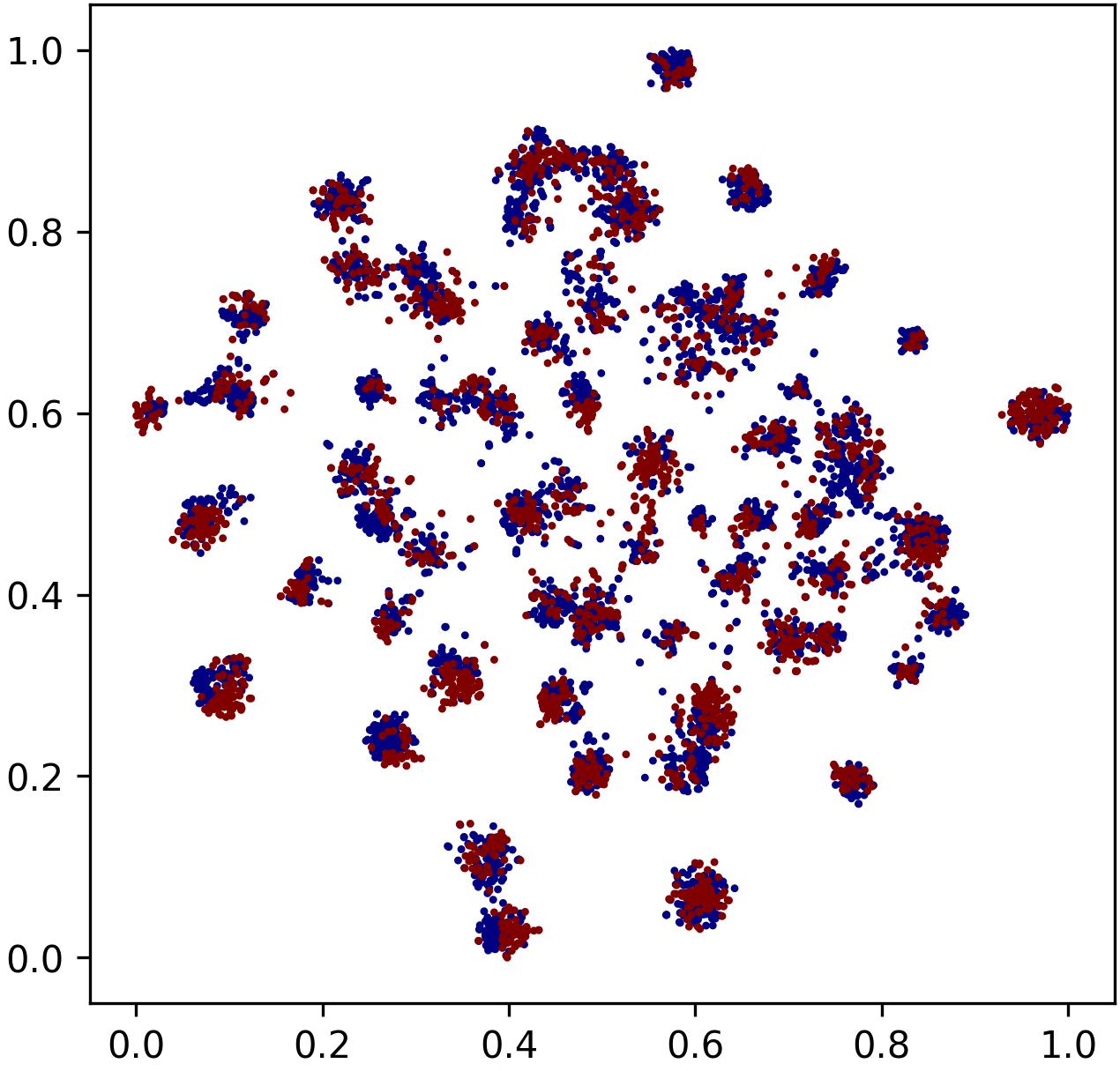}
\end{minipage}
}
\caption{T-SNE of adapted source and target data on task A from OfficeHome.}\label{tsneA}
\end{figure}

\begin{figure}[htbp]
\centering
\subfloat[Selected A and C]{
\begin{minipage}[t]{0.3\linewidth}
\centering
\includegraphics[height=2.5cm,width=2.5cm]{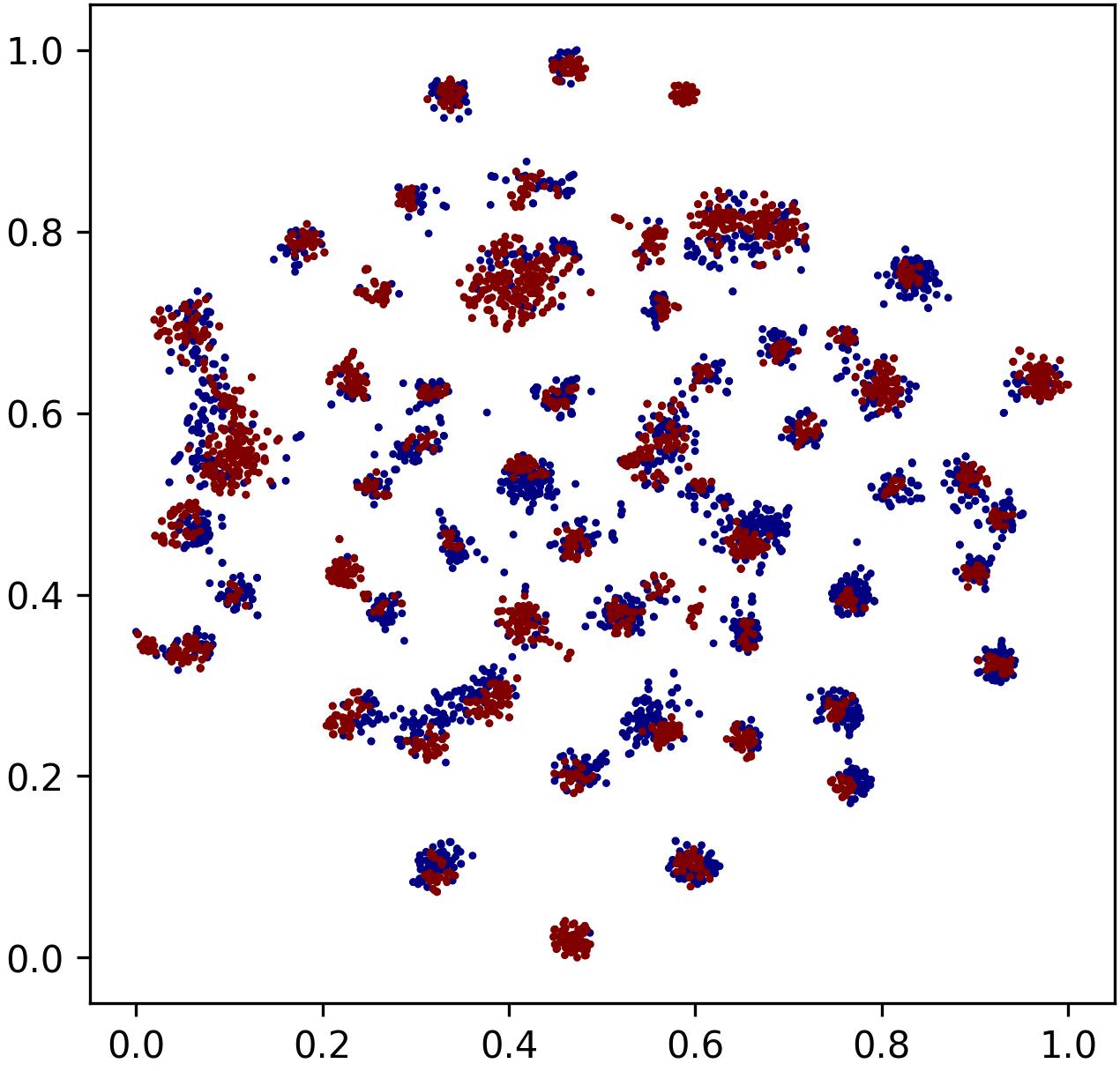}
\end{minipage}
}
\subfloat[Selected P and C]{
\begin{minipage}[t]{0.3\linewidth}
\centering
\includegraphics[height=2.5cm,width=2.5cm]{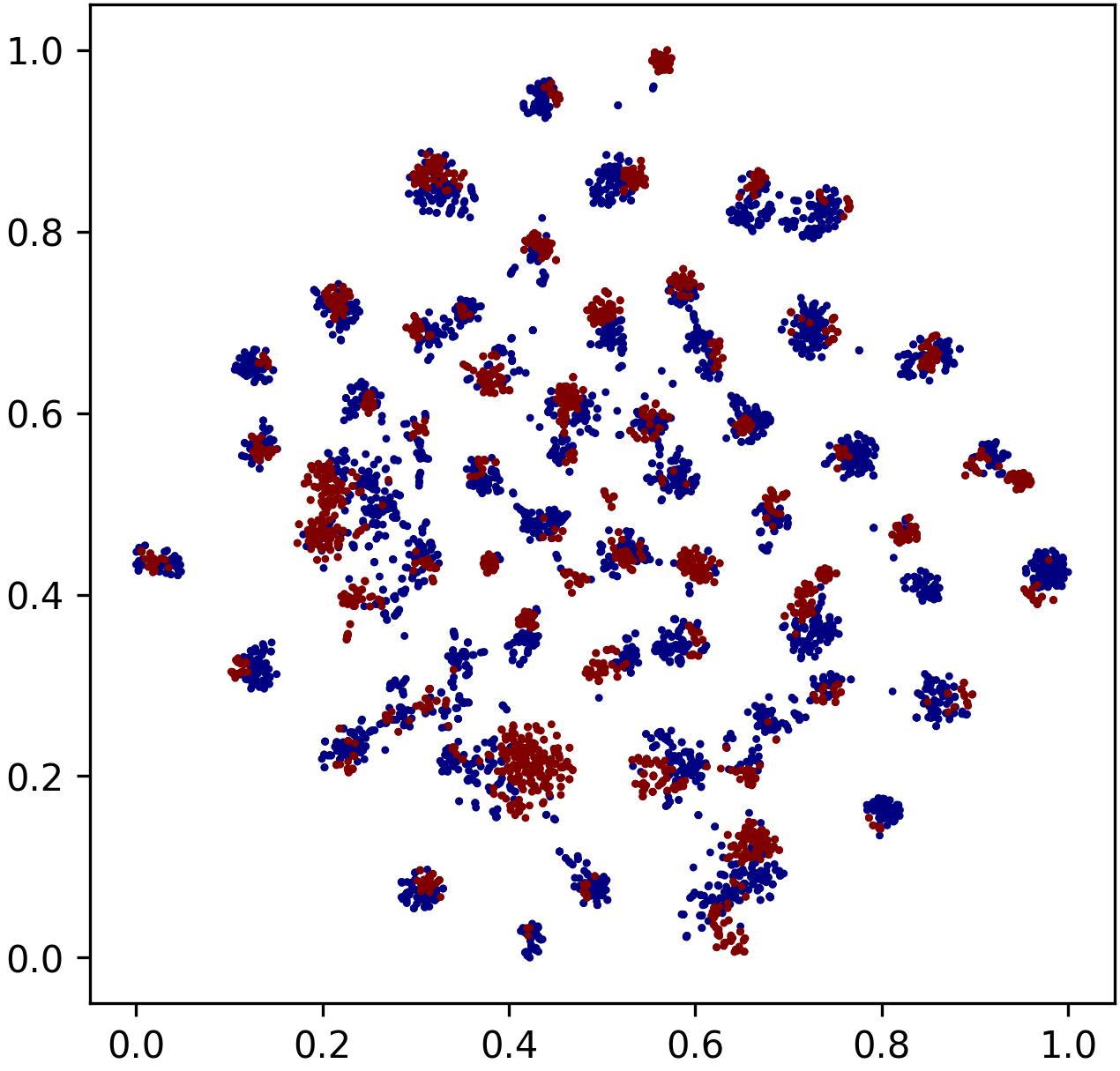}
\end{minipage}
}
\subfloat[Selected R and C]{
\begin{minipage}[t]{0.3\linewidth}
\centering
\includegraphics[height=2.5cm,width=2.5cm]{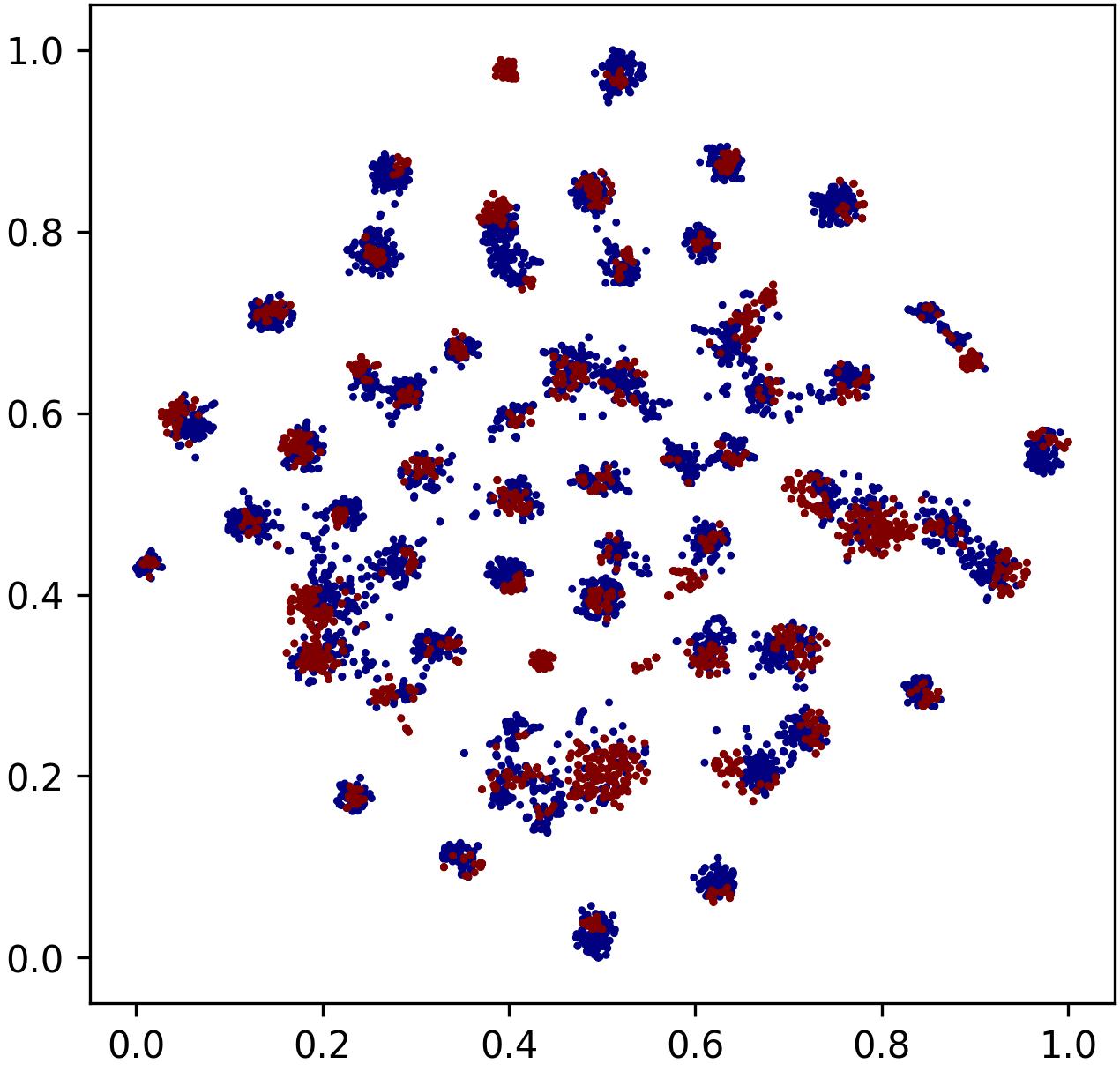}
\end{minipage}
}

\subfloat[Adapted A and C]{
\begin{minipage}[t]{0.3\linewidth}
\centering
\includegraphics[height=2.5cm,width=2.5cm]{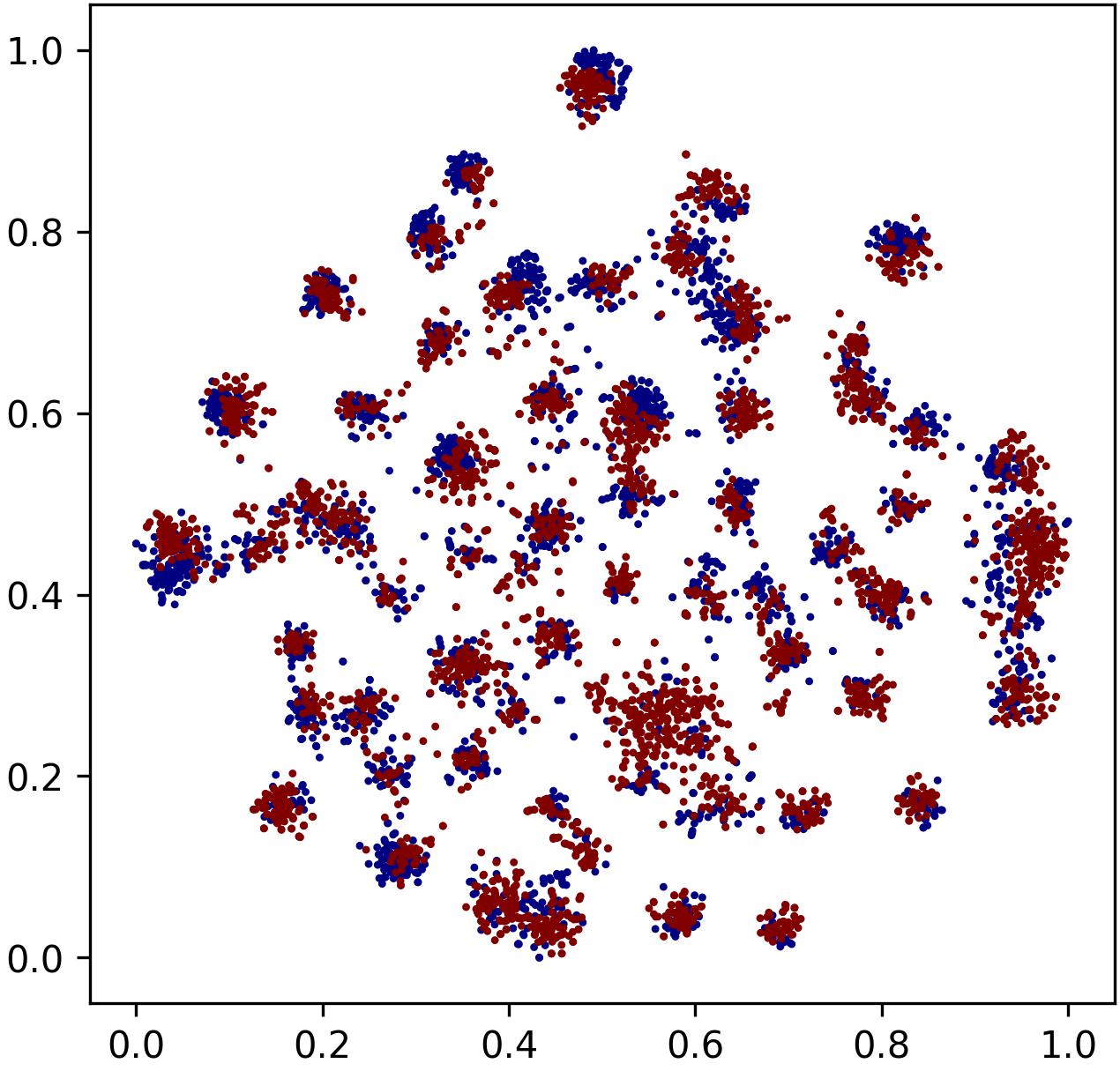}
\end{minipage}
}
\subfloat[Adapted P and C]{
\begin{minipage}[t]{0.3\linewidth}
\centering
\includegraphics[height=2.5cm,width=2.5cm]{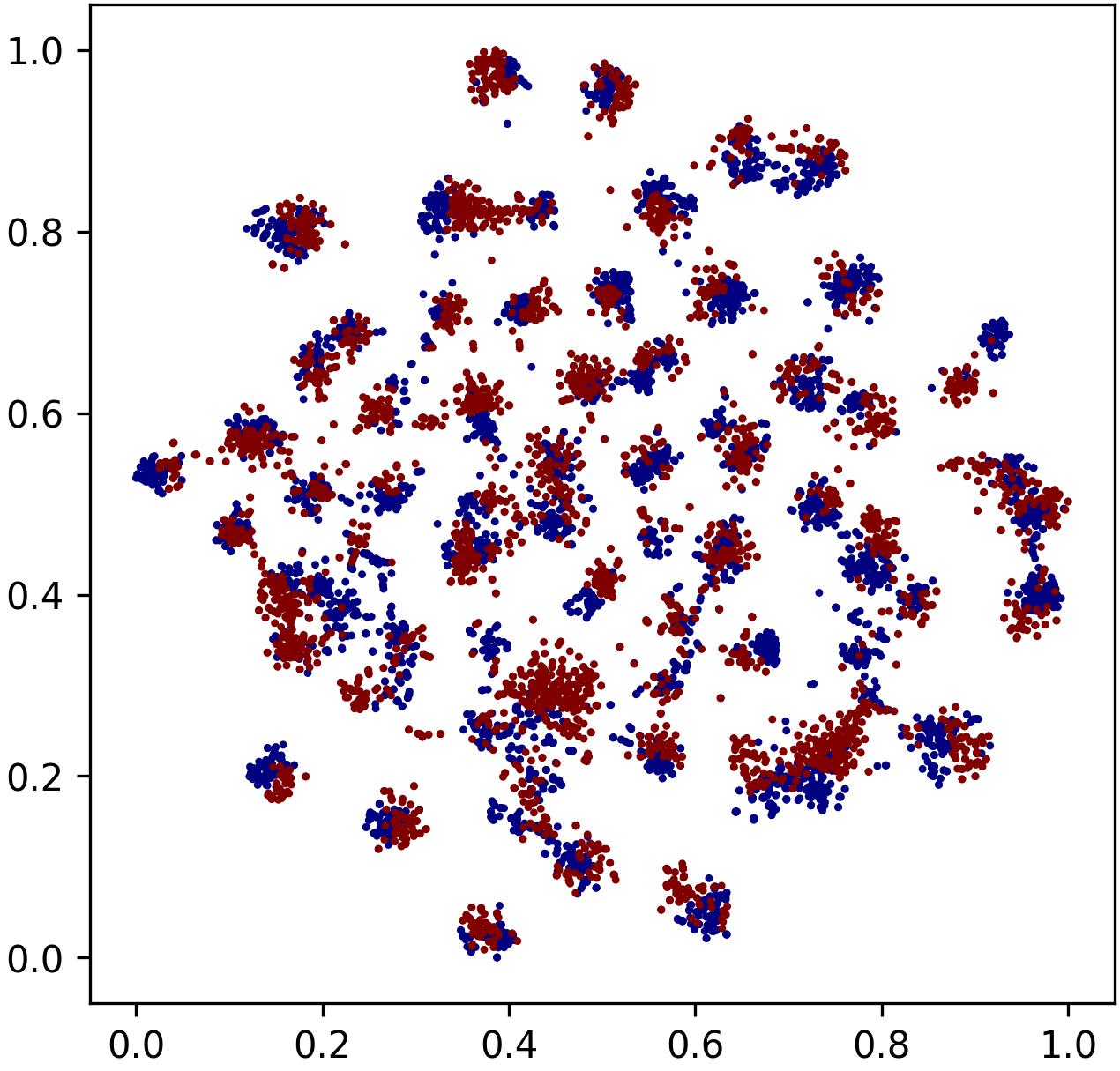}
\end{minipage}
}
\subfloat[Adapted R and C]{
\begin{minipage}[t]{0.3\linewidth}
\centering
\includegraphics[height=2.5cm,width=2.5cm]{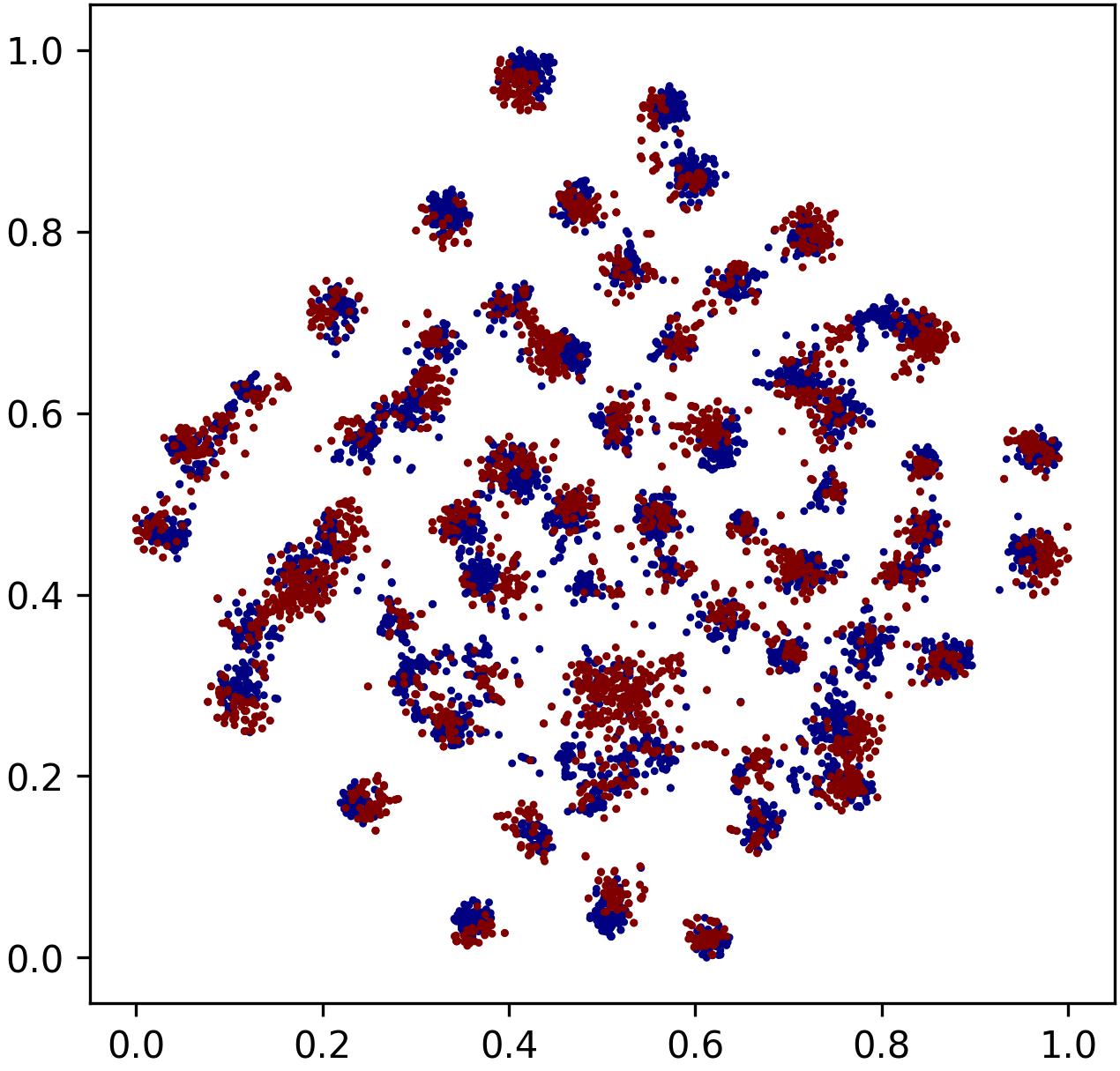}
\end{minipage}
}
\caption{T-SNE of adapted source and target data on task C from OfficeHome.}\label{tsneC}
\end{figure}

\begin{figure}[htbp]
\centering
\subfloat[Selected A and P]{
\begin{minipage}[t]{0.3\linewidth}
\centering
\includegraphics[height=2.5cm,width=2.5cm]{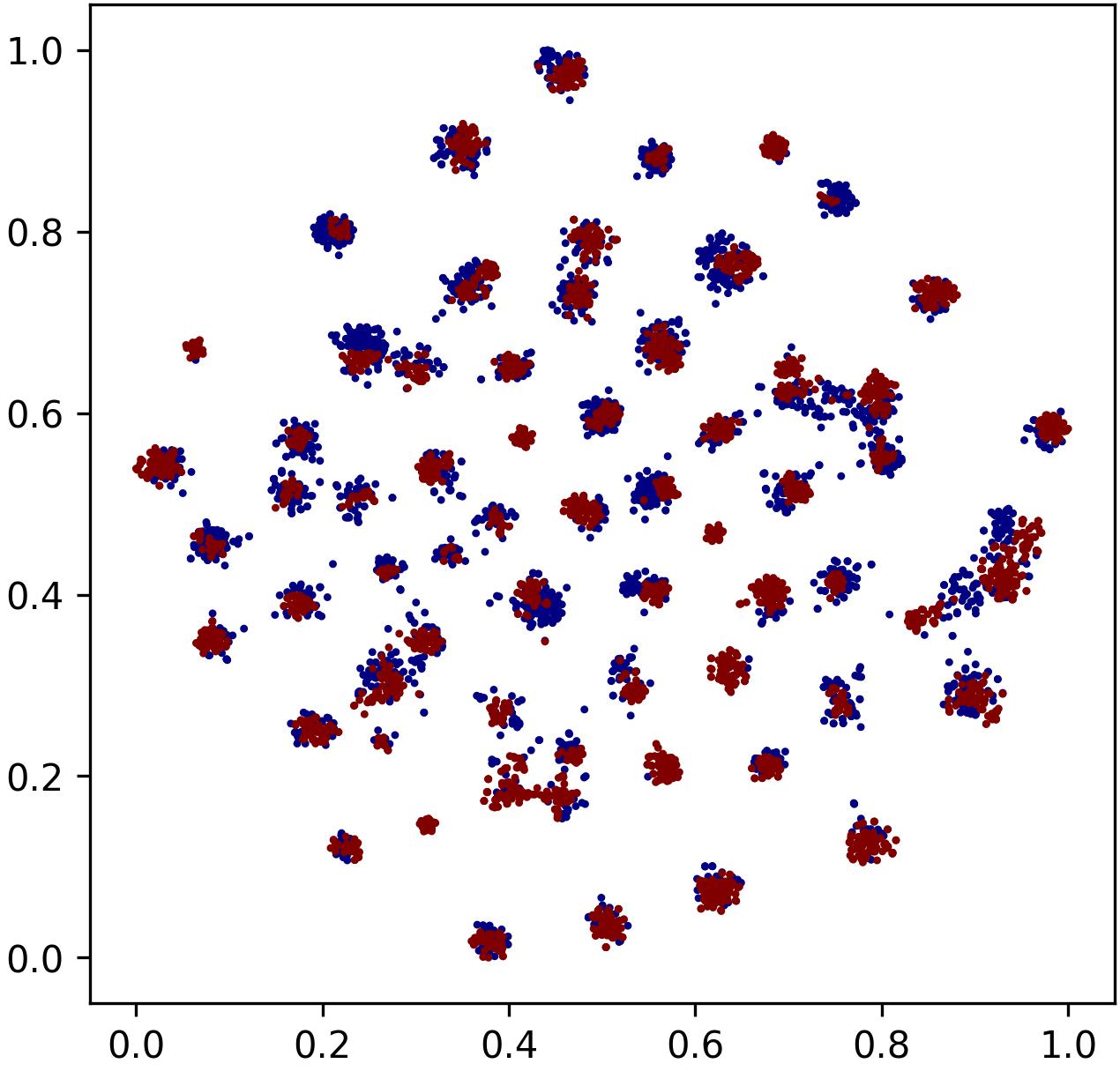}
\end{minipage}
}
\subfloat[Selected C and P]{
\begin{minipage}[t]{0.3\linewidth}
\centering
\includegraphics[height=2.5cm,width=2.5cm]{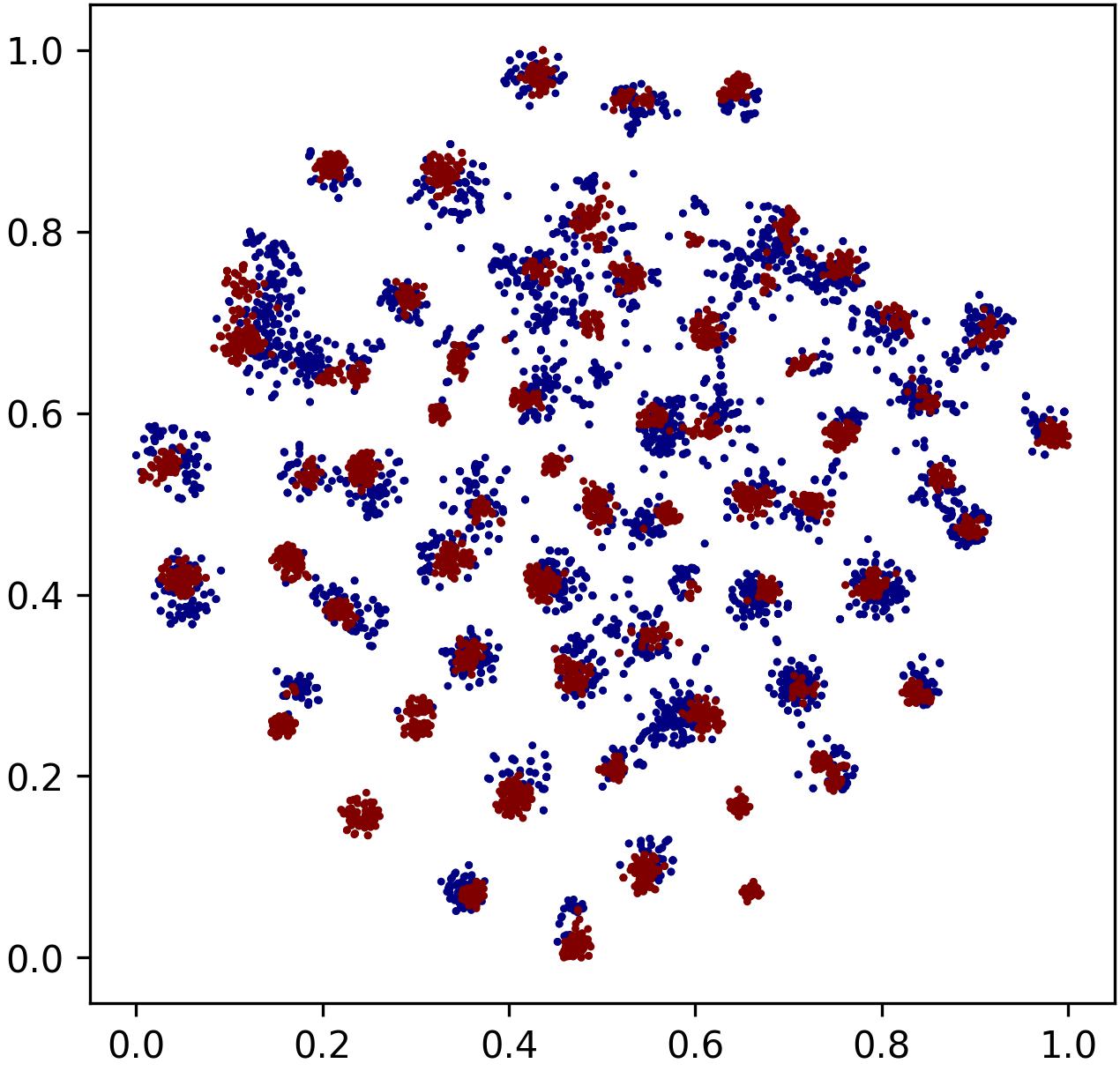}
\end{minipage}
}
\subfloat[Selected R and P]{
\begin{minipage}[t]{0.3\linewidth}
\centering
\includegraphics[height=2.5cm,width=2.5cm]{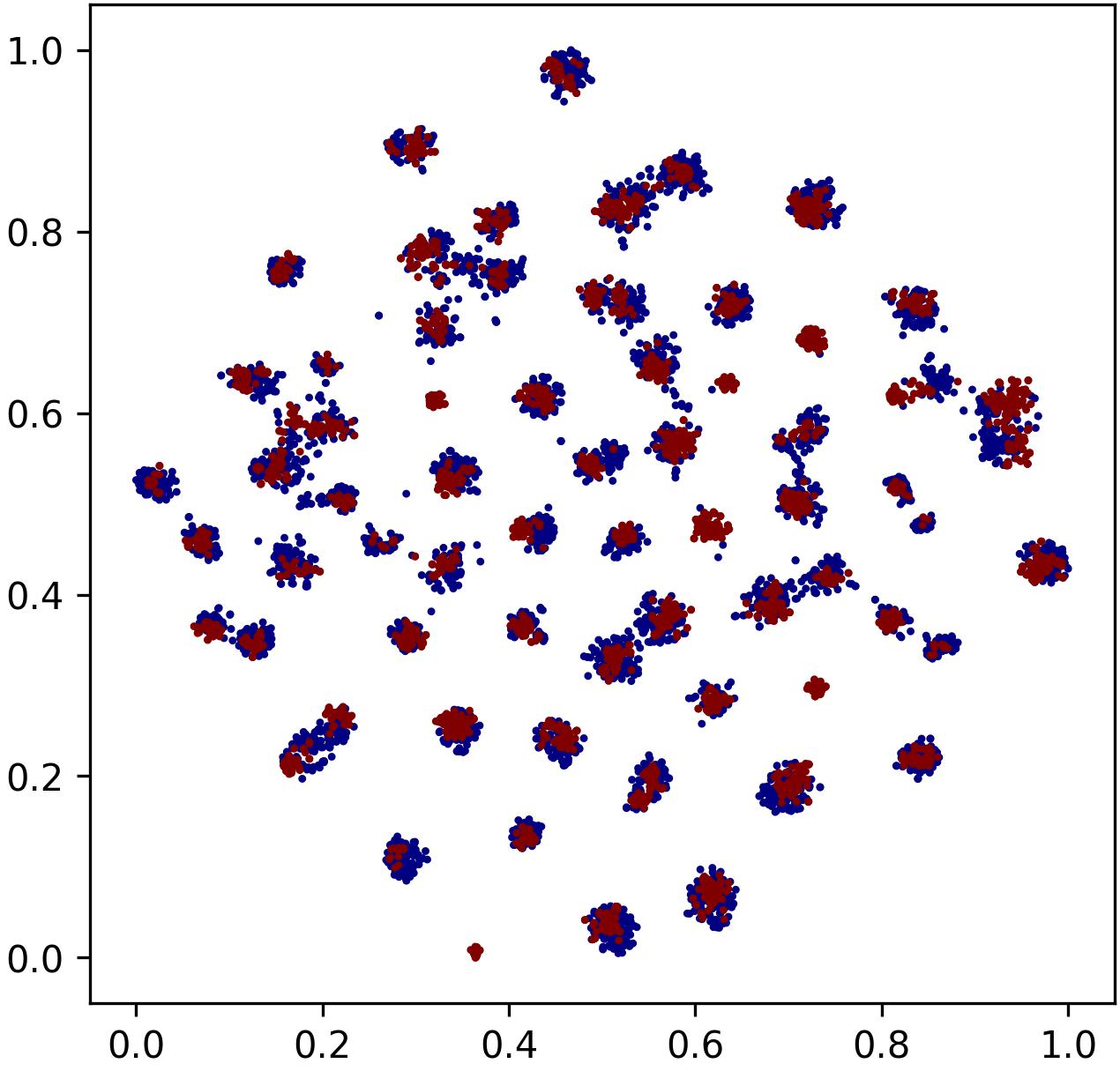}
\end{minipage}
}

\subfloat[Adapted A and P]{
\begin{minipage}[t]{0.3\linewidth}
\centering
\includegraphics[height=2.5cm,width=2.5cm]{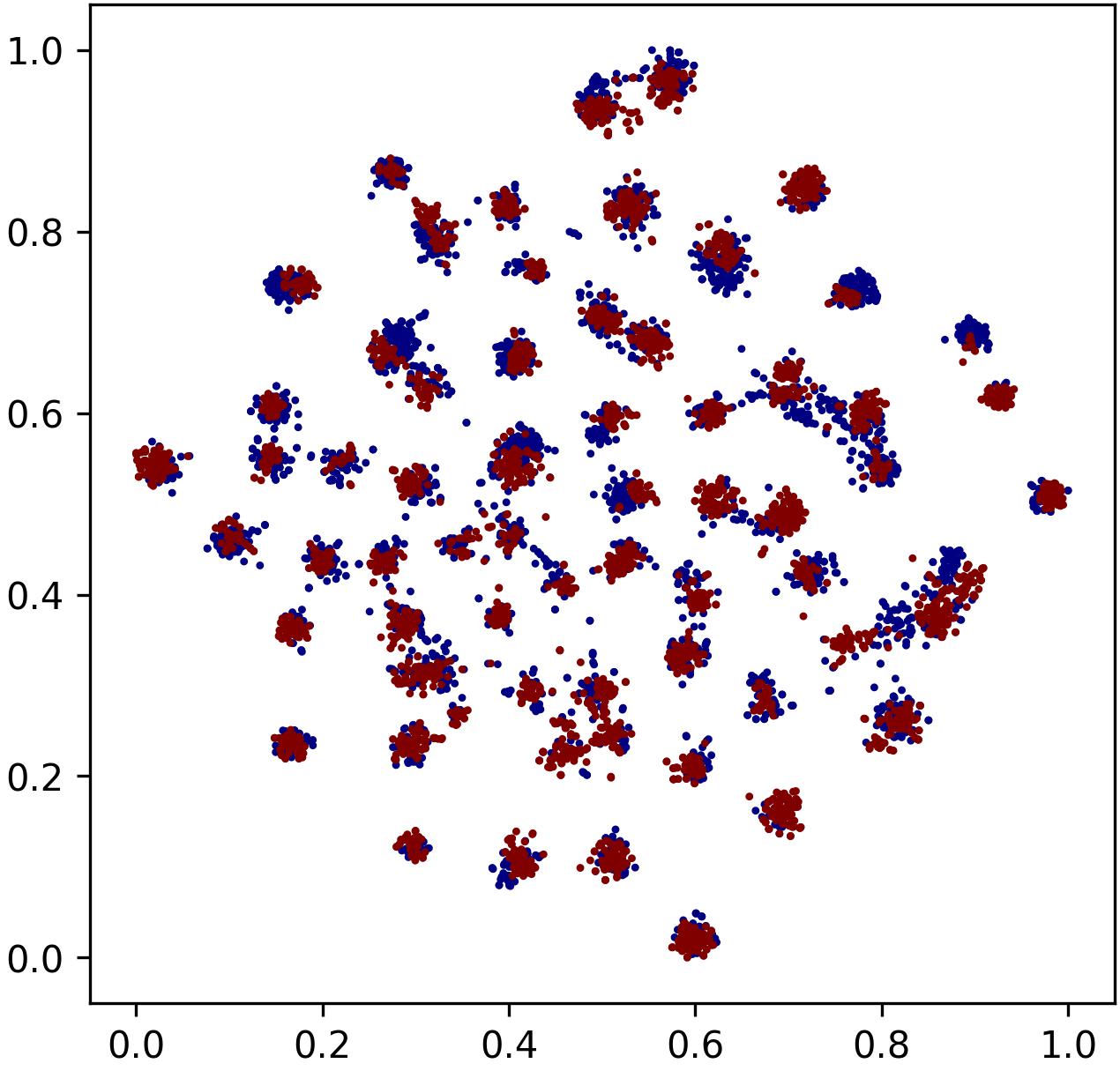}
\end{minipage}
}
\subfloat[Adapted C and P]{
\begin{minipage}[t]{0.3\linewidth}
\centering
\includegraphics[height=2.5cm,width=2.5cm]{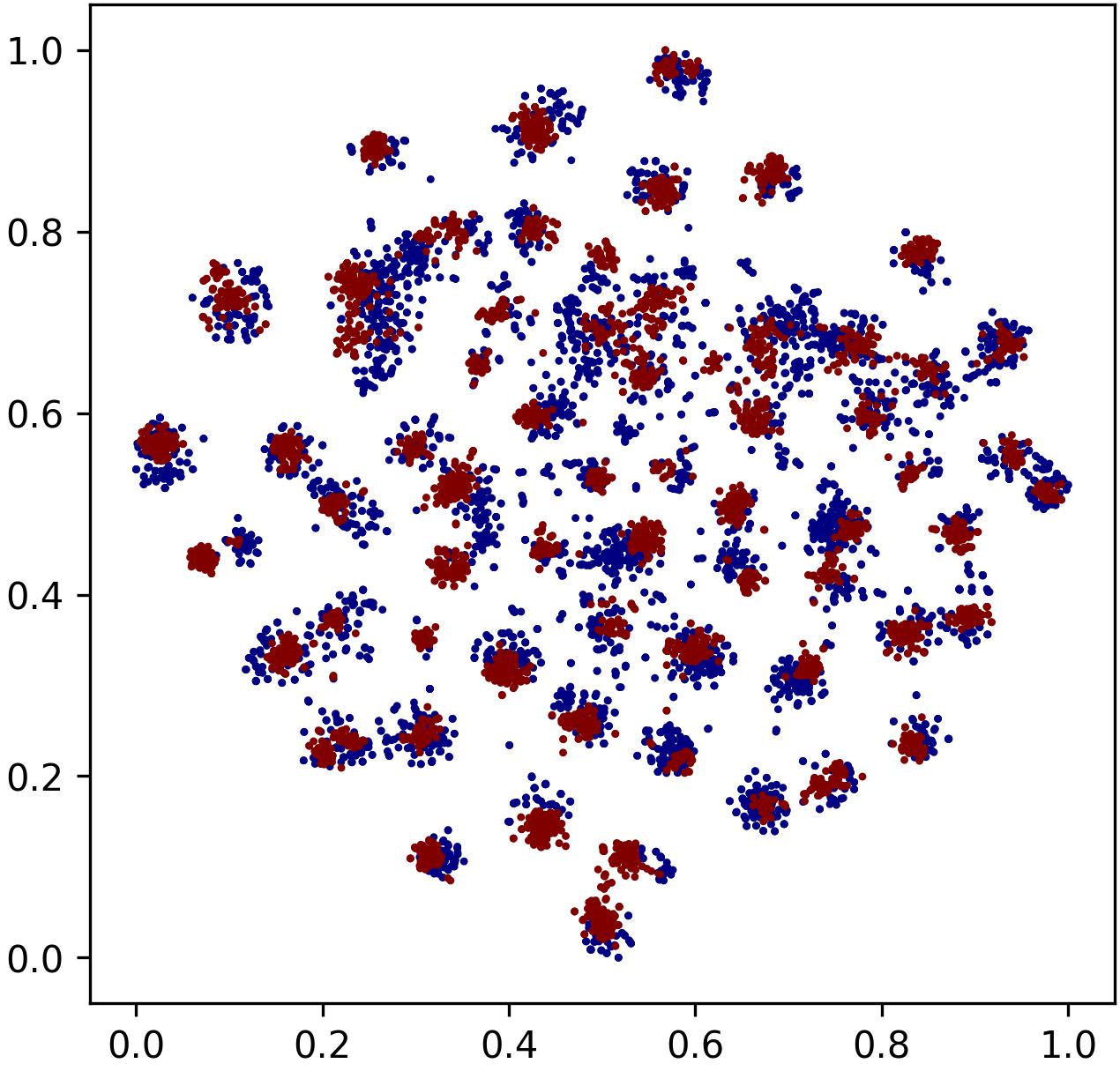}
\end{minipage}
}
\subfloat[Adapted R and P]{
\begin{minipage}[t]{0.3\linewidth}
\centering
\includegraphics[height=2.5cm,width=2.5cm]{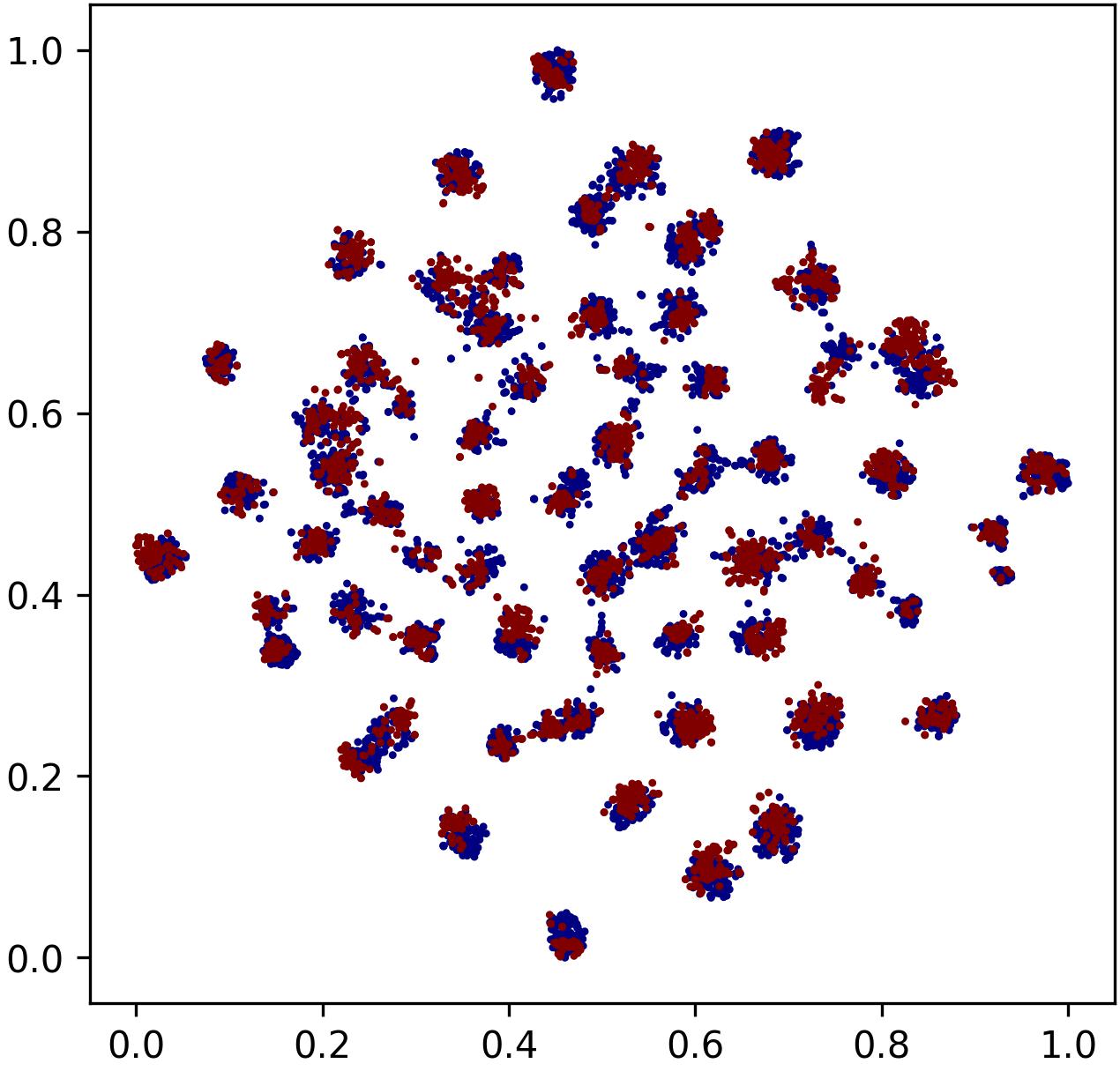}
\end{minipage}
}
\caption{T-SNE of adapted source and target data on task P from OfficeHome.}\label{tsneP}
\end{figure}

\begin{figure}[htbp]
\centering
\subfloat[Adapted A]{
\begin{minipage}[t]{0.3\linewidth}
\centering
\includegraphics[height=2.5cm,width=2.5cm]{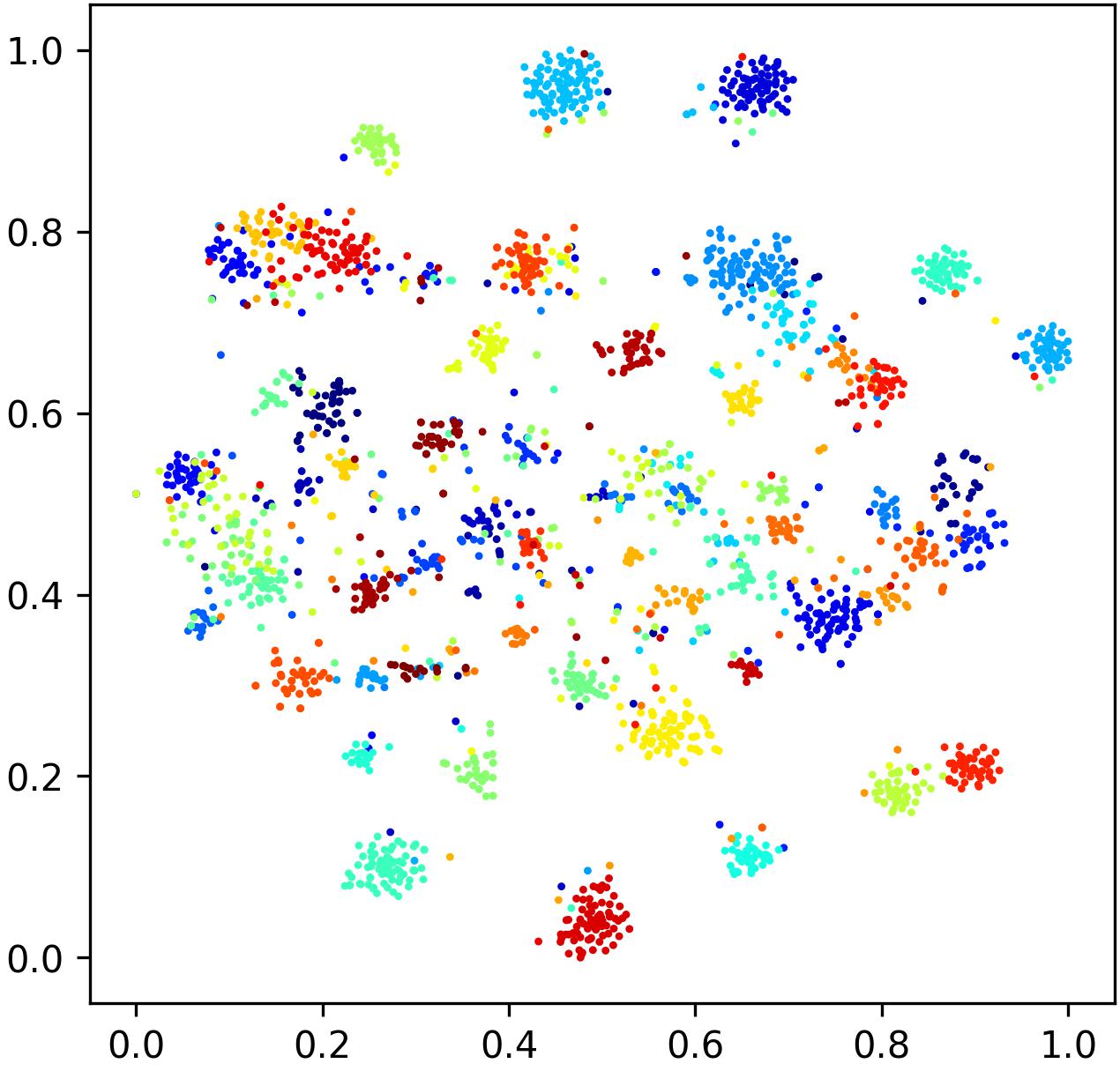}
\end{minipage}
}
\subfloat[Adapted C]{
\begin{minipage}[t]{0.3\linewidth}
\centering
\includegraphics[height=2.5cm,width=2.5cm]{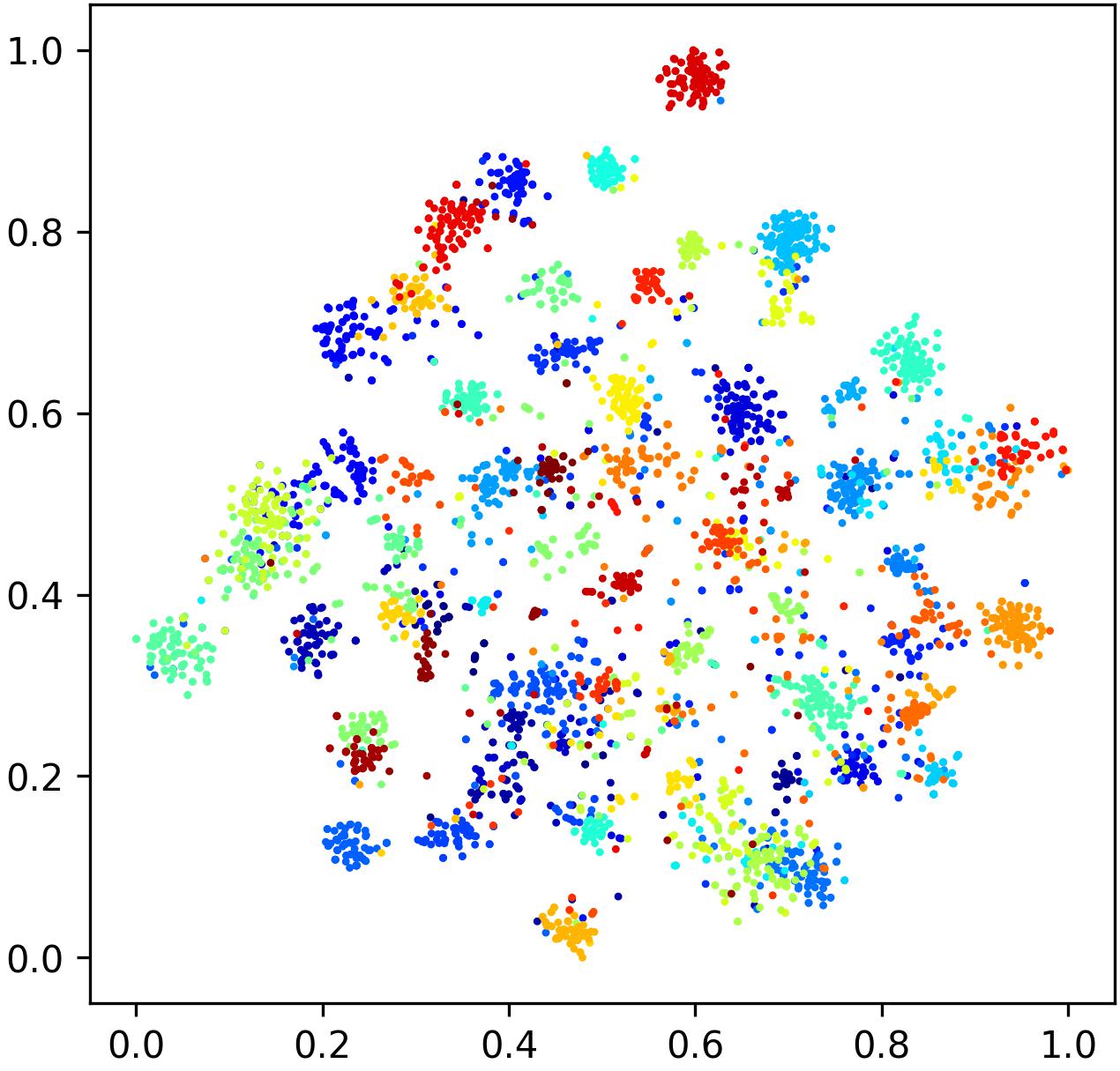}
\end{minipage}
}

\subfloat[Adapted P]{
\begin{minipage}[t]{0.3\linewidth}
\centering
\includegraphics[height=2.5cm,width=2.5cm]{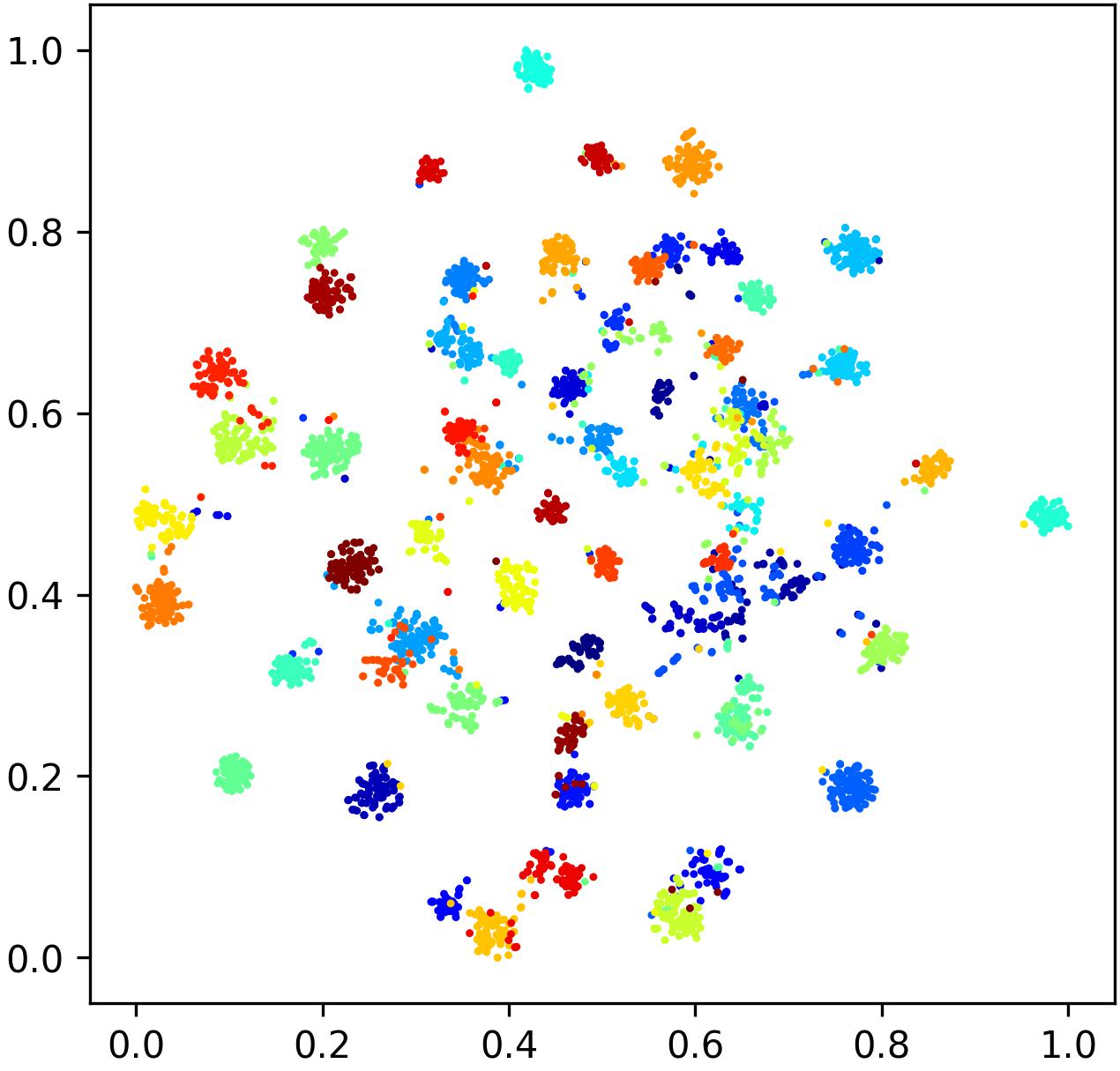}
\end{minipage}
}
\subfloat[Adapted R]{
\begin{minipage}[t]{0.3\linewidth}
\centering
\includegraphics[height=2.5cm,width=2.5cm]{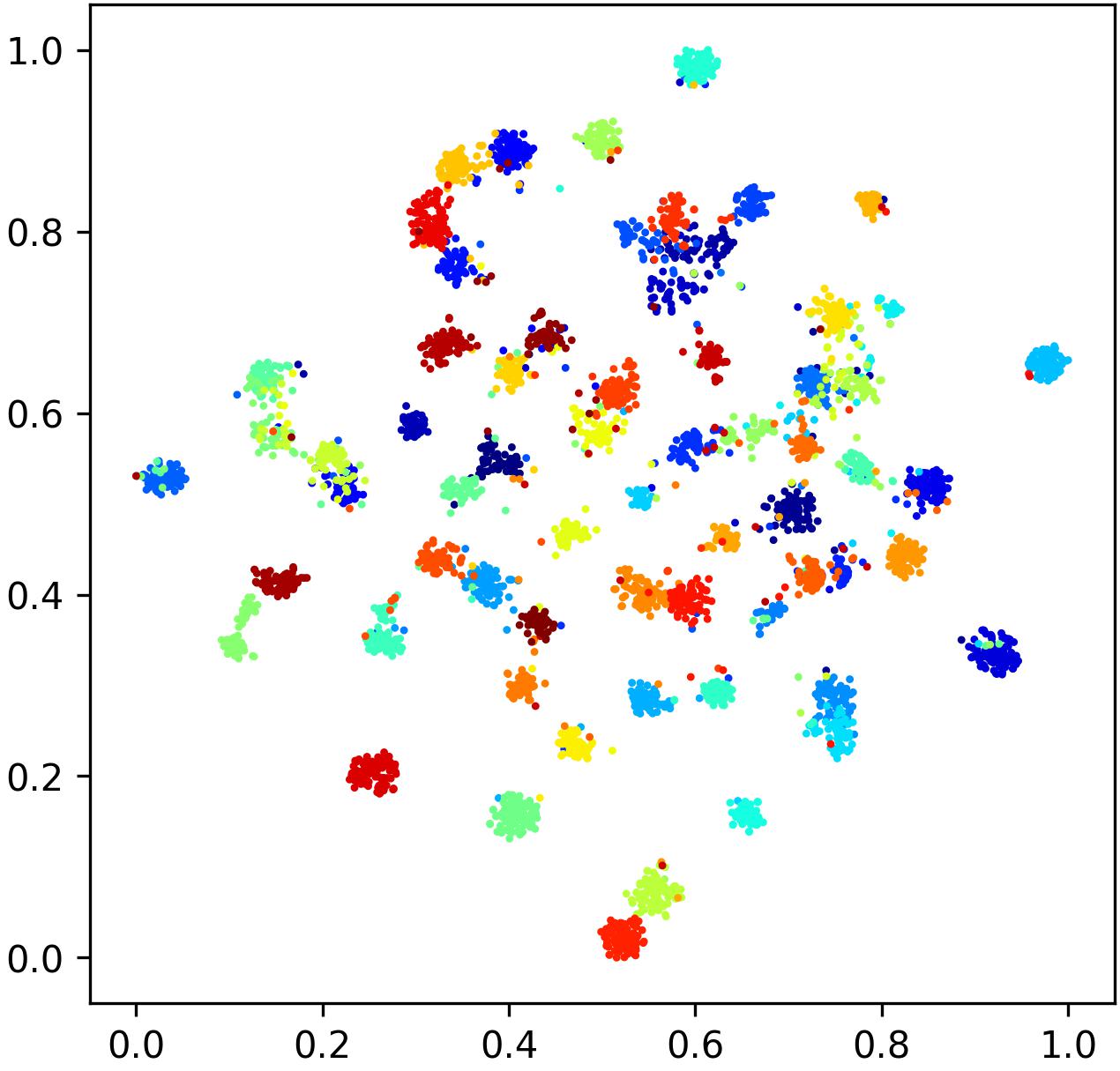}
\end{minipage}
}
\caption{T-SNE of target data under source-free setting from OfficeHome.}\label{tsnesf}
\end{figure}
\end{document}